\RequirePackage{fix-cm}

\documentclass[twocolumn]{svjour3}

\usepackage[numbers,sort&compress]{natbib}

\usepackage[table]{xcolor}
\smartqed  %
\usepackage{tikz,tikz-cd, pgf}
\usepackage{amsmath}
\usepackage{amssymb}
\usepackage{stmaryrd}
\usepackage{amsfonts}
\usepackage{algorithm}
\usepackage[noend]{algorithmic}
\usepackage{setspace}
\usepackage{verbatim}
\usepackage{comment} 
\usepackage{url}
\usepackage{enumitem}
\usepackage{float}
\restylefloat{table}

\usepackage{graphicx}
\usepackage{tabularx}
\usepackage{rotating}
\usepackage{booktabs}
\definecolor{lightgray}{gray}{0.95}
\usepackage{hyphenat}

\usepackage[normalem]{ulem}

\usepackage[colorlinks=true, allcolors=blue]{hyperref}
\usepackage{multirow}

\newcommand*\rot{\rotatebox{90}}

\DeclareMathOperator*{\argmin}{arg\,min}

\def\erfc{{\text{erfc}}}

\journalname{Journal of Mathematical Imaging and Vision}

\begin{document}\sloppy

\title{Image Anomalies: a Review and Synthesis of Detection Methods}

\author{Thibaud Ehret$^{\star}$ \and Axel Davy$^{\star}$ \and Jean-Michel Morel \and Mauricio Delbracio \thanks{$\star$ Both authors contributed equally to this work.\newline Work supported by IDEX  Paris-Saclay IDI 2016, ANR-11-IDEX-0003-02, ONR  grant N00014-17-1-2552,  CNES MISS project, Agencia Nacional de Investigaci\'on e Innovaci\'on (ANII, Uruguay) grant FCE\_1\_2017\_135458, DGA Astrid ANR-17-ASTR-0013-01, DGA ANR-16-DEFA-0004-01,  Programme ECOS Sud -- UdelaR - Paris Descartes U17E04, and MENRT. }}
\institute{Thibaud Ehret \and Axel Davy \and Jean-Michel Morel \at CMLA, ENS Cachan, CNRS, Universit\'e Paris-Saclay, 94235 Cachan, France \\ \email{thibaud.ehret@ens-cachan.fr}\\ \email{axel.davy@ens-cachan.fr} \and Mauricio Delbracio \at IIE, Facultad de Ingenier\'ia, Universidad de la Rep\'ublica, Uruguay}

\date{Received: date / Accepted: date}

\maketitle

\begin{abstract} We review the broad variety of methods that  have been  proposed for anomaly detection in images. Most methods found in the literature  have in mind a particular application.  Yet we focus on  a classification  of the methods based on the structural  assumption they make on the ``normal'' image, assumed  to obey a ``background model''. Five different structural assumptions emerge for the background model.  Our  analysis leads  us to reformulate  the  best  representative algorithms in each class by attaching to them  an \textit{a-contrario} detection that  controls the  number of  false positives and thus deriving a uniform detection scheme for all.  By combining the most general structural  assumptions expressing the background's normality with the proposed generic statistical detection tool,  we end up proposing several generic algorithms that seem to generalize or reconcile most methods.
We  compare the  six best  representatives  of our  proposed classes of algorithms on anomalous images taken from classic papers on the subject,  and on a synthetic database. Our  conclusion hints that it is  possible to perform automatic anomaly detection on a single image.

\end{abstract}


 \section{Introduction}

\begin{figure*}[t]
 \centering
\includegraphics[width=.2\textwidth]{Experiments/Grosjean/orig}
\includegraphics[width=.2\textwidth]{Experiments/Papers/mine2}
\includegraphics[width=.2\textwidth]{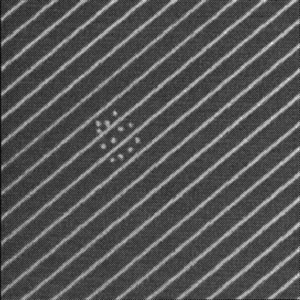}
\includegraphics[width=.2\textwidth]{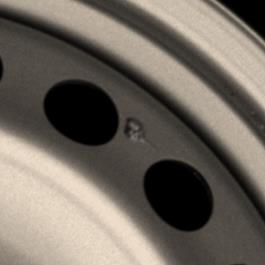}

\caption{Examples of industrial images with anomalies to detect. From left to right a suspicious mammogram~\cite{grosjean2009contrario}, an undersea mine~\cite{mishne2014multiscale}, a defective textile pattern~\cite{tsai1999automated} and a defective wheel~\cite{toutthesis}.}

\label{fig:teaser}
\end{figure*}

The automatic detection of anomalous structure in arbitrary images is concerned with the problem of finding non-confirming patterns with respect to the image normality. This is a challenging problem in computer vision, since there is no clear and straightforward definition of what is (ab)normal for a given arbitrary image. Automatic anomaly detection  has high stakes in  industry, remote sensing and medicine (Figure \ref{fig:teaser}).  It  is  crucial to be able to handle automatically massive data to  detect for example anomalous masses in mammograms~\cite{tarassenko1995novelty,grosjean2009contrario},
 chemical targets in multi-spectral and hyper-spectral satellite images \cite{ashton1998detection,schweizer2000hyperspectral,stein2002anomaly,du2008automated}, sea mines in side-scan sonar images~\cite{mishne2013multiscale}, or defects in industrial monitoring applications~\cite{zontak2010defect,yeh2010wavelet,tout2017automatic}.  This detection may be done using any imaging device from cameras to scanning electron microscopes~\cite{carrera2016scale}.
 
Our goal here  is  to review the broad variety of methods that  have been  proposed for this problem in the realm of image  processing. We would like to classify the methods, but also to decide if some arguably general anomaly  detection  framework emerges  from the analysis. This  is not obvious: most reviewed methods were  designed for a particular application, even if most claim some degree of generality. 

Yet, all anomaly  detection  methods  make a  general structural assumption  on the ``normal'' background that actually characterizes the method. By combining the most general structural assumptions with statistical detection tools controlling  the number of false alarms,  we shall converge to a few generic algorithms that seem to generalize or reconcile most methods. 

To evaluate our conclusions, we shall  compare   representatives  of the  main algorithmic classes on classic and diversified  examples. A fair comparison will require completing   them when necessary with a common statistical decision threshold.

 \paragraph{Plan of  the paper.}
 In the  next subsection \ref{sectionformalmodel}, we make  a first sketch of definition of the problem, define the main  terminology and  give the notation for the statistical framework used throughout the paper. Section \ref{sec:quick_review_reviews} reviews four anterior reviews and discusses their methodology. Section \ref{sec:whatanomalydetectionsisnt} circumscribes our field of  interest by excluding several related but  different questions.  In the central Section~\ref{sec:review} we propose a classification of the anomaly detectors into five classes depending on the main structural assumption made on the background model. This section contains the description and analysis of about 50 different methods. This analysis raises the question of defining a uniform detection scheme for all background structures. Hence, in Section~\ref{sec:generalization} we incorporate a uniform probabilistic detection threshold to the  most relevant methods spotted in Section~\ref{sec:review}.  This enables us in
Section~\ref{sec:experiments} to build three comparison protocols for six methods representative of  each class.  We finally conclude in Section~\ref{sec:discussion}.

   \subsection{Is there a formal generic framework for  the problem?\label{sectionformalmodel}}
   Because of the variety of methods proposed, it is virtually impossible to start with a formal definition of the problem.   Nevertheless, this subsection circumscribes it and lists the most important terms and concepts recurring in most papers. Each new  term will be indicated in italic.
   
   Our study is limited to \textit{image} anomalies for obvious experimental reasons: we need a common playground to compare methods. Images have a specific geometric structure and homogeneity which is different from (say) audio or text. For example, causal anomaly detectors based on predictive methods such as autoregressive conditional heteroskedasticity (ARCH) models fall out of our field. (We shall nevertheless study an  adaptation of ARCH to anomaly detection in sonar images.)
   
 Like in the  overwhelming majority of reviewed papers,  we assume that anomalies can be detected  in and from a single image, or from an image data set,  even if  they do contain anomalies.   Learning the  background or ``normal''  model from images containing anomalies nevertheless implies that anomalies are small,  both in size and proportion to the processed images, as stated for example in~\cite{olson2018manifold}:  
   \begin{quote}``We consider the problem of detecting points that are rare within a data set dominated by the presence of ordinary background points.''\end{quote}

Without loss of generality, we shall evaluate the methods on single images. It appears that for the overwhelming majority of  considered methods, a single image has enough samples to learn a background model. As a matter of fact, many methods are proceeded locally in the  image or in a feature space, which implies that the background model for each detection test is learned only on a well chosen portion of the image or of the samples. Nevertheless for industrial applications, using a fixed database representative of anomaly-free images can help reduce false alarms and computation time, and studied methods can generally be adapted to this scenario.
All methods extract vector samples from the images, either \textit{hyperspectral pixels} generally denoted by $x_i, x_j, x_r\cdots$, or  \textit{image patches}, namely sub-images of the image $u$ with moderate size, typically from $2\times 2 $ to $16\times 16$, generally denoted by $p_i,  p_j,  q_r, q_s,\cdots$.  The vector samples may be also obtained as a feature vector obtained by a linear transform (e.g. wavelet coefficients) or by a linear or nonlinear coordinate  transform such as, PCA, kernel PCA or diffusion maps, or  as  coordinates in a sparse dictionary. We denote  the  resulting  vector representing a sample by $\tilde x_i, \tilde y_i,\cdots$ or  $\tilde  p_i, \tilde q_i, \cdots$.
   
From these samples taken from an image (or from a collection  of images), all considered anomaly detection methods estimate (implicitly or explicitly) a \textit{background model},  also known as  model of \textit{normal} samples. The goal of the  background model is to provide for each sample a measure of its \textit{rareness}. This rarity measure is generally called a \textit{saliency map}.  It requires an \textit{empirical threshold} to decide which pixels or patches are salient enough to be called anomalies. If the  background model is stochastic,  a \textit{probability of false alarm} or  \textit{p-value} can be associated with each sample, under the  assumption  that it obeys the background model.  

The methods will be mainly characterized by the structure of their  background model.   This model may be \textit{global in the image},  which means common to all the image  samples, but also \textit{local} in the image (for \textit{ center-surround} anomaly detectors), or \textit{global in the sample space} (when a global  model is given for all samples regardless of their position in the  image).  The  model may remain \textit{local in the sample space} when the sample's anomaly is evaluated by comparing  it to its neighbors in the patch  space or in the space of hyperspectral pixels.  When samples are compared locally in  the sample space but can be taken from all over the image, the method is often called \textit{non-local}, though  it can actually be local in the  sample space.

Many methods proceed to a \textit{background  subtraction}. 
This operation, which can be performed in many  different ways that we will explore, aims at removing from the data all ``normal'' variations, attributable to the background model, thus enhancing the abnormal ones, that is, the  anomalies. 

At  the  end of  the game, all methods compute for each sample its \textit{distance to the background} or \textit{saliency}.  This distance must be larger than a given value (threshold) to decide if the  sample is anomalous. The  detection threshold may be empirical,  but is preferably obtained through a  statistical argument. To explicit the formalism, we shall  now detail  a classic method.

\citet{du2011random}~proposed to learn a Gaussian background model from randomly picked $k$ dimensional image  patches in a hyperspectral image.  Once this background model $p\sim {\mathcal{N}}\left(\mu ,\Sigma \right)$ with mean $\mu$ and covariance matrix $\Sigma$ is obtained, the  anomalous $(2\times  2)$ patches are detected using a threshold on their \textit{Mahalanobis distance} to the background 
$$
d_\mathcal{M}(p_i):=\sqrt{(p_i-\mu)\Sigma^{-1}(p_i-\mu)}.
$$ 
Thresholding  the Mahanalobis distance boils down to a simple $\chi^{2}$ test.  Indeed, one has  $d_\mathcal{M}^{2}(p)\sim \chi_{k}^{2}$, meaning that the  square of the Mahalanobis distance between $p$ and its expectation obeys a $\chi^2$ law  with $k$ degrees of freedom. Let  us denote by  $\chi_{k;1-\alpha }^{2}$ the quantile $1-\alpha$, then $$\mathbb {P} \left[d_\mathcal{M}^{2}(p)\leq \chi_{k;1-\alpha }^{2}\right]=1-\alpha =\mathbb{P}\left[p\in ZT_{\alpha }\right],$$  where $ZT_{\alpha}:=\left\{p\in \mathbb{R}^{k}\mid d_\mathcal{M}^{2}(p)\leq \chi _{k;1-\alpha }^{2}\right\}$ is the  \textit{$\alpha$-tolerance zone}.
Thus, $\alpha$ is the \textit{p-value} or \textit{probability of false alarm} for an anomaly under the Gaussian background problem: If indeed 
 $d_\mathcal{M}^{2}(p)> \chi_{k;1-\alpha }^{2}$,  then the probability  that $p$ belongs the background is lower than~$\alpha$.  
 
 Yet, thresholding the p-value may lead to  many false detections.  Indeed, anomaly detectors perform a very large number of tests, as  they typically test  each pixel.  For that reason, \citet{desolneux2004}, \cite{desolneux2007gestalt} pointed out  that in image analysis computing a \textit{number of false alarms} (NFA), also commonly called \textit{per family error rate} (PFER) is  preferable. 
 Assume that the above  anomaly test is performed for all $N$ pixels $p_i$ of an image. Instead of fixing a p-value for each pixel, it is sound to  fix a tolerable number $\alpha$ of false alarms  per image.  Then the  ``Bonferroni correction'' requires our test on $p$ to  be
 $d_\mathcal{M}^{2}(p)> \chi_{k;1-\frac \alpha N}^2$.
 We  then  have 
\begin{align*}
\mathbb{P}&\left(\bigcup_{i=1}^N[d_\mathcal{M}^{2}(p_i)> \chi_{k;1-\frac \alpha N}^2] \right) \\ \leq &\sum_{i=1}^N\mathbb{P}\left([d_\mathcal{M}^{2}(p_i)> \chi_{k;1-\frac \alpha N}^2]\right)=N\frac{\alpha}N=\alpha,
\end{align*}
which means that  the probability of detecting at least one  ``false anomaly'' in the background is equal to $\alpha$.  It  is convenient to reformulate  this Bonferroni estimate in terms of  expectation of  the number of false alarms:
 \begin{align*}\mathbb{E}&\left[\Sigma_{i=1}^N\mathbf{1}_{[d_\mathcal{M}^{2}(p_i)> \chi_{k;1-\frac \alpha N}^2]}\right]\\ = &\sum_{i=1}^N\mathbb{E}\mathbf{1}_{[d_\mathcal{M}^{2}(p_i)> \chi_{k;1-\frac \alpha N}^2]}=N\frac{\alpha}N=\alpha\end{align*}
 where $\mathbf{1}$ denotes the characteristic function  equal to 1 if and only if its argument is positive.
 This means that by fixing a lower threshold equal to $\chi_{k;1-\frac \alpha N}^2$ for the distance $d_\mathcal{M}^{2}(p)$, we  secure on average $\alpha$ false alarms per image. 

  We can compare this  unilateral  test to standard statistical decision  terms. The  final step of an anomaly detector would be to decide between two assumptions:  
  \begin{itemize}
      \item $\mathcal H_0$: the sample $p$ belongs to the background;
      \item $\mathcal H_1$: the sample $p$ is too exceptional under $\mathcal H_0$ and  is therefore an anomaly.
  \end{itemize}
Because  no model is at  hand for  anomalies, $\mathcal H_1$ boils down to a mere negation of $\mathcal H_0$.   $\mathcal H_1$  is chosen with a  probability of false alarm $\frac \alpha N$ and therefore with a number of false alarms (NFA) per image equal to $\alpha$.  We shall give  more  examples of NFA computations in Section~\ref{sec:generalization}.

\subsection{ A quick review  of  reviews}
\label{sec:quick_review_reviews}
More than 1000 papers  in  Google scholar contain the  key  words ``anomaly detection'' and ``image''. The existing review papers proposed a useful classification, but leave open the question of the  existence of  generic algorithms \textit{performing unsupervised  anomaly detection on  any image}. 
The 2009 review paper by~\citet{chandola2009anomaly} on anomaly detection is arguably the most complete review. It considered allegedly all existing techniques and all application fields and reviewed 361 papers. The review establishes a distinction between \textit{point anomaly, contextual anomaly, collective anomalies}, depending on whether the background is steady or evolving and the anomaly has a larger scale than the initial samples. It also distinguishes between \textit{supervised, mildly supervised} and \textit{unsupervised} anomalies. It revises the main objects where anomalies are sought for (images, text, material, machines,  networks, health, trading, banking operations, etc.) and lists the preferred techniques in each domain. Then it finally proposes the following classification of all involved techniques. 
\begin{enumerate}
\item {\bf Classification based anomaly detection, e.g., SVM, Neural networks.}
These techniques train a classifier to distinguish between normal and anomalous data in the given feature space. Classification is either multi-class (normal versus abnormal) or one-class (only trains to detect normality, that is, learns a discriminative boundary around normal data).
Among the one-class detection methods we have the \textit{Replicator Neural Networks} (\textit{auto-encoders}).

\item {\bf Nearest neighbor based anomaly detection.} The basic assumption of these methods is that normal data instances occur in dense neighborhoods, while anomalies occur far from their closest neighbors.
 This can be measured by the distance to the $k^{\text{th}}$ nearest neighbor or as  relative density.

\item {\bf  Clustering based anomaly detection.} 
Normal data instances are  assumed to belong to a cluster in the data, while anomalies  are defined as those standing far from the centroid of their closest cluster.  
\item {\bf  Statistical anomaly detection.} 
Anomalies are defined as observations unlikely to be generated by the ``background'' stochastic model. Thus, anomalies occur in the low probability regions of the background model.
Here the background models can be: \textit{parametric} (Gaussian, Gaussian mixture, regression), or \textit{non-parametric} and  built, e.g., by a kernel method.  

\item {\bf  Spectral anomaly detection.}
The main tool here is  principal component analysis (PCA) and its generalizations. Its principle is that an anomaly has deviant coordinates with respect to normal PCA coordinates. 

\item {\bf Information theoretic anomaly detection.} These techniques analyze the information content of a data set using information theoretic measures, such as, the Kolomogorov complexity, the entropy, the relative entropy, among others.  
\end{enumerate}
This excellent review is perhaps nevertheless biting off more than it could possibly chew. Indeed, digital materials like sound, text, networks, banking operations, etc.  are so different that it was impossible to examine in depth the role  of their specific structures for anomaly detection. By focusing on images,  we shall have a much focused discussion involving their specific structure yielding natural vector samples (color or  hyperspectral, pixels,  patches) and specific structures  for these samples, such as self-similarity and sparsity.

The above review by \citet{chandola2009anomaly} is fairly well completed by the more recent review  by \citet{pimentel2014review}. This paper presents a \textbf{complete} survey of novelty detection methods and introduces a classification into five groups. 

\begin{enumerate}
\item  {\bf Probabilistic novelty detection.}   These methods are based on estimating a generative probabilistic model of the data (either parametric or non-parametric).

\item {\bf Distance-based methods.} These methods rely on a distance metric to define similarity among data points (clustering, nearest-neighbour and self-similar methods are included here).

\item {\bf Reconstruction-based methods.}  These methods seek to model the normal component of the data (background), and the reconstruction error or residual is used to produce an anomaly score.

\item {\bf Domain-based methods.} They determine the location of the normal 
data boundary using only the data that lie closest to it, and do not 
make any assumption about data distribution.

\item {\bf Information-theoretic methods.} These methods require a measure (information content) that is sensitive enough to detect the effects of anomalous points in the dataset. Anomalous samples, for example, are detected by a local Gaussian model, which starts this list.
 
\end{enumerate}

Our  third reviewed  review was devoted to  anomaly detection  in hyperspectral imagery \cite{matteoli2010tutorial}. It  completes three previous comparative studies, namely \cite{stein2002anomaly}, \cite{hytla2007anomaly} and \cite{matteoli2007comparative}. \citet{matteoli2010tutorial}~conclude that  most of the techniques try to cope with background non-homogeneity,
and attempt to remove it by de-emphasizing the main structures in the image, which we can interpret as background subtraction. 

 For the same authors, an  anomaly
 can be defined as an observation that deviates in some way from the background clutter.  The
background itself can be identified  from a local neighborhood surrounding the observed pixel, or from
a larger portion of the image.  They also suggest that the  anomalies  must be sparse and small to make sense  as anomalies. Also, no a-priori knowledge about the target's spectral
signature should  be required. The question  in hyperspectral imagery therefore is to ``find  those pixels whose spectrum significantly differs from the
background''.
We can summarize the findings of this review by examining the five detection techniques that are singled out:

\begin{enumerate}
\item {\bf Modeling the background as a locally Gaussian model} \cite{reed1990adaptive} and detecting anomalous pixels by their Mahanalobis distance to the local Gaussian model learned from its surrounding at some distance. This famous method is called the RX (Reed-Xiaoli) algorithm. 
\item {\bf Gaussian-Mixture Model based Anomaly Detectors }\cite{stein2002anomaly,ashton1998detection,hazel2000multivariate,carlotto2005cluster}. The optimization  is done  by stochastic expectation  minimization \cite{masson1993sem}. The detection methodology is similar to the  locally Gaussian model, but the main difference is that background modeling becomes global instead of local. 

The technical difficulty raised by this more complex model is the  variety of clustering  algorithms that  can be used \cite{duran2005time,duran2007time}, and the  thorny question of finding the  adequate number of  clusters as  addressed in \cite{duran2006anomaly} and \cite{penn2002using}.

\item {\bf The Orthogonal Subspace Projection approach}. It performs a background estimation via a projection  of pixel  samples on their main components after an SVD has been applied to all samples. Subtracting the  resulting image amounts to a background subtraction and therefore delivers an image where noise and the  anomalies dominate.  
\item {\bf The kernel RX algorithm}
\cite{kwon2005kernel} which  proceeds by defining a  (Gaussian) kernel distance between pixel samples and considering that it represents a Euclidean distance in a higher dimension feature space.  (This technique is also proposed in \cite{mercier2006partially} for oil slick detection.)
A local variant of this method~\cite{ranney2006hyperspectral}
performs an OSP suppression of the background, defined as one of the  four  subspaces spanned by the pixels within four neighboring subwindows surrounding  the pixel at some distance.

\item  {\bf Background support region estimation by Support Vector Machine} \cite{banerjee2006support}. Here the idea is that it is not necessary to model the  background, but that  the main question is to model its support and to define anomalies as observations away from this support. 
\end{enumerate}

Our last  reviewed review, by~\citet{olson2018manifold}, compares ``manifold learning techniques for unsupervised anomaly detection''  on simulated and real images.
 Manifold methods assume that the background samples span a manifold rather than a linear space. Hence PCA might be suboptimal and must be replaced by a nonlinear change of coordinates. The authors of the review consider and  three kinds for this change of coordinates:
 
 \begin{enumerate}
     \item  {\bf Kernel PCA,}  introduced
by Sch\"{o}lkopf et al. \cite{scholkopf1998nonlinear} and adapted to the anomaly detection
problem by Hoffmann \cite{hoffmann2007kernel}.
     
     \item  {\bf The Parzen density estimator}, which is  actually  interpreted as the  simplest  instance  of kernel PCA \cite{parzen1962estimation,hoffmann2007kernel}.
     \item {\bf The  diffusion map} \cite{coifman2006diffusion,lafon2006data}, which in this framework appears as a variant of kernel PCA.
 \end{enumerate}
 We  shall review these techniques in more detail in Section~\ref{sec:review}.
   In these methods, the  sample manifold $\mathcal M$ is structured by a Gaussian ``distance'' $$k(x_j,x_j)=e^{-\frac 1 {h^2} ||x_i-x_j||^2}.$$  The methods roughly represent the samples by  coordinates computed from the  eigenvectors and eigenvalues of the matrix $K=(k(x_i,x_j))_{ij}$.   This amounts in  all cases to a nonlinear change of coordinates.  Then, anomalous samples are  detected as falling apart from the manifold. 
 The key parameter $h$ is  chosen in the examples so that the  isolevel  surface of the  distance function wraps tightly the inliers.  
 
 The review compares the  ROC curves of the different methods (PCA, kernel PCA, Parzen, diffusion map) and concludes that small ships on a sea landscape are better  detected by kernel  PCA. 
 Since the review only compares ROC curves between the different methods, it avoids addressing the detection threshold issue.

\paragraph{Discussion.}  The above four highly cited reviews made an excellent job of considering  countless papers and  proposing a categorization of methods.  Nevertheless, their final map of  the methods is an exhaustive inventory where methods are distributed according to what they do, rather than to what they assume on background and anomaly.
Nevertheless, the \citet{pimentel2014review} review is actually  close to  classify methods by structural assumptions on  the  background, and  we shall follow this  lead.
The above reviews do not conclude  on a unified statistical decision  framework.  Thus,  while  reusing  most  of their categories, we shall attempt at reorganizing the panorama according  to three main questions:
 \begin{itemize}
\item What is the structural assumption made on the  background:  in  other terms what is ``normal''?
\item How is the decision measurement computed?
\item How is the anomaly detection threshold defined and computed,  and  what  guarantees are met?
\end{itemize}

Our ideal goal would be to find out the weakest (and therefore most general) structural assumption on normal data, and to apply to it  the  most rigorous statistical test. In other words, the weaker the assumptions of normality, the more generic the detector will be.
Before proceeding to a  classification of anomaly detection  methods,  we shall examine several related questions which share some  of their tools  with anomaly  detection.

\subsection{What anomaly detection isn't \label{sec:whatanomalydetectionsisnt}}
\subsubsection{Not a classification problem}
 Most papers and reviews on anomaly detection  agree that multi-class classification techniques like SVM can be discarded, because anomalies are generally not observed in sufficient number and  lack statistical coherence. There are exceptions like the recent method introduced by~\citet{ding2014experimental}.  This paper assumes the disposition of enough anomalous samples to learn classification parameters from the data themselves.  Given several datasets with dimensions from 8 to 50 with moderate size (a few hundreds to a few thousand samples), this paper applies classic density estimators to  sizable extracts of the normal set (k-means, SVM, Gaussian mixture), then learns the optimal thresholds for each classifier and finally compares the performance of these classifiers.
 
 While in many surface defect detection problems, the defect can be of any shape or color, in some industrial applications known recurrent anomalies are the target of defect detectors. In this case a training database can be produced and the detection algorithm is tuned for the detection of the known defects \cite{jia2004intelligent, yang2002discriminative, xie2008review}. For example, \citet{soukup2014convolutional} proposed to detect rail defects in a completely supervised manner by training a classical convolutional neural networks on a dataset of photometric stereo images of metal surface defects. Another neural-network based method was proposed by~\citet{kumar2003neural}. This paper on the detection of local fabric defects, first performs a  PCA dimension reduction on $7\times7$ windows followed by the training  of a neural network on a  base of detects / non-detects, thus  again performing two-class classification. 

To detect changes on optical or SAR satellite images, many methods compare a pair of temporally close images, or more precisely the subtraction between them in the case of optical images \cite{bruzzone2002adaptive, Zanetti2015, Zanetti2018,Liu2017,Bovolo2006, Liu2015,Thonfeld2016}, or the log-ratio for SAR images \cite{bovolo2005adaptive, Li2018, celik2010change, jia2018novel}. However these methods often work on a pair of images where a change is known to have occurred (such as a forest fire \cite{bruzzone2002adaptive, celik2010change}, an earthquake \cite{washaya2018sar,Ferrentino2018} or a flood \cite{Li2018, Clement2017}), and thus have an \textit{a priori} for a two class distribution, which leads to classification techniques.

\paragraph{Conclusions.}

\subsubsection{More than a saliency measure\label{notasaliencymeasure}}
A broad related literature exists on saliency measures.  They associate to each image a saliency map, which is a scalar positive function that  can be visualized as an image where the brighter the  pixel, the  more salient it is. The goal of automatic saliency measures is to emulate the  human perception. Hence saliency measures  are often learned from a large set of  examples associating with images their average fixation maps by humans. For  example, \citet{tavakoli2011fast} designed an anomaly detector trained on average human fixation maps learning both the salient parts and their surround vectors as Gaussian vectors.  This reduced the problem to a two class Bayesian classification problem. 

The main difference with anomaly detectors is that many saliency measures try to mimic the human visual perception and therefore are allowed to introduce semantic prior knowledge related to the perceptual system (e.g., face detectors). This approach works particularly well with deep neural networks because attention maps obtained by gaze trackers can be used as a ground truth for the training step. SALICON by \citet{huang2015salicon} is one of these deep neural networks architecture achieving state of the art performance.

Saliency measures deliver saliency maps, in contrast to anomaly detectors that are requested to give a binary map of the anomalous regions. 
We can exclude from  our review supervised saliency methods based on learning from humans. Yet we cannot exclude the unsupervised methods that are based,  like anomaly detectors, on a structural model of the background. The only difference of  such saliency maps with anomaly detectors is that that anomaly detectors would require to add a last thresholding step after the saliency map is computed, to transform it into a binary detection map.

Interesting methods for example assign a saliency score to each tested pixel feature based on the inverse of the histogram bin value to which it belongs. 
In~\cite{riche2013rare2012} a saliency map is obtained by combining 32 multiscale oriented features obtained by filtering the image with oriented Gabor kernels. A weighted combination of the most contrasted channels for each orientation yields a unique multiscale orientation channel $c_o(i)$ for each orientation. Then the histograms $h_o$ of these channels $c_o$ are computed and each  pixel $i$ with value $c_o(i)$ is given a weight which is roughly inversely proportional to its value $h_o(c_o(i))$ in the histogram. 
The same rarity measurement is applied to the colors after PCA.  Summing all of these saliency maps one obtains something similar to what is observed with gaze trackers: the salient regions are the most visited.

Similarly,  image patches are represented by \citet{borji2012exploiting}  using their coefficients on a patch dictionary learned on natural images.   Local and global image patch rarities  are  considered as two ``complementary processes''.  Each patch is first represented by a vector of coefficients that linearly reconstruct it from a learned dictionary of patches from natural scenes (``normal'' data). Two saliency measures (one local and one global) are calculated and fused to indicate the saliency of each patch. The local saliency is computed  as the distinctiveness of a patch from its surrounding patches, while the global saliency is the inverse of a patch's probability of happening over the entire image. The final saliency map is built by normalizing and fusing local and global saliency maps of all channels from both color systems. (Patch rarity is  measured both in RGB and Lab color spaces.)

One can consider the work by~\citet{murray2011saliency}, as a faithful representative of the multiscale center-surround saliency methods. Its main idea is to:
 \begin{itemize}
\renewcommand{\labelitemi}{$\bullet$}
\itemsep 0em
\item apply a multi-scale multi-orientation wavelet pyramid to the image;
\item measure the local wavelet energy for each wavelet channel at each scale and orientation;
\item compute  a center-surround ratio for this energy; 
\item obtain in that way wavelet contrast coefficients that  have the same spatial  multi-scale sampling as the wavelet pyramid itself;
\item apply the reverse wavelet pyramid to the contrast coefficients to obtain a saliency map.
\end{itemize}
This is  a typical saliency-only  model, for  which an adequate detection  threshold is again  missing.

\paragraph{Conclusions.} Saliency detection methods learned from human gaze tracking are semantic methods that fall off our inquiry.  But  unsupervised saliency measures deliver a  map that  only needs to  be adequately thresholded to get an anomaly map. They therefore propose mechanisms and background structure assumptions that are relevant for anomaly detection.
Conversely, most anomaly detectors also deliver a saliency map before thresholding. 
The last three generic saliency measures listed are tantalizing. Indeed, they seem to do a very good job of enhancing anomalies by measuring rarity. Notwithstanding, they  come with no clear mechanism to transform the saliency map into a probabilistic one that  might allow  hypothesis testing and eventually statistically motivated detection  thresholds. 

\subsubsection{A sketch of our proposed classification}
The anomaly detection problem has been generally handled as a ``one-class'' classification problem. The 2003 very complete review by~\citet{markou2003noveltyA} concluded that most research on anomaly detection was driven by modeling background data distributions, to estimate the probability that test data do not belong to such distributions. Hence the mainstream methods can be classified by their approach to background modeling.
Every  detection  method has to do three things:  
\begin{enumerate}[label=(\alph*)]
\item to model the anomaly-free ``background''.  This background model may be constructed from samples of various sizes extracted from the given image (or an image database): pixels (e.g. in hyperspectral images), patches, local  features (e.g. wavelet  coefficients). 

\item to define a measure on the observed data evaluating how far its samples are from their  background model. Generally, this measure is a probability of false alarm (or even better, as we shall see, an expectation of the number of false alarms) associated with each sample.

\item to define the  adequate (empirically or statistically motivated) threshold value on the measure obtained in b).
\end{enumerate}
 The structure chosen for the background model appears to us as the most important difference between methods. Hence we shall  primarily classify the methods by the assumed structure of their  background model, and  the  way a distance  of samples to the background model is computed.  Section \ref{sec:generalization} will then be devoted to the computation  of the detection thresholds.

We shall examine in detail five generic structures for  the background: 
\begin{enumerate}
\item the  background can be  modeled by a \textit{ probability density function} (pdf), which is  either parametric, such as, a Gaussian, or a Gaussian mixture, or is obtained by interpolation from samples by a kernel density estimation method;  this structure leads to detect anomalies by hypothesis testing on the pdf;

\item the background is \textit{globally homogeneous} (allowing for a fixed reference image, a global Fourier or a convolutional neural network model generally followed by background subtraction);

\item the background is \textit{locally spatially homogeneous} (leading  to center-surround methods);

\item the background is \textit{sparse} on a given dictionary or base (leading to variational decomposition  models). 

\item the background is \textit{self-similar} (in the  non-local sense that for each sample there are other similar samples in  the  image).

\end{enumerate}

\section{Detailed analysis of the main anomaly detection families}
\label{sec:review}
The main anomaly detection families can be analyzed from their structural assumptions on the background model. In what follows we present and discuss the five  different families that we announced.

\subsection{Stochastic background models} 
\label{sec:review_parametric}
The principle of these anomaly detection methods is that anomalies occur in the low probability regions of the background model. The stochastic model can be parametric (Gaussian, Gaussian mixture, regression), or non-parametric.  For example in ``spectral anomaly detection'' as presented by \citet{chandola2009anomaly}, an anomaly is defined by having deviant coordinates with respect to normal PCA coordinates. This actually assumes a  Gaussian  model for the  background.

\paragraph{Gaussian background model.}
The Gaussian background assumption  may expand to image patches. \citet{du2011random} proposed to build a Gaussian background model from random  $2\times 2$ image  patches in a hyperspectral image. Once this background model $(\mu, \Sigma)$ is obtained, the  anomalous $(2\times 2)$ patches are detected using a threshold on their Mahalanobis distance to the background Gaussian model.  The selection of the image blocks permitting to estimate the Gaussian patch model $(\mu, \Sigma)$ is performed by a RANSAC procedure~\cite{fischler1987random}, picking random patches in the image and excluding progressively the anomalous ones.

\citet{goldman2004anomaly}, aiming  at  sea-mine detection, propose a detection scheme that does not rely on a statistical  model of the targets. It performs a background estimation in a local feature space of principal components (this again amounts to  building a Gaussian  model).  Then, hypothesis testing is used for the detection of anomalous pixels, namely those with an exceedingly high Mahalanobis distance to the Gaussian distribution (Section~\ref{sectionformalmodel}). This detects potentially anomalous pixels,  which are thereafter grouped and filtered by morphological operators.  This ulterior filter suggests that the first stage may yield many false alarms. 

\paragraph{Pdf estimation.}
Sonar images have a somewhat specific anisotropic structure that leads to model the background using signal processing  methods. For example, in \cite{mousazadeh2014two} the  authors proposed to adapt an ARCH model, thus obtaining a statistical detection model for anomalies  not explained by the non-causal model. This method is similar to the detection of scratches in musical records~\cite{oudre2015automatic}.

\citet{cohen1991automated} detect fabric defects using a Gaussian Markov Random Fields model. The method computes the likelihood of patches of size $32\times32$ or $64\times64$ according to the model learned on a database free of defects. The patches are then classified as anomalous or defect-free thanks to a likelihood ratio test.

\citet{tarassenko1995novelty} identify abnormal  masses in mammograms by assuming  that abnormalities are uniformly distributed outside the boundaries of normality (defined using an estimation of the probability density function from training data). 
If a feature vector falls in a low probability region (using a pre-determined threshold), then this feature vector is considered to be novel. The process to build the background model is complex and  involves selecting five local features, equalizing their means and variances to give them the  same importance, clustering the  data set into four  classes,  and estimating for each cluster its pdf by a non-parametric method (i.e.,  Parzen window  interpolation). Finally, a feature vector is considered anomalous if it has low  probability for each estimated pdf.  Such a non-parametric pdf estimate has of course an over-fitting or under-fitting risk, due to the fact that training data are limited. 

\paragraph{Gaussian Mixture.}
The  idea introduced  by~\citet{xie2007texems} is to learn a texture model based on Julezs' texton theory~\cite{julesz1981textons}. The textons are interpreted as image patches following a Gaussian model. Thus a random image patch is assumed to follow a Gaussian mixture model (GMM), which is therefore estimated from exemplar images by the expectation-maximization algorithm (EM). The method works at several scales in a Gaussian pyramid with fixed size patches (actually $5\times 5$). 
The threshold values for detecting anomalies are learned on a few images without defects in the following  way: At each scale, the minimum probability in the GMM over all patches is computed. These probabilities serve as detection thresholds. 
A  patch is then considered anomalous if its probability is lower than the minimum learned on the faultless textures on two consecutive dyadic scales in the  Gaussian pyramid.
A saliency map is obtained by summing up these consecutive probability excesses. 
Clearly, this model can  be transformed from a saliency map to an anomaly detector by using hypothesis testing on the background Gaussian mixture model. Gaussian mixture modeling has been long classical in hyperspectral imagery \cite{ashton1998detection} to detect anomalies. In that case, patches are not needed as each hyperspectral pixel already contains rich multidimensional information.

\paragraph{Gaussian Stationary process.}
\citet{grosjean2009contrario}~propose a method that models the background image as a Gaussian stationary process,  which can  also be  modeled as the result of the convolution of a white Gaussian noise model with an  arbitrary kernel, in other terms a colored  noise.
This background model is rather restrictive, but it is  precise and simple to estimate. The Gaussian model is first estimated. Then the image is filtered either with low-pass filters (to detect global peaks in the texture) or center-surround filters (to detect locally contrasted peaks in the texture). The Gaussian probability density function of each of these filtered images is easily computed. Finally, a probabilistic detection threshold for the filtered images is determined by bounding the NFA as sketched in Section~\ref{sectionformalmodel} (we shall give more details on this computation  in Section~\ref{sec:nfa_intro}.) 
\paragraph{Conclusions.}
To summarize, in the above methods relying on probabilistic background models, outliers are detected as incoherent with respect to a probability distribution estimated from the input image(s).  The anomaly detection threshold is a statistical likelihood test on the learned background model. In all cases, it gives (or could give) a p-value for each detection. So, by  tightening the detection thresholds, one can easily control the number of false alarms, as done by  \citet{grosjean2009contrario} (see Section~\ref{sectionformalmodel}).  
 
\subsection{Homogeneous background model} 
\label{sec:review_hom}
These methods \textit{estimate} and (generally) \textit{subtract} the background from the image to get a \textit{residual} image representation on  which detection is eventually performed. We shall examine different ways to  do so:  by using Fourier modeling, auto-encoder networks, or by subtraction of a smooth or fixed background.

\paragraph{Fourier background model.}
Perhaps the most successful background based method is the detection of anomalies in periodic patterns of textile~\cite{tsai1999automated,tsai2003automated,perng2010novel}. This can be done naturally by cutting specific frequencies in the Fourier domain and thresholding the residual to find the defects. 
For  example \citet{tsai1999automated} remove the background  by a frequency cutoff. Then a detection threshold using a combination of the mean and the variance of the residual yields a detection map.  

Similarly, \citet{tsai2003automated} propose an automatic inspection of defects in randomly textured surfaces which arise in sandpaper, castings, leather, and other industrial materials. The proposed method does not rely on local texture features, but on a background subtraction scheme in Fourier domain. It assumes that the spread of frequency components in the power spectrum space is isotropic, and with a shape that is close to a circle. By finding an adequate radius in the spectrum space, and setting to zero the frequency components outside the selected circle, the periodic, repetitive patterns of statistical textures are removed. In the restored image, the homogeneous regions in the original image get approximately flat, but the defective region is  preserved. According to the authors, this leads to convert the  defect detection in textures into a simple thresholding problem in non-textured images. This thresholding is done using a statistical process control (SPC) binarization method,
$$
f_b(x,y) =  \begin{cases} 
      255 & \text{ if } \mu - k\sigma \leqslant f(x,y) \leqslant \mu + k\sigma \\
      0 & \text{ otherwise,}  
   \end{cases}
$$
where $k$ is a control parameter, $\mu$ is the residual image average and $\sigma^2$ its variance. Regions set to zero are then detected.

\citet{perng2010novel} focus on anomaly detection during the production of bolts and nuts. The method starts by creating normalized unwrapped images of the pattern on which the detection is performed. The first step consists in removing the ``background'' by setting to zero some Fourier coefficients. Indeed, the background pattern being extremely periodic, is almost entirely removed  by canceling large Fourier coefficients. The mean $\mu$ and the variance $\sigma^2$ of the residual are then computed. This residual is then thresholded using the SPC binarization method of \citet{tsai2003automated}.

\citet{aiger2010phase} propose to learn a Gaussian background Fourier model of the image Fourier phase directly from the input image. The method assumes that only a few sparse defaults are  present in  the provided image.
First  a ``phase only transform (PHOT)'' is applied to the image. The Fourier transform of an image contains all the information of its source inside the modulus of the Fourier coefficients and their phase. The phase is known to contain key positional elements of the image, while the modulus relates more to the image  texture,  and therefore  to its background. To illustrate this fact, RPNs are  well known models for  a wide  class of ``microtextures'' as explained in \citet{galerne2011random}.  A RPN is a random image where the  Fourier coefficients have deterministic moduli (identical to the reference texture), but random, uniform, independent phases. Another illustration of  the role of phase and modulus is obtained noticing that a Gaussian noise has uniform random phase.   The PHOT amounts to invert the Fourier transform of an image after normalizing the Fourier coefficients modulus, thus keeping only the structural information contained in the phase.
A  local anomaly is expected to have a value in excess compared to the PHOT. Anomalous pixels are therefore detected as peaks of the Mahalanobis distance of their values to the background modelled as Gaussian distributed. Hence, a  probability of false alarm can be directly  computed in this ideal case.
The detection method  can be also applied after convolving the PHOT transformed image with a Gaussian, to detect blobs instead of single pixels.

\citet{xie2000golden} introduced a method to detect defects in periodic wafer images. By estimating the periods of the repeating pattern, the method obtains a ``golden template'' of the patterned wafer image under inspection. No other prior knowledge is required. The estimated defect-free background pattern image is then subtracted to find out possible defects.

\paragraph{Neural-network-based background model.} 
The general idea is to learn the background model by using a neural network trained on normal data.
Under the assumption that the background is homogeneous, the 
``replicator'' neural networks proposed by  \citet{hawkins2002outlier} can be used to learn this model. These networks were introduced in section \ref{sec:quick_review_reviews}.

Perhaps the most  important application of anomaly detection in industry is surface defect detection. \citet{iivarinen2000surface} proposes an efficient technique to detect defects in surface patterns. A statistical self-organizing map (SOM) is trained on defect-free data, using handpicked features from co-occurrence matrices and texture unit elements. The SOM is then able to separate the anomalies, which are supposed to have a different feature distribution.
As can be seen in \citet{xie2008review} which reviews surface defect detection techniques, many surface defect detection methods work similarly. Texture features are selected, and defects are detected as being not well explained by the feature model.

Similarly \citet{chang2009unsupervised} presented an unsupervised clustering-based  automatic wafer inspection system using  self-organizing neural networks. 
\citet{An2016} proposed to train a variational autoencoder (VAE), and to compute from it an average reconstruction probability, which is a different measure than just looking at the difference between the input and output. Given a new data point, a number of samples are drawn from the trained probabilistic encoder. For each code sample, the probabilistic decoder outputs the corresponding mean and variance parameters. Then, the probability of the original data being generated from a Gaussian distribution having these parameters is calculated. The average probability, named reconstruction probability, among all drawn samples is used as an anomaly score.

\citet{mishne2017diffusion} presented an encoder-decoder deep learning framework for manifold learning. The encoder is constrained to preserve the locality of the points, which improves the approximation power of the embedding. Outliers are detected based on the autoencoder reconstruction error. The work of~\citet{Schlegl2017} is in the same direction as using an autoencoder and looking at the norm between the original and the output. A Generative Adversarial Network (GAN)~\cite{goodfellow2014generative} is trained (generator + discriminator) by using anomalous-free data. Then, given a new test image a representation in latent space is computed (by backpropagation), and the GAN reconstruction is compared to the input. The discriminator cost is then used alongside the representation of the input by the network to find the anomalies. There is,  however, no guarantee that the latent representation found would do good for anomaly free examples.  Hence, it is not  clear why the discriminator cost would detect anomalies.

\paragraph{Smooth or fixed background model.}

Many surface defect detectors fall into that category. For example, a common procedure to detect defects in semiconductors is to use a fixed reference clean image and apply some detection procedure to the difference of the observed image and the reference pattern~\cite{hiroi1994precise,dom1995recent,tsai2005eigenvalue,shankar2005defect,tsai2005quantile}. Since for different chips, the probability of defects existing
at the same position is very low, one can extract a standard reference image by combining at least three images (by replacing  pixels located in defects by the pixels located
in the corresponding location of another image)\cite{liu2010defect}. Similar ideas have been exploited for the detection of defects in patterned fabrics~\cite{ngan2011automated}. In~\cite{ngan2005wavelet}, nonconforming regions are detected by subtracting a golden reference image and processed in the Wavelet domain.

A very recent and exemplary method to detect anomalies in smooth materials is the one proposed by~\citet{tout2016fully}. In this paper, the  authors develop a method for the fully automatic detection of anomalies on wheels surface. 
First, the wheel image are registered to a fixed position.  For each wheel patch in  a given position, a linear deterministic background model is designed. Its basis is made of a few low degree polynomials combined with a small number of basis functions  learned as the  first basis vectors of a PCA applied to exemplar data.
The acquisition noise is accurately modeled by a two-parameter Poisson noise. The parameters are easily estimated from the data.  The background estimation  is a mere projection  of each observed patch on the background subspace. The residual, computed as the difference between the input and the projection, can contain only noise-and anomalies.  Thus, classic hypothesis testing  on the norm of the residual of each  patch will yield an automatic detection  threshold.  This method is clearly adapted to defect detection on smooth surfaces.

\paragraph{Conclusions.} Homogeneous background model based anomaly detection methods are compelling detectors used in a wide variety of applications. They avoid proposing a stochastic model for an often complex background by computing the distance to the background or doing background subtraction. However this simplification comes at a cost: some algorithms are hard to generalize to new applications, and the detection decision mechanism is generally not statistically justified, with the exception of some methods, like \citet{tout2016fully}.

\subsection{Local homogeneity models: center-surround detection.} 
These methods are often used for creating saliency maps. Their rationale is that anomalies (or saliency) occur as local events contrasting with their surroundings.

In one  of the early papers on this topic, \citet{itti1998model} propose to compute a set of center-surround linear filters based on color, orientation and intensity. The filters are chosen to only have positive output values. The resultant maps are normalized by stretching  their response so that the max is at a pre-specified value.
These positive feature maps are then summed up to produce a final saliency map. Detection is then done on a simple winner-takes-all scheme on the maximum of the response maps. 
This method is applied in~\citet{itti2000saliency} to detect vehicles via their saliency in huge natural or urban images. It has also been generalized to video in~\citet{mahadevan2010anomaly}.

The method was expanded by~\citet{gao2008discriminant}. The features in this paper are basically the same as those proposed by~\citet{itti2000saliency}, that is, color features, intensity features, and a few orientation filters (Gabor functions, wavelets). This last paper does detection on image and video with center-surround saliency detector. It directly compares its  results to those of \citet{itti2000saliency} and takes similar features, but works differently with them. In particular it computes center-surround  discrimination scores for  the features, and puts in doubt the linearity of  center-surround filters and the need for computing a (necessarily nonlinear) probability of false alarm in  the  background  model. In fact, they claim~\cite{gao2008discriminant}:
\begin{quote}
``In particular, it is hypothesized that, in the absence of high-level goals, the most salient locations of the visual field are those that enable the discrimination between center and surround with smallest expected probability of error.'' 
\end{quote}

The difficulty of center-surround anomaly detection is faced by \citet{honda2001finding}, who introduced a generic method which tentatively works on all types of images. The main idea is to estimate a probability density for sub-regions in an image, conditioned upon the areas surrounding these sub-regions. The estimation method employs independent component analysis and the 
Karhunen-Lo\`eve transform (KLT) to reduce dimensionality and find a compact representation of the region space and its surroundings,  with elements  as  independent as possible. Anomaly is  again defined as a subregion with low  conditional probability with  respect to  its surrounding.  This is both a coarse grained and complex method.

\citet{scholkopf2000support} and \citet{tax2004support}~extended SVM to the problem of one-class detection (support estimation). The general idea is that by assuming that only a small fraction of the training data consist of anomalies, we can optimize the decision function of a classifier to predict if a point belongs or not to the normal class. The goal is to find the simplest or smallest region that is compatible to observing a given fraction of anomalies in the training set. 
In~\cite{gurram2012sparse}, the authors presented an ensemble-learning anomaly detection approach by optimizing an ensemble of kernel-based one-class classifiers.

Very recently, \citet{ruff2018deep} introduced a novel approach to detect anomalies using deep-learning that is inspired in the same ideas. The method, named Deep Support Vector Data Description (Deep SVDD), trains a deep neural network by minimizing the volume of a hypersphere that encloses the network representations of the data.

In the  famous Reed-Xiaoli (RX) algorithm  \cite{reed1990adaptive} the  pixels of a hyperspectral optical image are assumed  to follow a Gaussian non-stationary multivariate random process with a rapidly fluctuating space-varying mean vector and a more slowly space-varying covariance matrix. This  ``local normal model'' for  the  background pixels is learned from an outer window from which a guard window has been subtracted, as it might  contain the  anomaly. Then detection is performed by thresholding the  Mahanalobis distance  of the pixel of interest to the  local Gaussian model,  as described in Section~\ref{sectionformalmodel}. It may be noticed that a previous rough background subtraction  is  performed by a local demeaning using a sliding window \cite{margalit1985adaptive,chen1987detection}. \citet{matteoli2010tutorial} points out  two main limitations of the  RX method: first, the difficulty of estimating locally a high dimensional covariance matrix, and second the fact that \textit{a local anomaly is not necessarily a global anomaly:} an isolated tree in a meadow would  be viewed as an anomaly, even if its stands close to a wood of the same trees. Nevertheless, RX remains a leading  algorithm and it  has even online  versions: See, e.g.,~\cite{fowler2012anomaly} for the successful application of  RX after a dimensional reduction by random projections, inspired from compressed sensing.

 \paragraph{Conclusions.}  Most presented center-surround anomaly detectors produce a saliency map, but as previously mentioned in Section~\ref{notasaliencymeasure}, while saliency detectors are tantalizing since they propose simple and efficient rarity measurements, they provide no detection mechanism (threshold value). Several above reviewed center-surround methods attempt to remedy that.  But then, the method becomes quite heavy as it requires estimating a \textit{local stochastic model for both the center and  surround.} Hence  we  are forced back to two-class classification  with fewer samples and a far more complex methodology.

\subsection{Sparsity-based background models and  its variational implementations} \label{sec:review_sparsity}
One recent non-parametric trend is to learn a sparse dictionary representing the background (i.e., \emph{normality}) and to characterize outliers by their non-sparsity. 

\citet{margolin2013makes}~propose a method for building salient maps by a conjunction of pattern distinctness and color distinctness.  They claim that for pattern distinctness, patch sparsity is enough to characterize visual saliency.  They proceed by: 
\begin{enumerate}[label=(\alph*)]
\item Computing the PCA of all patches (of fixed size --typically $8\times 8$) in the image;
\item Computing the pattern saliency of a patch $p$ as $P(p):=\|p\|_1$ where the $l^1$ norm is computed on the PCA coordinates.
\item The pattern saliency measure is combined (by multiplication) with a color distinctness measure, that measures the distance of each color super pixel to its closest color cluster.   
The final map therefore is $D(p):=P(p)C(p)$ where $C(p)$ is  the color distinctness. 
\item  The final result is a product of this saliency map with (roughly) a Gaussian centered in the center of mass of the previous saliency map. 
\end{enumerate}

We now look at sparsity models that learn the background model as a dictionary on which ``normal'' patches would have to be represented by a sparse linear combination of the elements of the dictionary (and anomalous patches tentatively would not). Sparse dictionary learning, popularized by the K-SVD algorithm \cite{aharon2006k} and \cite{rubinstein2010dictionaries} and online learning methods \cite{mairal2009online}, has been successful for many signal representation applications and in particular for image representation and denoising \cite{mairal2009non}.

\citet{cong2011sparse,zhao2011online}~proposed a completely unsupervised sparse coding approach for detecting abnormal events in videos based on online sparse reconstructibility of query signals using a learned event dictionary. These methods are based on the principle that normal video events are more likely to be reconstructible from an event dictionary, whereas unusual events are not.

\citet{li2015low} introduced a low-rank and sparse tensor representation of hyperspectral imaginary HSI) data based on the observation that the HSI data volume often displays a low-rank structure due to significant correlations in the spectra of neighboring pixels.

The anomaly detector in hyperspectral images  proposed by \citet{li2015hyperspectral} soundly considers learning a background model and not an anomaly model. Its main contribution is perhaps to justify the use of sparsity to estimate a background model even in the presence of a minority of outliers. This detector belongs to the  class of center-surround detectors considered in the  previous section. In a neighbour of each pixel deprived of a ``guard'' central square, a sparse model of the background is learned by orthogonal matching pursuit. It is expected that the vectors of the  sparse basis will not contain any anomaly. Thus, the  projection of the  central pixel on the orthogonal space to this basis should have a norm much higher than the  average norm  observed in the surround if it is anomalous. The detection threshold is based on the ratio between these two  numbers and  is not further specified. It might nevertheless use a $\chi^2$ model, as the background residual could be modeled as white Gaussian noise.

For~\citet{boracchi2014novelty}, the background model is a learned patch dictionary from a database of anomaly-free data.  The abnormality of a patch is measured as the Mahalanobis distance to a 2D Gaussian learned on the parameter pairs composed by the $\ell_1$ norm of the coefficients and of their reconstruction error.  
In what follows we detail this method.

Although the method looks general, the initial question addressed by~\citet{boracchi2014novelty} is how to detect anomalies in complex homogeneous textures like microfibers. A model is built as a dictionary $\hat D$ learned from all patches $p_i$ by minimizing
$$
J_\lambda(X, D)=\| D X-P\|_F^2 +\lambda \|X\|_1,
$$
where $P$ is the matrix whose columns are the reference patches, the dictionary  $D$ is represented as a matrix where the columns are the elements of the dictionary,  $X$ is a matrix where the $i$-th column represents the coefficients of patch $p_i$ on $D$, and the data-fitting error is measured by the Frobenius norm of the first term.
The $\ell_1$ norm on $X$ must be understood as the sum of the absolute values of all of its coefficients. Once a minimizer $\hat D$ is obtained, the same functional can be used to find a sparse representation $\mathbf{x}$ for each patch $p$ by minimizing 
$$
J_\lambda(\mathbf{x})=\|\hat D \mathbf{x}-p\|^2 +\lambda \|\mathbf{x}\|_1.
$$

The question then arises: how to decide from this minimization that a patch $p$ is anomalous?  The authors propose to associate to each patch the pair of values
$\phi(p):=(\|\hat D \mathbf{x}-p\|, \|\mathbf{x}\|_1)$. 
The first component is a data-fidelity term measuring how well the patch is represented in $\hat D$. The second component measures the sparsity (and therefore the adequacy) of this representation. An empirical 2D Gaussian model $(\mu, \Sigma)$ is then estimated for these pairs calculated for all patches in the reference anomalous-free dataset. Under this Gaussian assumption, the \emph{normality} region can be defined for the patch model by fixing  an adequate  threshold $\gamma$ on the Mahanalobis distance of samples to this Gaussian model (see section \ref{sectionformalmodel}). According  to the authors fixing $\gamma$ is a ``suitable question'' that we shall address in Section \ref{sec:nfa_boracchi}.

 The~\citet{boracchi2014novelty}  method is directly related to the sparse texture modeling previously introduced by~\citet{elhamifar2012see}, where a ``row sparsity index'' is defined to distinguish outliers in a dataset. 
The outliers are added to the dictionary.  Hence, in any variational sparse decomposition of themselves, they will be used primarily as they cannot be sparsely decomposed over the inlier dictionary. In the words of the authors~\cite{elhamifar2012see}, 
\begin{quote}``We use the fact that outliers are often incoherent with
respect to the collection of the true data. Hence, an outlier
prefers to write itself as an affine combination of itself,
while true data points choose points among themselves as
representatives as they are more coherent with each other.''
\end{quote}
As we saw the ~\citet{boracchi2014novelty} method is  extremely well formalized. It was completed in~\citet{carrera2016scale} by adding a multiscale detection framework measuring the anomaly's non-sparsity  at several scales. The 2015  variant by~\citet{carrera2015detecting} of the above models introduces the tempting idea of building a convolutional sparse dictionary. This is done by minimizing
\begin{multline*}
L(\textbf{x}_m, d_m) = \\
\sum_{p\in \mathcal P} \left(\left\|\sum_{m=1}^M d_m\ast \mathbf{x}_m - p \right\|^2+\lambda \sum_{m=1}^M \|\mathbf{x}_m \|_1\right),
\end{multline*}
subject to $\|d_m\|_2=1$, $m=1,\cdots, M$, 
where $(d_m)_m$ and $(\mathbf{x}_m)_m$ denote a collection of $M$ filters and $M$ coefficient vectors respectively. As usual in such sparse dictionary models, the  minimization can be done on both the filters $(d_m)$ and  coordinates $x_m$ and summing for a learning set of patches.  Deprived of the sum over $p$, the  same functional can be minimized for a given input patch $p_0$ to compute its coordinates $\textbf{x}_m$ and evaluate its sparsity.

Defining anomaly detection as a variational problem, where anomalies are detected as non-sparse, is also the core of the method proposed by~\citet{adler2015sparse}. In a nutshell,  the $\ell_1$ norm of the coefficients on a learned background dictionary is used as an anomaly measure. More precisely, assuming a dictionary $D$ on which normal data would be sparse, the method performs the minimization
$$
\min_{X,E}\|Y-DX-E\|_F^2 + \alpha \|X\|_{1,q} + \beta\|E\|_{2,1},
$$
where $q=1$ for if sparsity is enforced separately on each sample and $q=2$ for enforcing joint sparsity of all samples and $\|E\|_{2,1}=\sum_{i}\|E(:,i)\|_2$ is the $l_{2,1}$ norm.
Here $Y$ is the data matrix where each column is a distinct data vector. Similarly $D$ is a matrix whose columns are the dictionary's components.  $X$ is the matrix of coefficients  of these data vectors on $D$ which  is forced by the $\|X\|_{1,q}$ term to become sparse.  Yet anomalies, which are not sparse on $D$, make a residual whose norm is measured as $\|E\|_{2,1}$, therefore their number should be moderated. Of course this functional depending on two parameters $(\alpha, \beta)$ raises the question of their adequate values. The final result is a decomposition $Y\simeq DX+E$ where the difference between $Y$ and $DX+E$ should be mainly noise and therefore we can write this
$$
Y=DX+E+N
$$
where  $N$ is the noisy residual, $DX$ the sparse part of $Y$ and $E$ its anomalies.

In appendix \ref{ap:dual_sparsity} we prove that the dual variational method amounts to finding directly the  anomalies. Furthermore,  we  have seen that  these methods cleverly solve the decision problem by applying  very simple hypothesis testing to the  low  dimensional variables formed by the values of  the terms of the functional. Hence, the method is generic,  applicable to all images  and can be completed by computing a number of false alarms, as we shall see.  Indeed, we interpret  the apparent over-detection by a neglect of  the  multiple testing. This can be fixed by the \textit{a-contrario} method and we shall do it in  Section~\ref{sec:nfa_boracchi}.

\paragraph{Dual interpretation of sparsity models.} 
Sparsity based variational methods lack the  direct interpretation  enjoyed by other methods  as  to the proper definition of an anomaly. By reviewing the first simplest method of this kind proposed by~\citet{boracchi2014novelty}, we shall see that its dual interpretation  points to the detection of the most deviant anomaly.
Let $D$ a dictionary representing ``normal'' patches. Given a new patch $p$ we compute the representation using the dictionary,
$$
\hat{x} = \argmin_x \left\{ \frac{1}{2}\|p-Dx\|_2^2 + \lambda \|x\|_1 \right\},
$$
and then build the ``normal'' component of the patch as $D\hat{x}$.

One can derive the following Lagrangian dual formulation (see Appendix~\ref{ap:dual_sparsity}),
\begin{equation} 
\label{dualsparsity}
\hat{\eta} = \argmin_\eta \left\{\frac{1}{2}\|p - \eta\|_2^2 + \lambda' \|D^T\eta\|_\infty\right\},
\end{equation}
where the vector $\eta$ are the Lagrangian multipliers.

While $D\hat{x}$ represents the ``normal'' part of the patch $p$, $\hat{\eta}$ represents the anomaly. Indeed, the condition $\|D^T\eta\|_\infty \leq \lambda$ imposes to $\eta$ to be far from the patches represented by $D$. 
Moreover, for a solution $\eta^*$ of the dual to exist (and so that the duality gap doesn't exist) it requires that $\eta^* = p - Dx^*$ \textit{i.e.} $p = Dx^* + \eta^*$ which confirms the previous observation. Notice that  the  solution of \eqref{dualsparsity}   exists by an obvious compactness argument and is unique by the strict convexity of  the dual functional.

\bigskip

\paragraph{Conclusions.}  The great advantage of the background models assuming sparsity is that they make a very  general structural assumption on the  background,  and  derive a variational model that  depends on one or two parameters only, namely the relative weights given to  the  terms  of the energy to be  minimized.

\subsection{Non-local self-similar background models} 
\label{sec:review_nonlocal} 

The non-local self-similarity principle is invoked as a qualitative regularity prior in many image  restoration  methods, and particularly for image denoising methods such as the bilateral filter~\cite{tomasi1998bilateral} or  non-local means~\cite{buades2005non}. It was first introduced for texture synthesis in the pioneering work of~\citet{efros1999texture}.

The basic assumption of this generic background model, applicable to most images, is that in normal data, each image patch belongs  to a dense cluster in the image's patch  space. 
Anomalies instead occur far from their closest neighbors.  This definition of  an anomaly can be implemented by clustering the image patches (anomalies being detected as far away from the centroid of their own cluster), or by a nearest neighbor search (NNS) leading to a  direct rarity measurement.

As several anomaly detectors derive from NL-means~\cite{buades2005non}, we shall here give a short overview of this image denoising algorithm. 
For each patch $p$ in the input image $u$, the $n$ most similar patches denoted by $p_i$ are searched and averaged to produce a self-similar estimate,
\begin{equation}
\hat{p} = \frac{1}{Z}\sum_{i=1}^{n} \exp \left(-\frac{ \|p - p_i\|_2^2 }{h^2}\right) p_i
\label{eq:nlmeans}
\end{equation}
where $Z=\sum_{i=1}^{n} \exp \left(-\frac{ \|p - p_i\|_2^2 }{h^2}\right)$ is a normalizing constant, and $h$ is a parameter (which should be set according to the noise estimation) and $\hat{p}$ is the denoised patch.
 
\paragraph{NL-means inspired model.}
An example of anomaly detector with non-local self-similar background model is \cite{seo2009static}, Seo and Milanfar propose to directly measure  rarity as an inverse function of resemblance. 
At each pixel $i$ a descriptor $F_i$ measures the likeness of a pixel (or voxel) to its surroundings. Then, this descriptor $F_i$ is compared to the corresponding descriptors of the pixels in a wider neighborhood. The saliency at a pixel $i$ is measured by 
\begin{equation}
S_i=\frac{1}{\sum_{j=1}^N \exp\left(\frac{\rho(F_i,F_j)-1}{h^2}\right)},
\label{saliencybydescriptorsimilarity}
\end{equation}
where $\rho(\cdot,\cdot)$ is the cosine distance between two descriptors, $F_i$ is the local feature, and $F_j$ for $j=1,\ldots,N$, the $N$ closest features to $F_i$ in the surrounding, and $0<h<1$ is a parameter. 

The formula reads as follows: if all $F_j$ are not aligned to $F_i$, the exponentials in~\eqref{saliencybydescriptorsimilarity} will be all small and therefore the saliency will be high. If instead only one $F_j$ correlates well with $F_i$, the saliency will be close to one, and if $k$ different  $F_j$s correlate well with $F_i$, $S_i$ will be approximately equal to $\frac 1 k$.  This method cannot yield better than a saliency measure, as no clear way of having a detection mechanism  emerges:  how do we set a detection threshold?
 
The algorithm in~\citet{zontak2010defect} is closely inspired   from NL-means: 
For a reference patch $p$, a similarity parameter $h^2$ 
and a set of $n$ neighboring patches $(p_i)$, an anomaly is detected when 
$$
\sum_{j=1}^n e^{\frac{-\|p-p_j\|_2^2}{h^2}} \leqslant \tau
$$
where $\tau$ is an empirical  parameter. 
The anomaly detection is applied to strongly self-similar wafers, and the authors also display  the difference between their actual denoised source image by the NL-means denoising  algorithm, and an equally denoised reference image.
We can interpret the  displayed experiments, if not the method, as a form of background subtraction followed  by a detection threshold on the  \textit{residual}.  In section \ref{zontakcohenNFA} we  shall propose a statistical method for fixing $\tau$.

A similar idea was proposed by~\citet{tax1998outlier}: 
\begin{quote}
``The distance of the new object and its nearest neighbor in the training set is found and the distance of this nearest neighbor and its nearest neighbor in the training set is also found. The quotient between the first and the second distance is taken as indication of the novelty of the object.’’
\end{quote}
 
As demonstrated more recently by the SIFT method~\cite{lowe1999object}  this  ratio is a  powerful tool. In SIFT a descriptor in a first  image is  compared to all other descriptors in a target image.  If the ratio of distances between the closest descriptor and  the  second closest one is  below a certain threshold,  the match between both descriptors is considered meaningful.  Otherwise,  it is considered casual. 

In \citet{davy2018reducing} the  authors of  the present review addressed this last step. They proposed to perform background modeling on the \textit{residual image} obtained by background subtraction.  As for the above mentioned self-similarity based methods, the background is assumed self-similar. Thus, to remove it, a variant of  the NL-means algorithm is applied. 
The background modeling consists in replacing each image patch by an average of the most similar ones. These similar patches are found outside a ``guard region'' centered at the query patch. This precaution  is  taken to prevent anomalies with some self-similar structure to be kept  in the background.  

Equation~\eqref{eq:nlmeans} used to reconstruct the background is the same as for NL-means.
Since each pixel belongs to several different patches, it  receives several distinct estimates that can be averaged to give the  final background image $\hat{u}$. Finally, the residual image is built as $r(u)=\hat{u}-u$. 
Anomalies, having no similarities in the image, should remain in the residual $r(u)$. In the absence of the anomalies, the residual should instead be unstructured and  therefore akin to a noise.  
Then, the method uses the~\citet{grosjean2009contrario}  \textit{a-contrario} method to detect fine scale anomalies on the residual. A pyramid of images is used to detect anomalies at all scales. The method is shown to deliver similar results when producing the residual from features obtained from convolutional neural networks instead of the raw RGB features (see~\cite{davy2018reducing}).
Still, there is something  unsatisfactory in the method: it assumes  like \citet{grosjean2009contrario} that the background is an uniform Gaussian random field, but no evidence is given that the residual would obey such a model.

\citet{boracchi2014exploiting} proposed to detect structural changes in time-series by exploiting the self-similarity. Their general idea is that a normal patch should have at least one very similar patch along the sequence. Given a temporal patch (a small temporal window) the residual with respect to the most similar patch in the sequence is computed. This leads to a new residual sequence (i.e., change indicator sequence). The final step is to apply a traditional change detector test (CDT) on the residual sequence. CDTs are statistical tests to detect structural changes in sequences, that is, when the monitored data no longer conform to the independent and identical distributed initial model. CDTs run in an online and sequential fashion.
The very recent method~\cite{napoletano2018anomaly} is similar to the above commented \cite{boracchi2014exploiting}.  Its main  difference  is the usage of convolutional neural network features instead of image patches.

\paragraph{Kernel PCA background model.}
Manifold and PCA kernel methods reduce the computational expense by a uniform random sampling of a small fraction of the data, which has high chance of being uncontaminated by anomalies.
The kernel PCA method for anomaly detection introduced by~\citet{hoffmann2007kernel} defines a Gaussian kernel on the  dataset $x_i$, $i=1,\ldots, M$ by setting $k(x_i,x_j)=e^{-\frac 1{h^2}||x_i-x_j||^2}$,  $i,j=1,\ldots, M$.   
This ``kernel'' is actually assumed to represent the actual scalar product between feature vectors of the samples $\Phi(x_i)$ and $\Phi(x_j)$ in a high-dimensional feature space ($\Phi$ being implicitly defined). The trick of kernel PCA consists in performing implicitly a PCA in this feature space with computations only involving $k$. It is possible to compute the distance between $\Phi(z)$ and $\Phi_0 = \sum_{i=1}^M \Phi(x_i)$ using only $k$:
$$p(z)=k(z,z)-\frac 2M\sum_{i=1}^Mk(z,x_i)+\frac 1{M^2}\sum_{i,j=1}^M k(x_i,x_j).$$
Since the first term is 1 and the last term constant, it  follows that
$$p(z)=C-\frac 2M\sum_{i=1}^Mk(z,x_i),$$ which is opposite to the Parzen density estimation of the sample set using a Gaussian  kernel with standard deviation $h$. Thus, anomalies will be detected by setting a threshold on this density computed from the background samples. A more complete background subtraction can be performed by subtracting its $q$ first PCA components.
  
\paragraph{Diffusion map background model~\cite{olson2018manifold}.}
The diffusion map construction~\cite{coifman2006diffusion} views the data as a graph  where a kernel  function  $k(x_i,x_j)$ measures vertex similarity. Like in kernel PCA, consider the matrix $K_{ij} = e^{-\frac 1{h^2}||x_i-x_j||^2}$  associated with a Gaussian kernel, and transform it into a probability matrix by setting $p_{ij}=\frac{K_{ij}}{\sum_j K_{ij}}$. 
This matrix is interpreted as the probability that a random walker will jump from $x_i$ to $x_j$. The probability for  a random walk in the graph  moving from $x_i$ to $x_j$ in $t$  time steps is  given by $(P^t)_{ij}$, where  $P=(p_{ij})_{ij}$.  
The eigenvalues $\lambda_k$, and eigenvectors $\alpha^k$ of the $t$-th transition matrix
provide  diffusion map coordinates. Using these coordinates one can easily computes the distance (called \textit{diffusion distance}) between two graph nodes.
A background manifold is learned from these samples.  Unsampled data are the projected on a local plane tangent to the  manifold. The projection error can be then used as an anomaly detection statistic. 
The distance of a new sample $\theta'$ from the manifold is approximated by selecting a subset of $k$ nearest neighbors on the  manifold, finding the best least-squares plane through those points, and approximating the distance of the new point from the plane.
An adequately threshold on this distance is all that is needed to detect anomalies.
We refer to~\cite{madar2011non} for an actually very complex anomaly detector  based on a diffusion map of an image's hyperspectral pixels.

More recently, the self-similarity measurement proposed by \citet{goferman2012context},  finds for each $7\times 7$ patch $p_i$ its $K=64$ most similar patches $q_k$ in a spatial neighborhood, and computes its saliency as 
\begin{equation}
S_i= 1-\exp{\left(-\frac 1 {K} \sum_{k=1}^{K} d(p_i, q_k)\right)}.
\label{eq:Si_general}
\end{equation}

The distance between  patches is a combination of Euclidean distance of color maps in LAB coordinates and of the Euclidean distances of patch positions,
\begin{equation}
d(p_i, p_j)= \frac{\|p_i-p_j\|}{1+3 \|i-j\|},
\label{Gofermandistance}
\end{equation} 
where the norm is the Euclidean distance between patch color vectors or between patch positions $p_i, p_j$.

The algorithm computes saliency measures at four different scales and then averages them to produce the final patch saliency.
This is a rough measure: all the images are scaled to the same size of 250 pixels (largest dimension) and take patches of  size $7\times7$.   The four scales are 100\%, 80\%, 50\% and 30\%.
A pixel is considered salient if its saliency value exceeds a certain threshold ($S= 0.8$ in the examples shown in the paper).

The patch distance~\eqref{Gofermandistance} used in~\citet{goferman2012context} is almost identical to the descriptor distance proposed by \citet{mishne2014multiscale}. %
Like in their previous paper \citet{mishne2013multiscale}, the authors perform first a dimension reduction of the patches. To that aim  a nearest neighbor graph on the set of  patches is built, where the weights  on the edges between patches are decreasing functions of their Euclidean distances, $w(p_i,p_j)=\exp\left(-\frac{\|p_i-p_j\|^2}{h^2}\right)$. These positive weights allow  to define a graph Laplacian.  Then the basis of eigenvectors of the Laplacian is computed. The  first coordinates of each patch on this  basis yield a low dimensional embedding of the  patch space. (There is an equivalence between this  representation of patches  and the application to the patches of the NL-means algorithm, as pointed out in~\cite{singer2009diffusion}.)

The anomaly score involves the distance of  each patch to the first $K$ nearest neighbors, using the new patch coordinates $\tilde p_i$. This yields the following anomaly score for a given patch $p_i$ with coordinates  $\tilde p_i$:
$$
S_i=1-\exp\left(-\frac 1 K\sum_{k=1}^K \frac{\|p_i-p_j\|/2h}{1+c\|\tilde p_i- \tilde p_j\|}\right).
$$
Note  the intentional similarity of this formula with~\eqref{eq:Si_general} and \eqref{Gofermandistance}. Mishne and Cohen indeed state that they are adapting the Goferman score to the embedding space. Similar methods have been developed for video~\citet{boiman2007detecting}.

All of the mentioned methods so far have no clear specification of their anomaly threshold. This comes from the  fact that  the self-similarity principle is merely qualitative.  It  does not  fix a rule to  decide if  two patches are alike or not.

 \paragraph{Conclusions on  self-similarity.}
  Like sparsity, self-similarity is a powerful qualitative model, but we have pointed out that in all of  its applications except one, it  lacks a rigorous mechanism to fix an anomaly detection threshold.   The only exception is  \cite{davy2018reducing}, extending the \citet{grosjean2009contrario} method and therefore obtaining a rigorous detection threshold under the assumption that the residual image is a Gaussian random field. 
 The fact that  the residual is  more akin to a random noise than the background image is believable, but not  formalized.

 \subsection{Conclusions, selection of the  methods, and their  synthesis}
 \label{sec:review_conclusion}

\begin{table*}[h]
  \centering
  \begin{tabular}{l l l}
  \toprule Background category & Background sub-category & Reviewed methods \\\midrule
  \multirow{4}{*}{Stochastic} & Gaussian & \cite{du2011random, goldman2004anomaly} \\
   & Non-parametric pdf & \cite{mousazadeh2014two, cohen1991automated, tarassenko1995novelty} \\
   & Gaussian mixture & \cite{xie2007texems, ashton1998detection} \\
   & Gaussian stationary process & \cite{grosjean2009contrario} \\
   \midrule
  \multirow{3}{*}{Homogeneous} & Fourier & \cite{tsai1999automated,tsai2003automated,perng2010novel,aiger2010phase,xie2000golden} \\
   & Neural-network & \cite{hawkins2002outlier, iivarinen2000surface, chang2009unsupervised, An2016, mishne2017diffusion, Schlegl2017} \\
   & Smooth/fixed & \cite{hiroi1994precise, dom1995recent, tsai2005eigenvalue, shankar2005defect, tsai2005quantile, liu2010defect, ngan2005wavelet, tout2016fully} \\
   \midrule
  Locally Homogeneous &  & \cite{itti1998model, itti2000saliency, gao2008discriminant, honda2001finding, mahadevan2010anomaly, gurram2012sparse, ruff2018deep, reed1990adaptive, margalit1985adaptive, chen1987detection, fowler2012anomaly} \\
  \midrule
  Sparsity based &   & \cite{margolin2013makes, cong2011sparse, zhao2011online, li2015low, li2015hyperspectral, boracchi2014novelty, elhamifar2012see, carrera2015detecting, carrera2016scale, adler2015sparse} \\
  \midrule
  \multirow{3}{*}{Non-local self-similar} &  NL-means inspired & \cite{seo2009static, zontak2010defect, tax1998outlier, davy2018reducing, boracchi2014exploiting, napoletano2018anomaly} \\
   &  Kernel PCA & \cite{hoffmann2007kernel} \\
   &  Diffusion maps & \cite{olson2018manifold, coifman2006diffusion, madar2011non, goferman2012context, mishne2013multiscale, mishne2014multiscale, boiman2007detecting} \\\bottomrule
  \end{tabular}

  \caption{Synopsis of the examined anomaly detectors.}
  \label{tab:rel_work} 
\end{table*}

Table~\ref{tab:rel_work} recapitulates the analyzed papers in Section \ref{sec:review}. We observed that the methods giving a stochastic background model are powerful when the images belong to a restricted class of homogeneous objects, like textiles or  smooth painted  surfaces.  Indeed, the method furnishes rigorous detection thresholds based on the estimated probability density function.  But, regrettably, stochastic background modeling is not applicable on generic images. For the same reason, homogeneous background models are restrictive and do not rely on provable detection thresholds. We saw that center-surround methods are successful for saliency enhancement, but generally again lack a detection mechanism.  We also saw that the center-surround methods  proposing a detection  threshold have to estimate two stochastic models,  one for the center and one for the  surround,  being therefore quite complex and coarse grained.
The last two categories, namely the \textit{sparsity} and the  \textit{self-similarity} models are tempting and thriving. Their big advantage is their universality: they can be applied to all background images,  homogeneous or not, stochastic or not. But again, the  self-similarity model lacks a rigorous detection mechanism, because it works on a feature  space that is not easily modeled. 
Nevertheless,  several sparsity  models that we examined do propose a hypothesis testing  method based on  a pair of  parameters derived from the  variational method.  But these parameters have no justifiable  model and anyway do not take  into account  the multiple testing.  This last objection can be  fixed though, by computing  a number of false alarms as  proposed in \cite{grosjean2009contrario}, and  we shall do it in the  next section.   

As pointed out in \citet{davy2018reducing}, abandoning the goal of building a stochastic background model  does not imply abandoning the idea of a well-founded probabilistic threshold.
Their work hints that background subtraction is a powerful way to get rid of the hard constraint to model background and to work only on the residual.
But in \cite{davy2018reducing} no final argument is given demonstrating that the residual can be modeled as a simple noise. Nevertheless, this paper shows that the parametric \citet{grosjean2009contrario} detection works better on the residual than on the original image (see Section~\ref{GrosjeanandDavyNFA}).

We noticed that at least one paper (\citet{aiger2010phase}) has  proposed a form of  background whitening.  It  seems therefore advisable to improve background subtracting methods by applying  the PHOT to the  residual.  This post-processing step will remove the potential background leftovers of the NL-means inspired background subtracting method, and thus slightly enhance the detection results.
  
 Our conclusion is that we might be closer to a  fully generic anomaly detection by combining the  best advances that  we have  listed.   To summarize we see two different combinations of  these advances that might  give a competitive result:
 \begin{enumerate}
 \item The  sparsity method joined by an \textit{a-contrario} decision:
 \begin{itemize}
 \item model the background by a sparse dictionary~\cite{carrera2016scale};
 \item estimate a Gaussian on the distance parameters (these are actually statistics on the residual) \cite{carrera2015detecting};
 \item apply the \textit{a-contrario} detection  framework on this estimated Gaussian to control the  NFA \cite{desolneux2007gestalt}.
 
 \end{itemize}
 \item Background subtraction by self-similarity and residual whitening
 \begin{itemize}
 \item apply a variant of NL-means (using patches from the whole image) excluding a local search region to define the background;
 \item obtain the residual by subtracting the background \cite{davy2018reducing};
 \item whiten the residual by the phase only transform (PHOT) \cite{aiger2010phase};
 \item apply the \citet{grosjean2009contrario} center-surround detection criteria to the whitened residual.
 \end{itemize}
 \end{enumerate}
 These two proposals have the  advantage of taking  into account all the advances in anomaly detection that we pointed out. They cannot be united; sparsity and self-similarity are akin but different  regularity models.
 We notice  that  both methods actually work on a  residual. In the  second proposed  method the residual is computed explicitly. In the first one, the decision  method is taken on a Gaussian model for a pair of parameters where one  is actually the norm of the  residual and the other one a sparsity measure.
In   Section~\ref{sec:generalization}
 we  develop the  tools necessary to compare the selected methods. We need a unified anomaly detection criterion, and we shall see that the \textit{a-contrario} framework, introduced in Section~\ref{sec:nfa_intro}, gives one.

\section{Estimating a number of false alarms for all compared methods}
\label{sec:generalization}
In Section~\ref{sec:review}, we classified anomaly detection methods into several   families  based on their background models: stochastic, homogeneous, local homogeneous, sparsity-based and non-local self-similar models. 
Our final goal is to compare the results of these  families by selecting state of the art representatives for each family. 

All methods presented in section \ref{sec:review} require  a detection threshold.  
These thresholds are not always explicit and remain empirical in  many papers: instead of a universal threshold, most  methods propose a range from which to choose depending on the application or even on the image. 

To perform a fair comparison of  the  selected methods, we must  automatically set their detection threshold,  based on an uniform criterion. This will done by computing for  each method a Number of False Alarms,  using the \textit{a-contrario} framework introduced by \citet{desolneux2004}, \cite{desolneux2007gestalt}.  This detection criterion  is already  used in  two of  the examined papers, \cite{grosjean2009contrario} and \cite{davy2018reducing}.  We  give in  the next section a general framework to the explanations given in  section \ref{sectionformalmodel} on the particular example of the Mahanalobis distance.

\subsection{Computing  a number of false alarms in the \textit{a-contrario} framework}
\label{sec:nfa_intro}

The \emph{a-contrario} framework  is classical in many  detection or estimation computer vision tasks, such as line segment detection~\cite{grompone2010lsd,ipol.2012.gjmr-lsd}, ellipse detection~\cite{patraucean2012parameterless},  spot detection \cite{grosjean2009contrario}, vanishing points detection~\cite{lezama2014finding,ipol.2017.148}, fundamental matrix estimation \cite{moisan2004probabilistic}, image registration \cite{ipol.2012.mmm-oh},  mirror-symmetry detection~\cite{patraucean2013detection}, cloud detection~\cite{dagobert:tel-01691611}, among others. 

The a-contrario framework is a general methodology to automatically fix a detection threshold in terms of hypothesis testing. This is done by linking the number of false alarms (NFA) and the probability of false alarm, typically used in hypothesis testing. It relies on the following simple definition.
\begin{definition} \label{defnfa}\cite{grosjean2009contrario} \textit{
Given a set of random variables $(X_i)_{i \in [1,N]}$ with known distribution under a null-hypothesis $(\mathcal{H}_0)$, a test function $f$ is called an NFA if it guarantees a bound on the expectation of its number of false alarms under $(\mathcal{H}_0)$, namely:
$$
\forall{\varepsilon>0}, \mathbb{E}[\#\{i, f(i, X_i) \le \varepsilon\}] \le \varepsilon.
$$}
\end{definition}
To put it in words, raising a detection every time the test function is below $\varepsilon$ should give under $(\mathcal{H}_0)$ an expectation of less than $\varepsilon$ false alarms. An observation $\mathbf{x}_i$ is said to be ``$\varepsilon$-meaningful'' if it satisfies $f(i, \mathbf{x_i})\leq \varepsilon$, where $\varepsilon$ is the predefined target for the expected number of false alarms. The lower $f(i, \mathbf{x})$ the  ``stronger'' the detection.

Notice  that the  function $f(i,X_i)$ is called an NFA function but we call also its value for  a given sample an NFA. Thus we can use expressions like ``the NFA of $X_i$ is  lower than $\varepsilon$''.

While the definition of the background model $(\mathcal{H}_0)$ doesn't contain any \textit{a priori} information on what should be detected, the design of the test function $f$ reflects expectations on what is an anomaly. 
A common way to build an NFA is to take
\begin{equation}f(i, \mathbf{x_i}) = N \mathbb{P}_{\mathcal H_0}(X_i \ge \mathbf{x}_i)\label{eq:nfa_simple}\end{equation}
or
\begin{equation}f(i, \mathbf{x_i}) = N \mathbb{P}_{\mathcal H_0}(|X_i| \ge |\mathbf{x}_i|) ,\label{eq:nfa}\end{equation}
where $N$ is the number of tests, $i$ goes over all tests,  and $\mathbf{x}_i$, are the observations which excess should raise an alarm.  These test functions are typically used when anomalies are expected to have higher values than the background in the first case, or when anomalies are expected to have higher modulus than the background.  
If for example the $(X_i)$ represent the pixels of an image, there would be one test per pixel and per channel. Hence $N$ would  be the product  of the image dimension by the number of image channels.

\citet{grosjean2009contrario} proved that the test function~\eqref{eq:nfa_simple} satisfies  Definition \ref{defnfa}. Since the only requirement of their proof is that $X_i$ has to be a real-valued random variable, a more general result can be derived for any function $g$ and multi-dimensional $X_i$ if $g(X_i)$ is a real-valued random variable. Under these conditions, the following function
\begin{equation}
f(i, \mathbf{x}) = N \mathbb{P}_{\mathcal H_0}(g(X_i) \ge g(\mathbf{x}_i))
\label{eq:nfa_complex}
\end{equation}
also is a NFA.

In  short, applying the \textit{a-contrario} framework just requires a stochastic background model $(\mathcal{H}_0)$ giving  the laws of the random variables $X_i$, and a test function $f$. 

In~\citet{davy2018reducing} for example,  $X_i$ denote the pixels of the residual image $r(u)$, that presumably follow a Gaussian colored noise model. This Gaussian model defines the null hypothesis $(\mathcal{H}_0$), and $N$ is the total number of tested pixels (considering all the scales and channels), and the test function is given by~\eqref{eq:nfa}.

\begin{proposition}
Consider the simplest case where all tested variables  are equally distributed under $(\mathcal{H}_0)$, and assume that their cumulative distribution function is invertible.  Assume  that the  test function is given by~\eqref{eq:nfa}.  Then testing  if  $|\mathbf{x}_i|$ is above  $\gamma_\varepsilon$ defined by
\begin{equation}\mathbb{P}(|X| \ge \gamma_\varepsilon) = \frac{\varepsilon}{N}
\label{eq:nfa_threshold_simple}
\end{equation}
ensures a number of  false alarms lower that  $\varepsilon$.
\end{proposition}
In the particular \emph{a-contrario} setting given by Eq.~\eqref{eq:nfa_threshold_simple}, the number of false alarms gives a  result similar to the Bonferroni correction~\cite{bland1995multiple}, used to compensate for multiple testing. It is also interpretable as a per family error rate \cite{hochberg1987multiple}. Deeper results can be found in \cite{desolneux2007gestalt}.

In the  next sections we specify the \textit{a-contrario} framework for the  methods that we will be comparing.
\subsection{The \citet{grosjean2009contrario} stochastic parametric background model  and the \citet{davy2018reducing} self-similarity model \label{GrosjeanandDavyNFA}}

\citet{grosjean2009contrario} proposed to model the input image as a colored Gaussian stationary process. The method is designed to detect bright local spots in textured images, for example, mammograms. Three different ways to compute a NFA are proposed by locally assuming (i) no context, (ii) contrast related to the context, and (iii) a conditional context. Method (i) comes down to convolving the image with disk kernels, and testing the tails of the obtained Gaussian distributions, while method (ii) comes down to convolving with center-surround kernels. Their second method is preferred since with strong noise correlation the local average in their background model can be far from 0. 

In \citet{davy2018reducing}, a residual image is produced with a self-similarity removal step, which contains a normalization step to make the noise more Gaussian. The residual is then supposed to behave as  colored Gaussian noise. Then the method comes down to convolving the residual with disk kernels, and testing the tails of the obtained Gaussian distributions.

Both methods do combine the detection at several scales of the input image.
Thus, both methods share a similar detection mechanism and can be expressed in the same terms.
Under their $(\mathcal{H}_0)$, the result of the convolutions of the image for the former, and of the residual for the latter, with the testing kernels are colored Gaussian noise which mean and variance can be  estimated accurately from the filtered  image itself. Hence, the NFA test function applied on all the residual values (pixel/channel/residual) is exactly the function (\ref{eq:nfa}). Both methods assume the anomaly impact on the variance estimation is negligible (small anomaly).

\subsection{The Fourier homogeneous background model of~\citet{aiger2010phase}}
\label{sec:nfa_aiger}

In the \citet{aiger2010phase} method, a residual is obtained by setting the value of the modulus of the Fourier coefficients of the image (PHOT) to 1. The residual is then modeled \textit{a-contrario} as a simple Gaussian white noise whose mean and variance are estimated from the image.  
Anomalous pixels are therefore detected by using a threshold on the Mahalanobis distance between the pixel value and the background Gaussian model.
Let $(\mathcal{H}_0)$ be the null hypothesis under which the residual values $(X_i)$ follow a Gaussian distribution with mean $\mu$ and variance $\sigma^2$. Then we  have
\begin{align}
\mathbb{P}\left(\left|\frac{X_i - \mu}{\sigma}\right| \geqslant \gamma_\varepsilon\right) &=
 2 \int_{\gamma_\varepsilon}^{\infty} \frac{e^{-\frac{u^2}{2}}}{\sqrt{2\pi}}du\\
&= \erfc\left(\frac{\gamma_\varepsilon}{\sqrt{2}}\right).
\end{align}
Thus, the associated function
$$
f(i, \mathbf{x}_i) = N\mathbb{P}\left(\left|\frac{X_i - \mu}{\sigma}\right| \geqslant \left|\frac{\mathbf{x}_i - \mu}{\sigma}\right|\right)
$$
 is an NFA of  the form  (\ref{eq:nfa_complex}),
where the number of tests $N$ corresponds to the number of pixels in the image.  This  NFA leads to  detect an anomalous pixel when $\left|\frac{\mathbf{x}_i - \mu}{\sigma}\right|$ is above $\gamma_\varepsilon$ verifying
$$
\gamma_\varepsilon := \sqrt{2} \erfc^{-1}\left(\frac{\varepsilon}{N}\right).
$$
The impact  of anomalies impact on the PHOT is assumed to be negligible,  which implicitly assumes small or low intensity anomalies with respect to the background.

\subsection{The \citet{zontak2010defect} non-local self-similar model\label{zontakcohenNFA}}

In this method, the detection test is based on the NL-means weights. If the sum of these weights is smaller than a threshold $\tau$ (before normalization of these weights), then it is considered an anomaly. In what follows, we discuss how to choose this threshold $\tau$ by computing a NFA. 
We restrict ourselves to the case where the distance between patches is the $\ell_2$ distance.

Let us recall that for a reference patch $p$, a similarity parameter $h^2$ 
and a set of $n$ neighboring patches $(p_i)$, an anomaly is detected when 
\begin{equation}
\sum_{j=1}^n e^{\frac{-\|p-p_j\|_2^2}{h^2}} \leqslant \tau.
\end{equation}

Under $(\mathcal{H}_0)$, every patch $X_i$ of the image is associated with $n$ spatially close patches $P_{i,j}$. At least one of these patches is similar and only differs by the realization of the noise, the noise-free content assumed to be identical. The noise is supposed to be for each pixel an independent centered Gaussian noise of variance $\sigma^2$.
We know that 
\begin{equation}
\label{eq:zontak_g}
f(i, \mathbf{x}) = N\mathbb{P}\left(\sum_{j=1}^n e^{\frac{-\|X_i-P_{i,j}\|_2^2}{h^2}} \leqslant \sum_{j=1}^n e^{\frac{-\|\mathbf{x}_i-p_{i,j}\|_2^2}{h^2}}\right),
\end{equation}
verifies the NFA property (this is just equation (\ref{eq:nfa_complex}) with a well chosen $g$).

By hypothesis, at least one of the $P_{i,j}$ - we shall name $P_{i}^*$ one of these patches - is a realization of the same content than $X_i$ but with different noise (that we suppose to be of standard deviation $\sigma$).

By event inclusion,
$$
\mathbb{P}\left(\sum_{j=1}^n e^{\frac{-\|X_i-P_{i,j}\|_2^2}{h^2}} \leqslant \tau\right) \leqslant \mathbb{P}\left(e^{\frac{-\|X_i-P_{i}^*\|_2^2}{h^2}} \leqslant \tau\right).
$$

Moreover
\begin{multline*}
\mathbb{P}\left(e^{\frac{-\|X_i-P_{i}^*\|_2^2}{h^2}} \leqslant \tau\right) = \mathbb{P}\left(\frac{\|X_i-P_{i}^*\|_2^2}{h^2} \geqslant -\log(\tau)\right)\\
= 1 - \mathbb{P}\left(\frac{\|X_i-P_{i}^*\|_2^2}{2\sigma^2} \leqslant -\frac{h^2}{2\sigma^2}\log(\tau)\right).
\end{multline*}
Here we suppose that the candidate is indeed the same as the patch modulo the noise. Therefore the distance follows a $\chi^2$ law of degree the size of the patch.

That is,
$$
\mathbb{P}\left(e^{\frac{-\|X_i-P_{i}^*\|_2^2}{h^2}} \leqslant \tau\right) = 1-\text{chi2}\left(-\frac{h^2}{2\sigma^2}\log(\tau)\right)
$$
where \text{chi2} is the cumulative density function of the $\chi^2$ distribution of the degree the size of the patch.

Thus, by bounding (\ref{eq:zontak_g}) from above, and using the fact that a function whose value is always above a NFA  is also a NFA  (there will be fewer or an equal number of detections), the following test function also is a NFA:
\begin{equation*}
\resizebox{.99 \columnwidth}{!} 
{
$f(i, \mathbf{x}) = N\left(1-\text{chi2}\left(-\frac{h^2}{2\sigma^2} \log\left(\displaystyle\sum_{j=1}^n e^{\frac{-\|\mathbf{x}_i-p_{i,j}\|_2^2}{h^2}}\right)\right)\right)$
}
\end{equation*}
Thus, by definition of a NFA, a detection is raised if
$$
f(i, \mathbf{x}) \leqslant \varepsilon,
$$
which leads to a threshold $\tau_\varepsilon$ on  $\sum_{j=1}^n e^{\frac{-\|\mathbf{x}_i-p_{i,j}\|_2^2}{h^2}}$ satisfying
$$
\tau_\varepsilon := \exp\left(-\frac{2\sigma^2}{h^2}\text{chi2inv}\left(1-\frac{\varepsilon}{N}\right)\right).
$$

In order to fit the $(\mathcal{H}_0)$ hypothesis we can estimate $\sigma^2$ using \citet{ponomarenko2007automatic} noise level estimation, in the implementation proposed by \citet{ipol.2013.45}.

\subsection{The \citet{boracchi2014novelty} sparsity-based background model }
\label{sec:nfa_boracchi}

In this method the detection is done using a threshold on the Mahalanobis distance. \citet{chen2007new} has shown, as a generalization of Chebyshev's inequality, that for a random vector $X$ of dimension $d$ with covariance matrix $C$ we have 
$$
\mathbb{P}((X-\mathbb{E}(X))^TC^{-1}(X-\mathbb{E}(X)) \geqslant \gamma) \leqslant \frac{d}{\gamma},
$$
Moreover, it has been shown in~\cite{navarro2014can} that this inequality is sharp if no other assumptions are made on $X$. Therefore, in the case of this method, for a candidate $X_i$  and a reference set $P$,
\begin{equation}
\label{eq:chen_majoration_boracchi}
\mathbb{P}(d_\mathcal{M}(X_i) \geqslant \gamma) \leqslant \frac{2}{\gamma^2},
\end{equation}
where the Mahalanobis distance $d_\mathcal{M}(\cdot)$ is computed with respect to the empirical mean and covariance of the set $P$.
Hence, the function 
$$
f(i, \mathbf{x}) = N\mathbb{P}(d_\mathcal{M}(X_i) \geqslant d_\mathcal{M}(\mathbf{x}_i))
$$
is  clearly an NFA associated to  the  method.
Using (\ref{eq:chen_majoration_boracchi}) and the obvious fact that a function whose value is always above an NFA  also is an NFA, we deduce that the  test function
$$
f(i, \mathbf{x}) = \frac{2N}{d_\mathcal{M}(\mathbf{x}_i)^2}
$$
 also is a NFA.
Thus, a detection is made if
$$
\frac{2N}{d_\mathcal{M}(\mathbf{x}_i)^2} \leqslant \varepsilon,
$$
which leads to a threshold $\gamma_\varepsilon$, such that
$$
d_\mathcal{M}(\mathbf{x}_i) \geqslant \gamma_\varepsilon := \sqrt{\frac{2N}{\varepsilon}}.
$$
While the method was originally presented as using an external database of anomaly free detections, we use it on the image itself \textit{i.e.} the dictionary is learned on the image, under the  assumption  that  it  presents too  few anomalies to disturb the dictionary.

\subsection{The \citet{mishne2014multiscale} non-local self-similar model}

There is  no obvious way to formalize this method under the \emph{a-contrario} framework.   For the experiments that we present in Section~\ref{sec:experiments}, we use the detection threshold suggested in the original paper even though there is no actual theoretical justification.

\section{Experiments}
\label{sec:experiments}
In this section we shall compare the six methods analyzed in Section~\ref{sec:generalization}. In what follows, we detail the different variants that we finally compare:
\begin{itemize}
\item The  \citet{grosjean2009contrario} stochastic parametric background model as explained in Section~\ref{GrosjeanandDavyNFA}. The NFA computation has been adapted to take into account both tails of a pixel's distribution, with tests performed on all pixels. We denote this method by~\texttt{Grosjean}. 
\item The \citet{aiger2010phase} Fourier homogeneous model using the \textit{a-contrario} detection threshold as specified in Section~\ref{sec:nfa_aiger}. We denote this method by~\texttt{Aiger}.
\item The \citet{zontak2010defect} non-local self-similar model using the \textit{a-contrario} detection  threshold as specified in Section~\ref{zontakcohenNFA}. We denote this method by~\texttt{Zontak}.
\item The sparsity-based background model of \citet{boracchi2014novelty} using the \textit{a-contrario} detection  threshold as specified in Section~\ref{sec:nfa_boracchi}. We denote this method by~\texttt{Boracchi}.
\item The non-local self-similar model of~\citet{mishne2014multiscale} with the detection threshold as detailed in the original publication. We denote this method by~\texttt{Mishne}.
\item The non-local self-similar model of~\citet{davy2018reducing} where the phase only transform (PHOT) is applied before the distribution normalization. The  NFA is computed as explained in Section~\ref{GrosjeanandDavyNFA}. We denote this method by~\texttt{Davy}.
\end{itemize}

\medskip\noindent
We propose two types of experimental comparison.
\begin{itemize}
\item The first comparison is a \textbf{qualitative} sanity check. For this qualitative analysis we tested on synthetic examples having obvious anomalies of different types (color,  shape,  cluster), or  inexistent  (white noise).  These toy examples provide a sanity check since one would expect all  algorithms  to perform perfectly on them.   We will also examine the results of the competitors on challenging examples taken from anomaly detection articles.

\item The second protocol is a \textbf{quantitative} evaluation. We generated  anomaly-free images as samples of colored random Gaussian noise. Being  a spatially homogeneous random process, such images should remain neutral for an anomaly detector.
We then introduced small anomalies to these images and evaluated whether these synthetic anomalies were detected by the competitors.  This leads to evaluate a true positive detection rate (TP) for  each method on these images. 
We also evaluated how much of the anomaly free background was wrongly detected, namely the false positive detection rate (FP). Disposing of TP-FP pairs yields ROC curves  that will be  opportunely discussed.  Undoubtedly, the colored Gaussian noise used in this experiment could be replaced by any other spatially homogeneous random process.  We varied the background texture by varying strongly  the  process's power spectrum. 
\end{itemize}

\subsection{Qualitative evaluation \label{qualitative evaluation}}
\label{sec:exp_qualitative}
\begin{figure*}
\centering
\footnotesize
\begin{minipage}[c]{\textwidth}
\centering
\rot{\hspace{1.5em}\vphantom{A}\vphantom{[}input image}
\includegraphics[height=0.106\textheight]{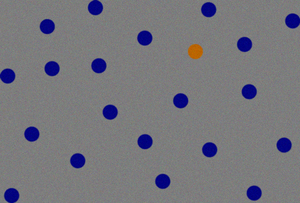}
\includegraphics[height=0.106\textheight]{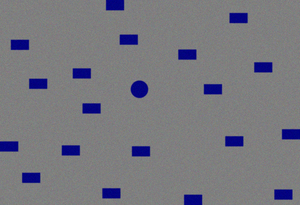}
\includegraphics[height=0.106\textheight]{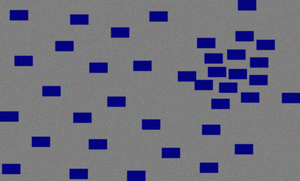}
\includegraphics[height=0.106\textheight]{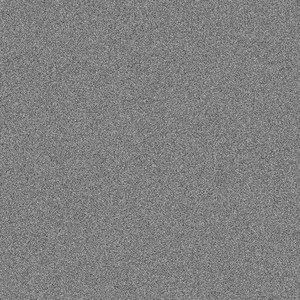}
\includegraphics[height=0.106\textheight]{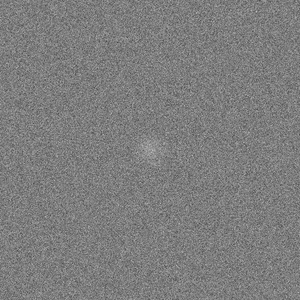}

\end{minipage}

\begin{minipage}[c]{\textwidth}
\centering
\rot{\hspace{1.5em}\vphantom{A}\vphantom{[}residual \cite{davy2018reducing}}
\includegraphics[height=0.106\textheight]{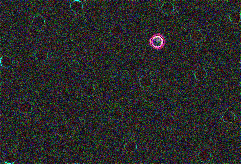}
\includegraphics[height=0.106\textheight]{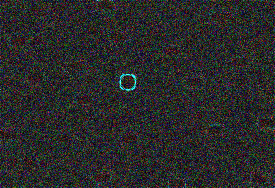}
\includegraphics[height=0.106\textheight]{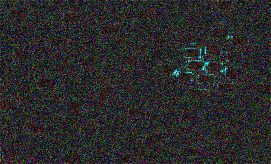}
\includegraphics[height=0.106\textheight]{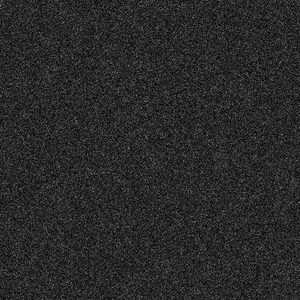}
\includegraphics[height=0.106\textheight]{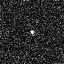}

\end{minipage}

\begin{minipage}[c]{\textwidth}
\centering
\rot{\hspace{1.7em}\vphantom{A}\vphantom{[} \texttt{Davy}~\cite{davy2018reducing}}
\includegraphics[height=0.106\textheight]{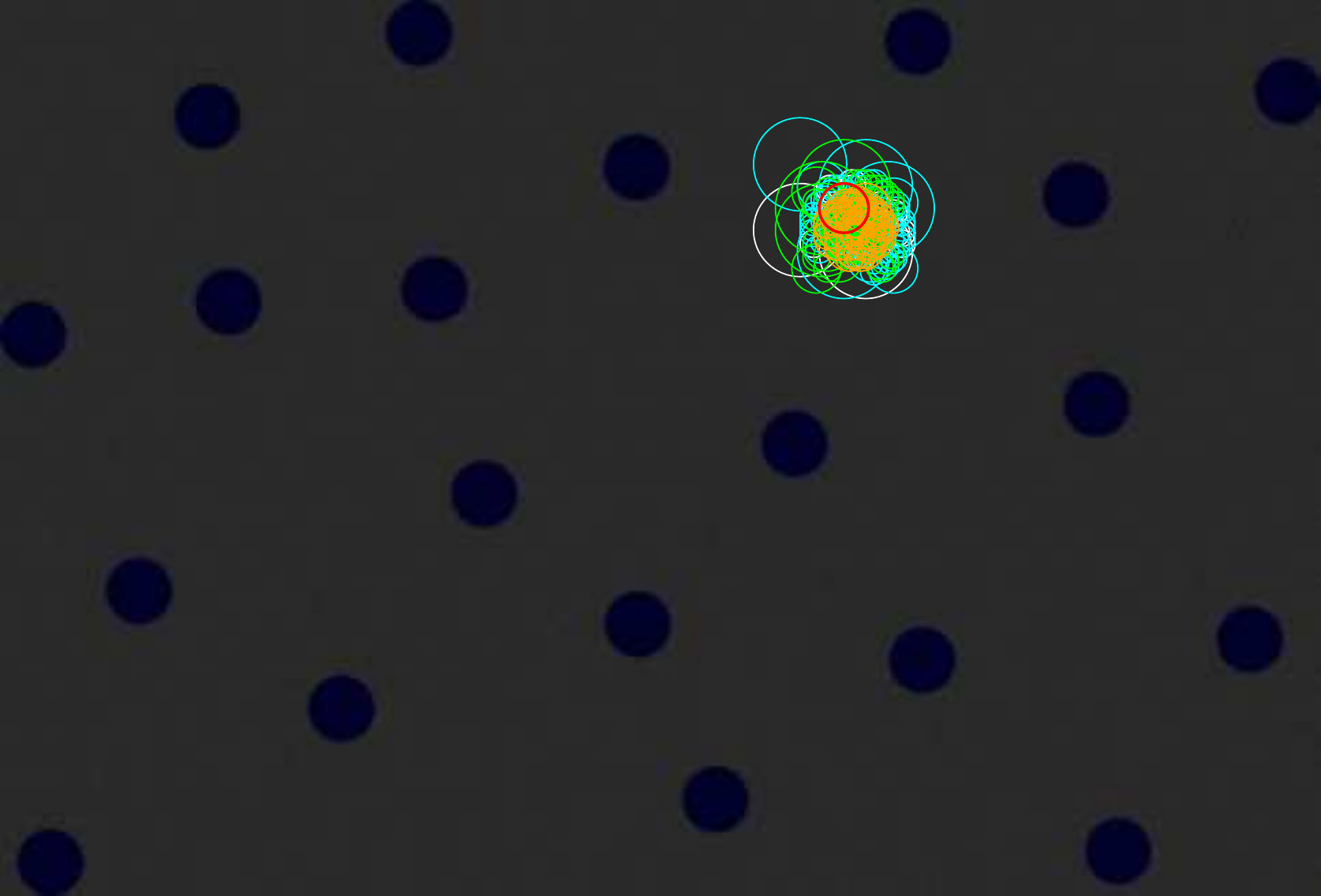}
\includegraphics[height=0.106\textheight]{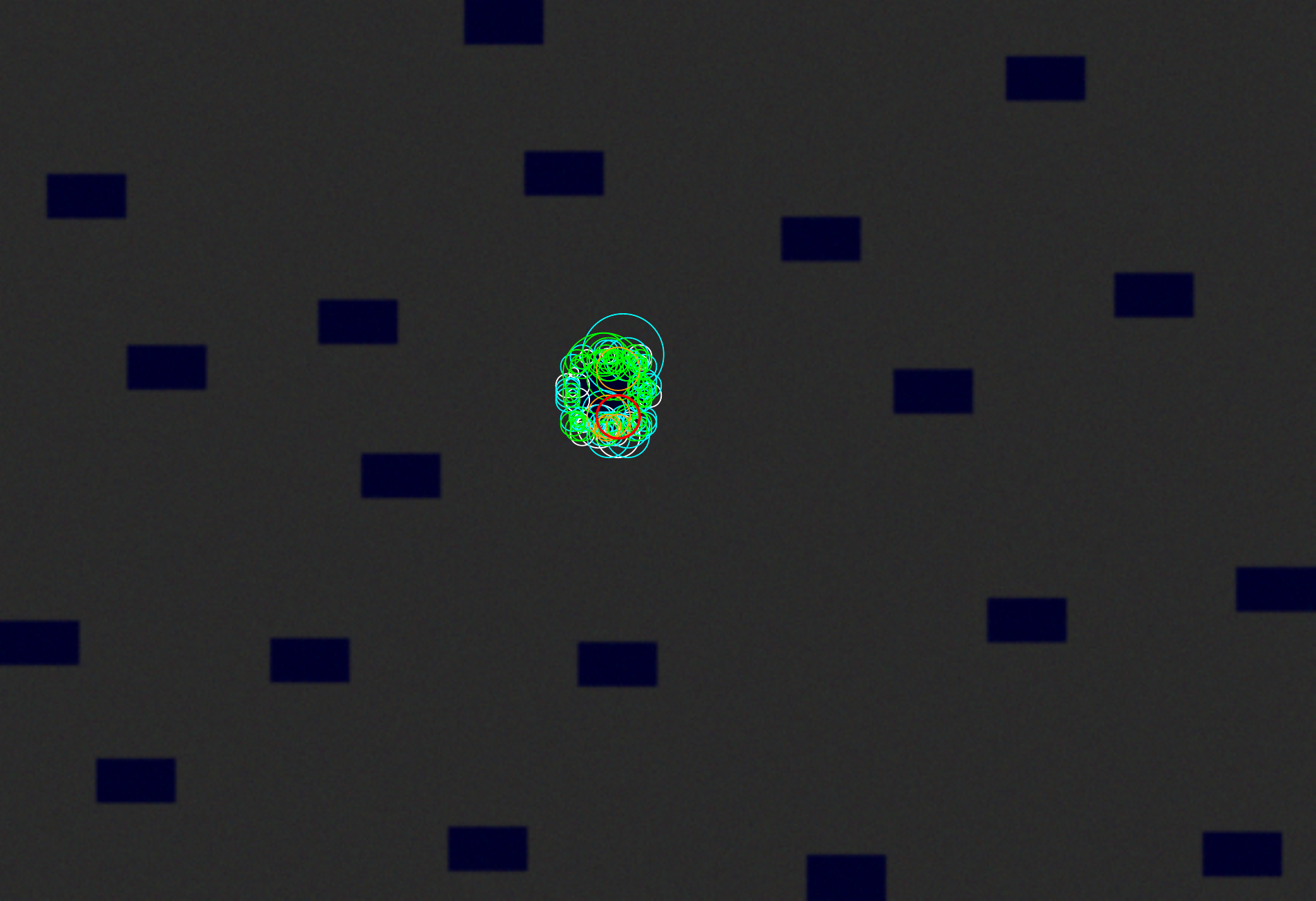}
\includegraphics[height=0.106\textheight]{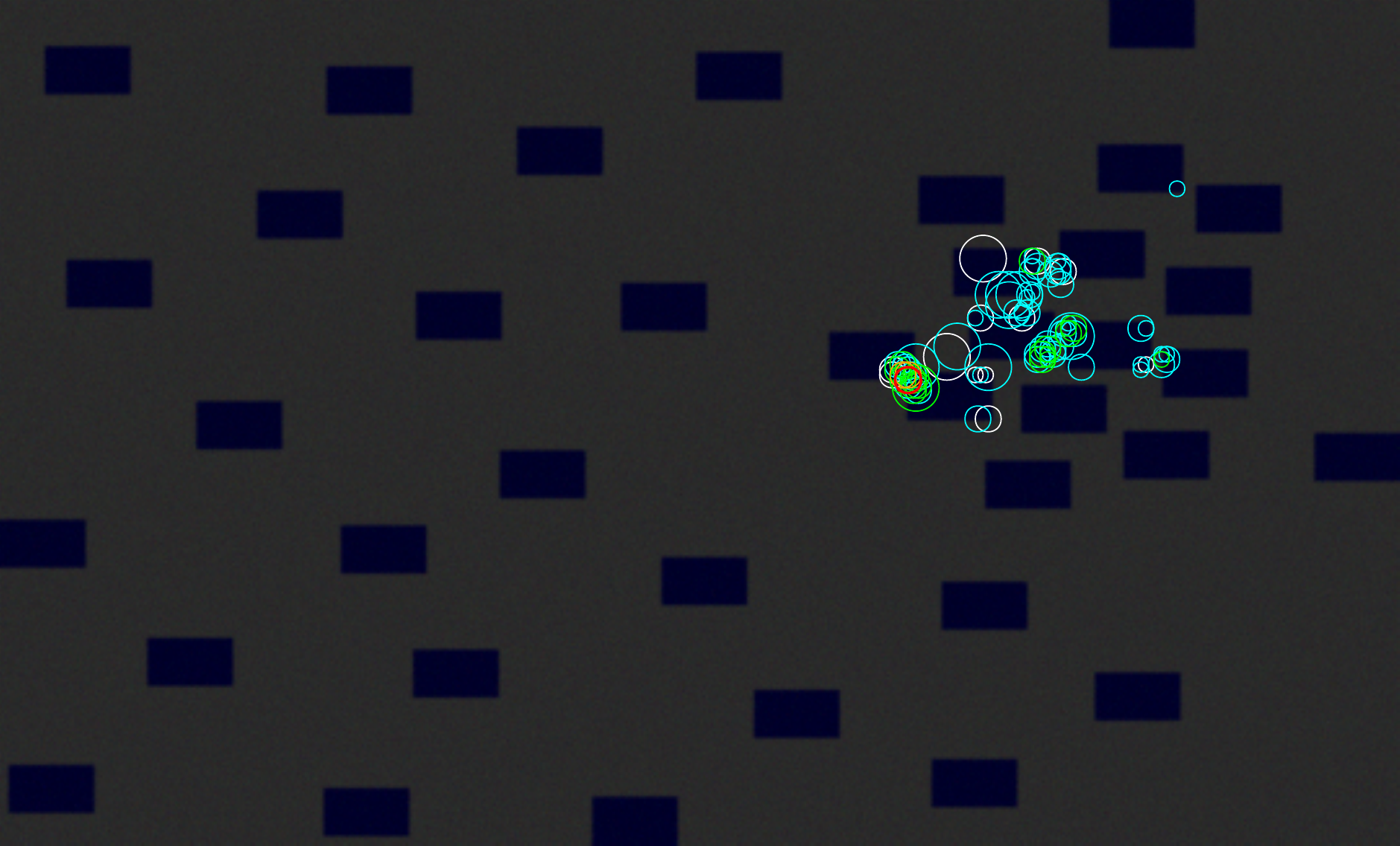}
\includegraphics[height=0.106\textheight]{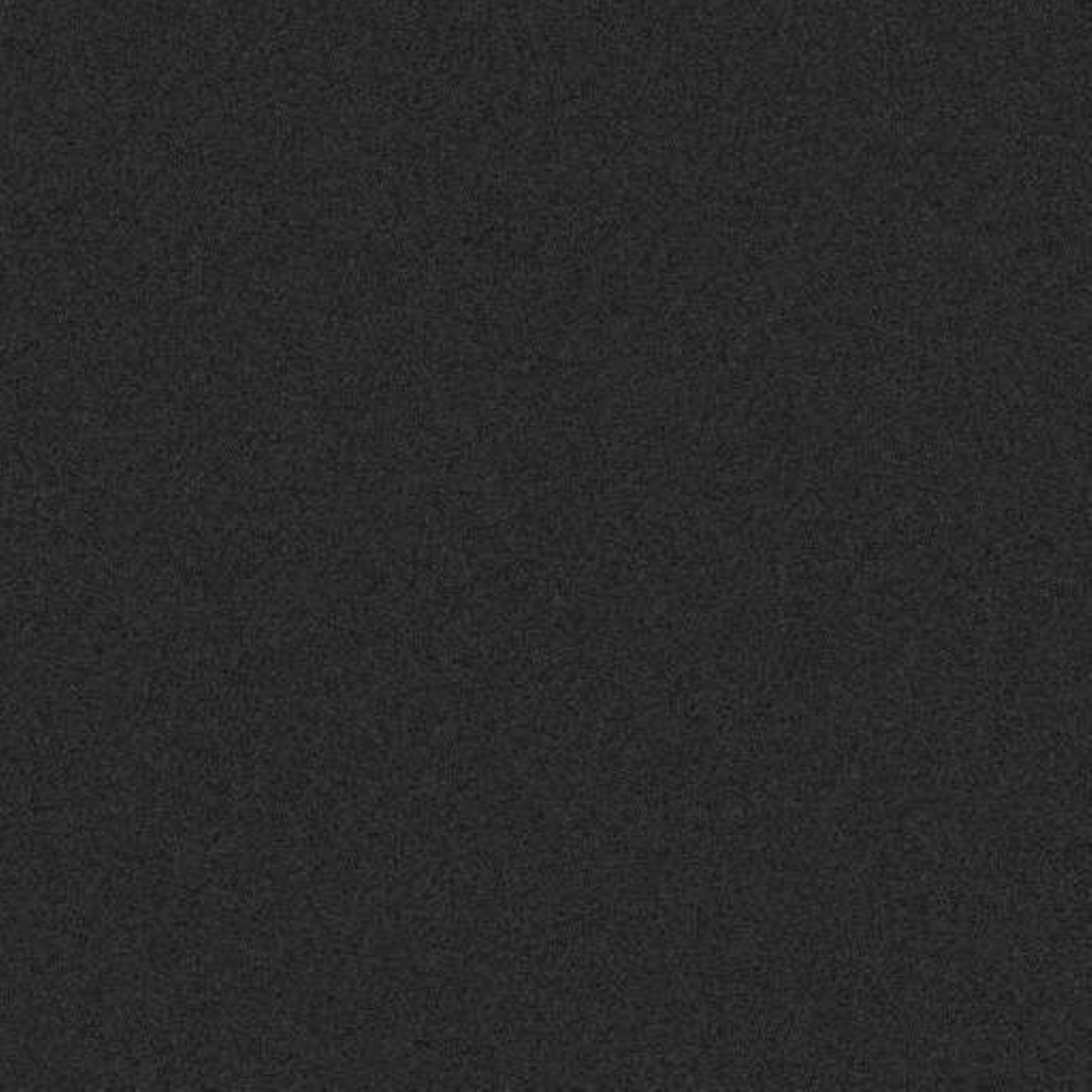}
\includegraphics[height=0.106\textheight]{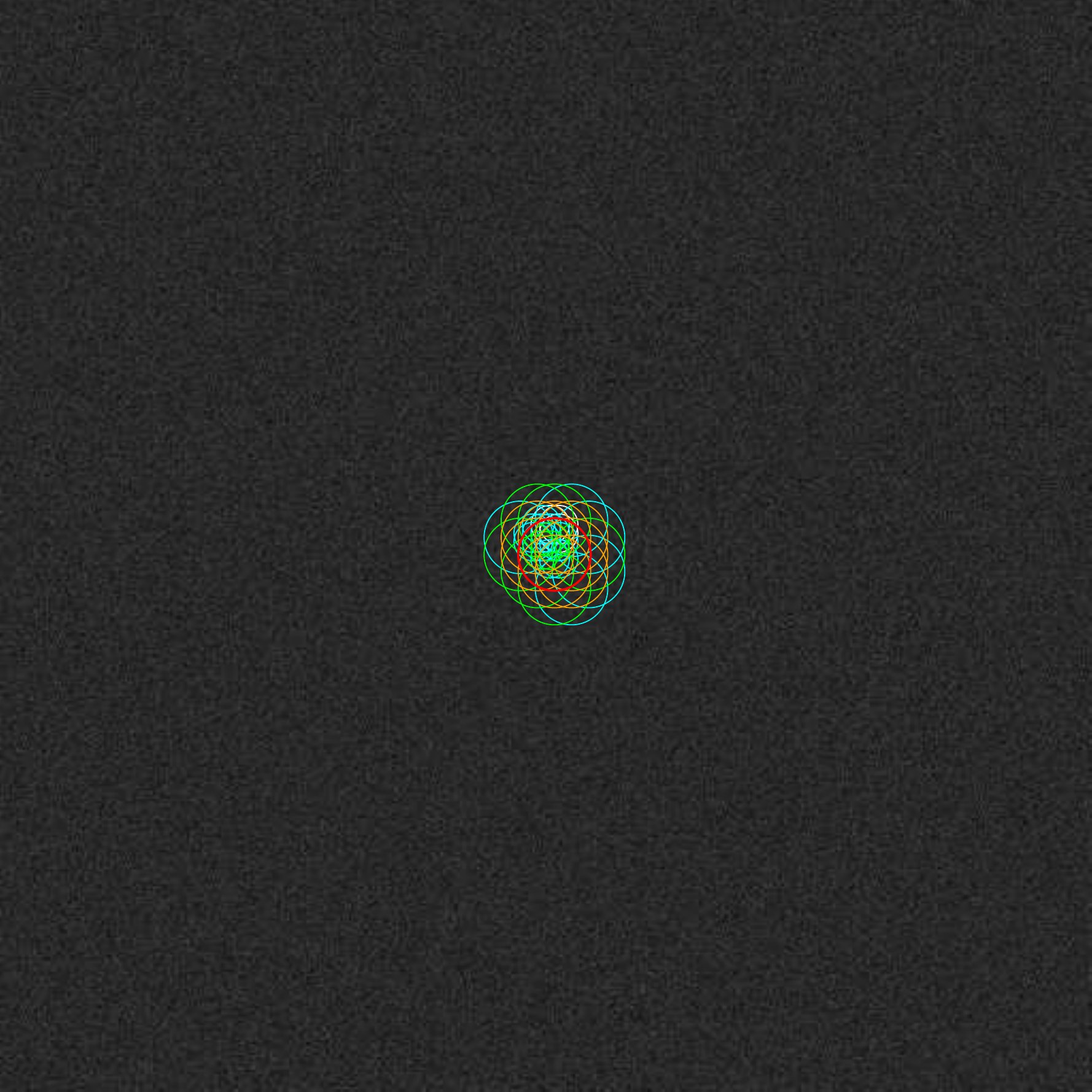}

\end{minipage}

\begin{minipage}[c]{\textwidth}
\centering
\rot{\hspace{3em}\vphantom{A}\vphantom{[}\texttt{Aiger}~\cite{aiger2010phase}}
\includegraphics[height=0.106\textheight]{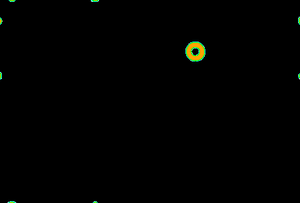}
\includegraphics[height=0.106\textheight]{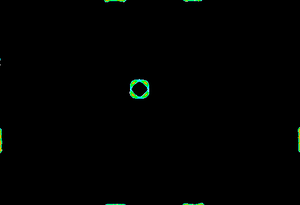}
\includegraphics[height=0.106\textheight]{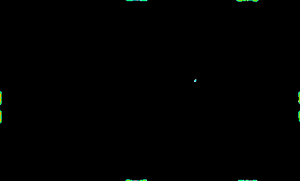}
\includegraphics[height=0.106\textheight]{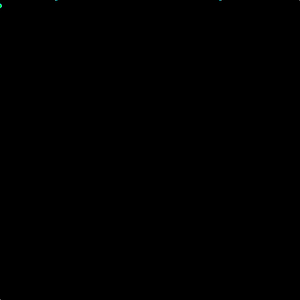}
\includegraphics[height=0.106\textheight]{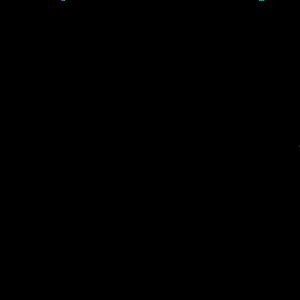}

\end{minipage}

\begin{minipage}[c]{\textwidth}
\centering
\rot{\hspace{2.5em}\vphantom{A}\vphantom{[}\texttt{Zontak}~\cite{zontak2010defect}}
\includegraphics[height=0.106\textheight]{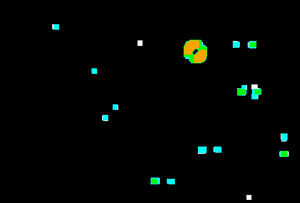}
\includegraphics[height=0.106\textheight]{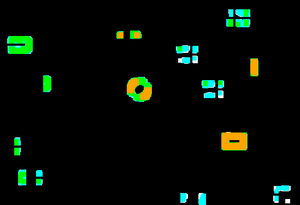}
\includegraphics[height=0.106\textheight]{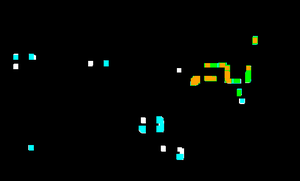}
\includegraphics[height=0.106\textheight]{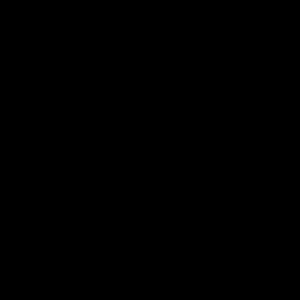}
\includegraphics[height=0.106\textheight]{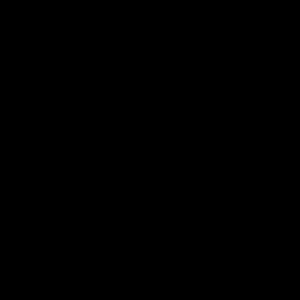}

\end{minipage}

\begin{minipage}[c]{\textwidth}
\centering
\rot{\hspace{2.2em}\vphantom{A}\vphantom{[}\texttt{Mishne}~\cite{mishne2014multiscale}}
\includegraphics[height=0.106\textheight]{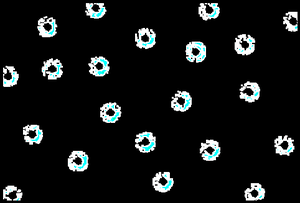}
\includegraphics[height=0.106\textheight]{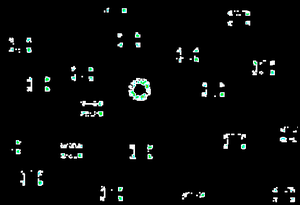}
\includegraphics[height=0.106\textheight]{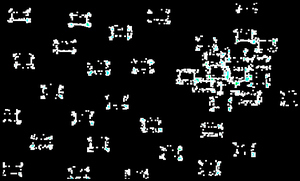}
\includegraphics[height=0.106\textheight]{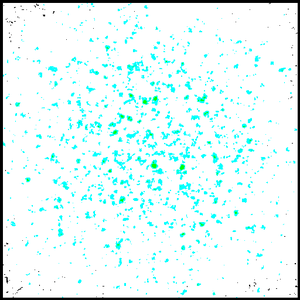}
\includegraphics[height=0.106\textheight]{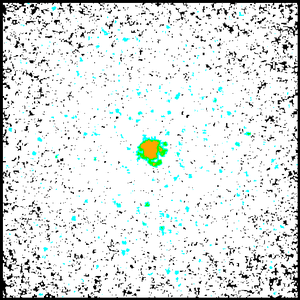}

\end{minipage}

\begin{minipage}[c]{\textwidth}
\centering
\rot{\hspace{1.3em}\vphantom{A}\vphantom{[}\texttt{Grosjean}~\cite{grosjean2009contrario}}
\includegraphics[height=0.106\textheight]{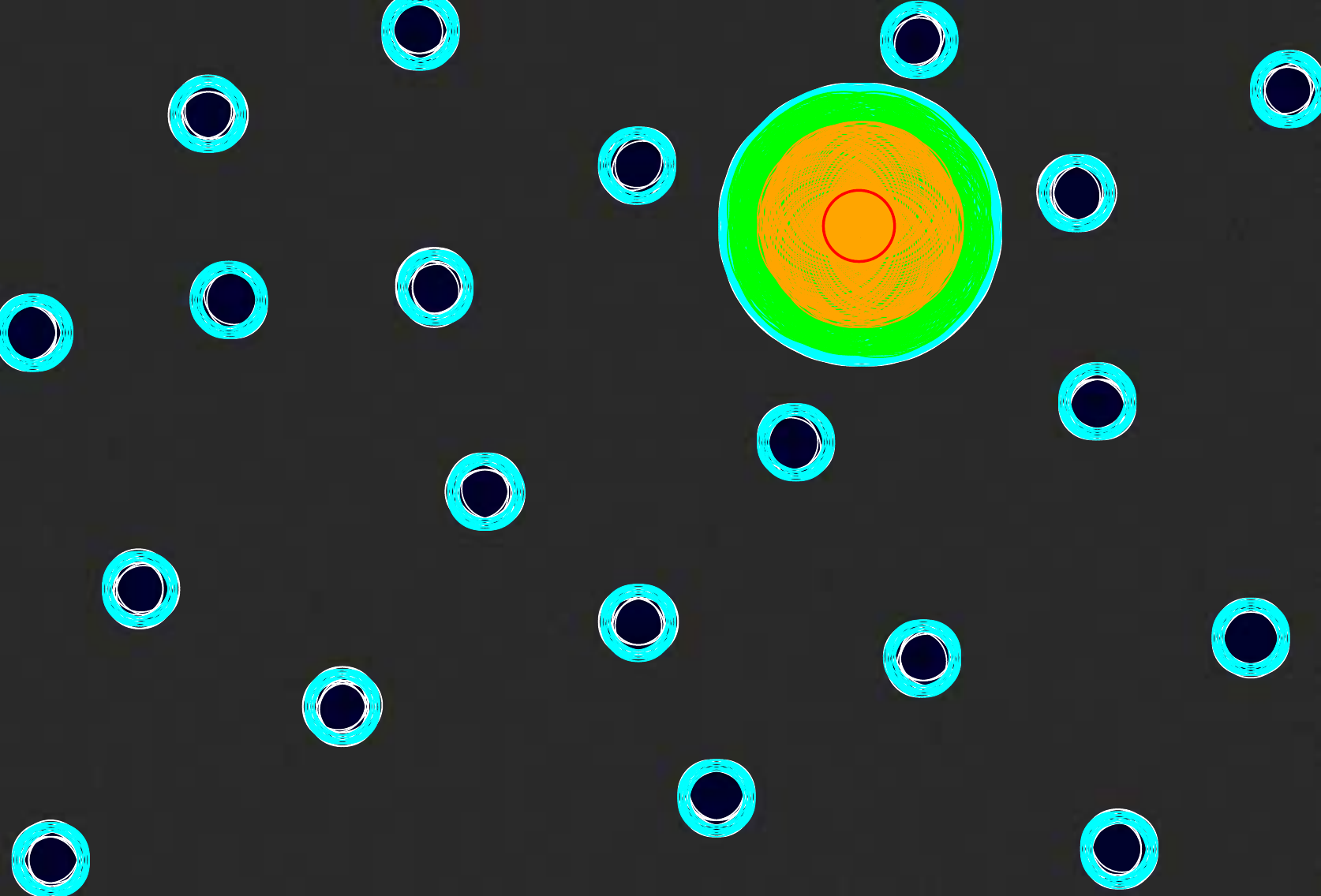}
\includegraphics[height=0.106\textheight]{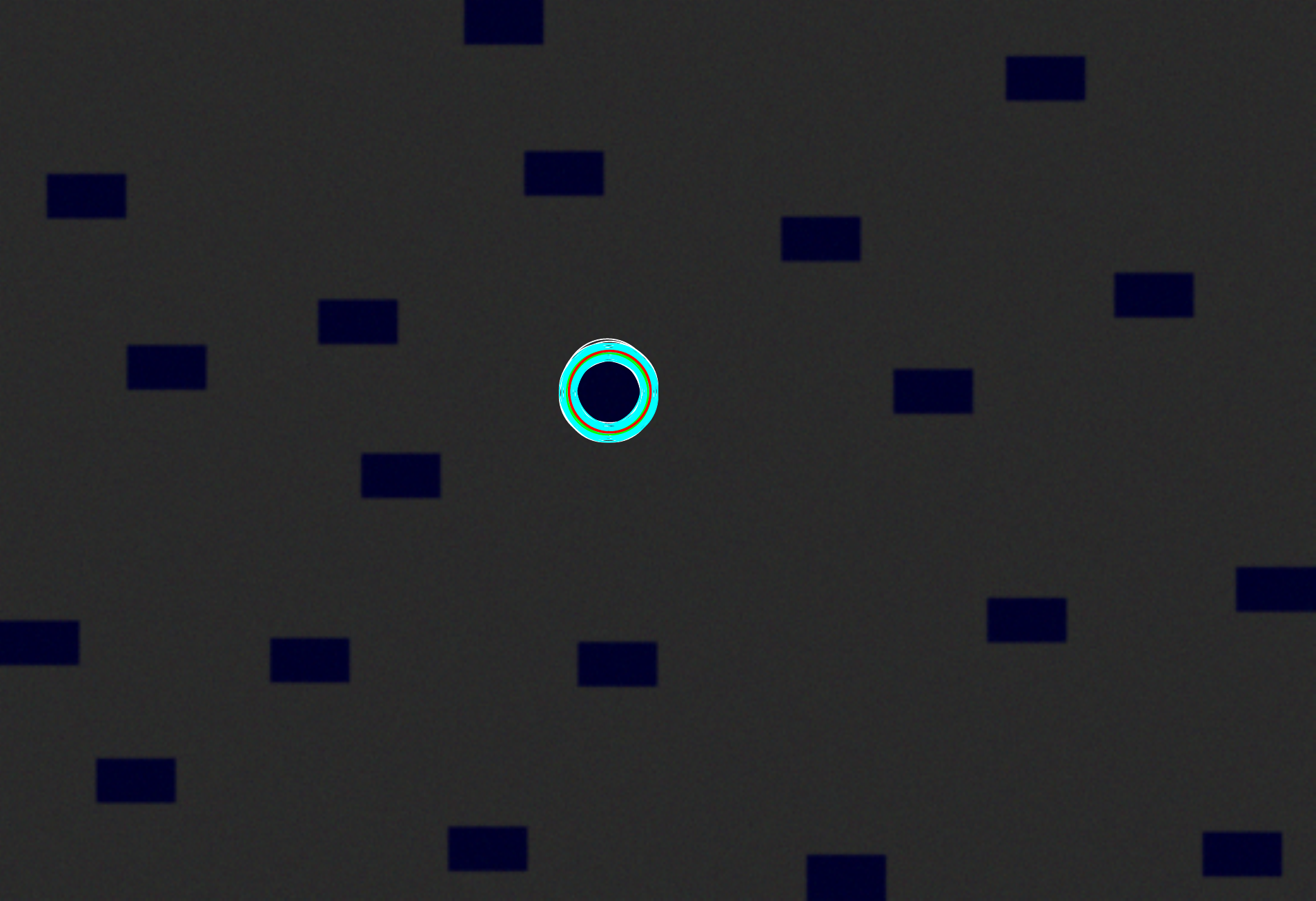}
\includegraphics[height=0.106\textheight]{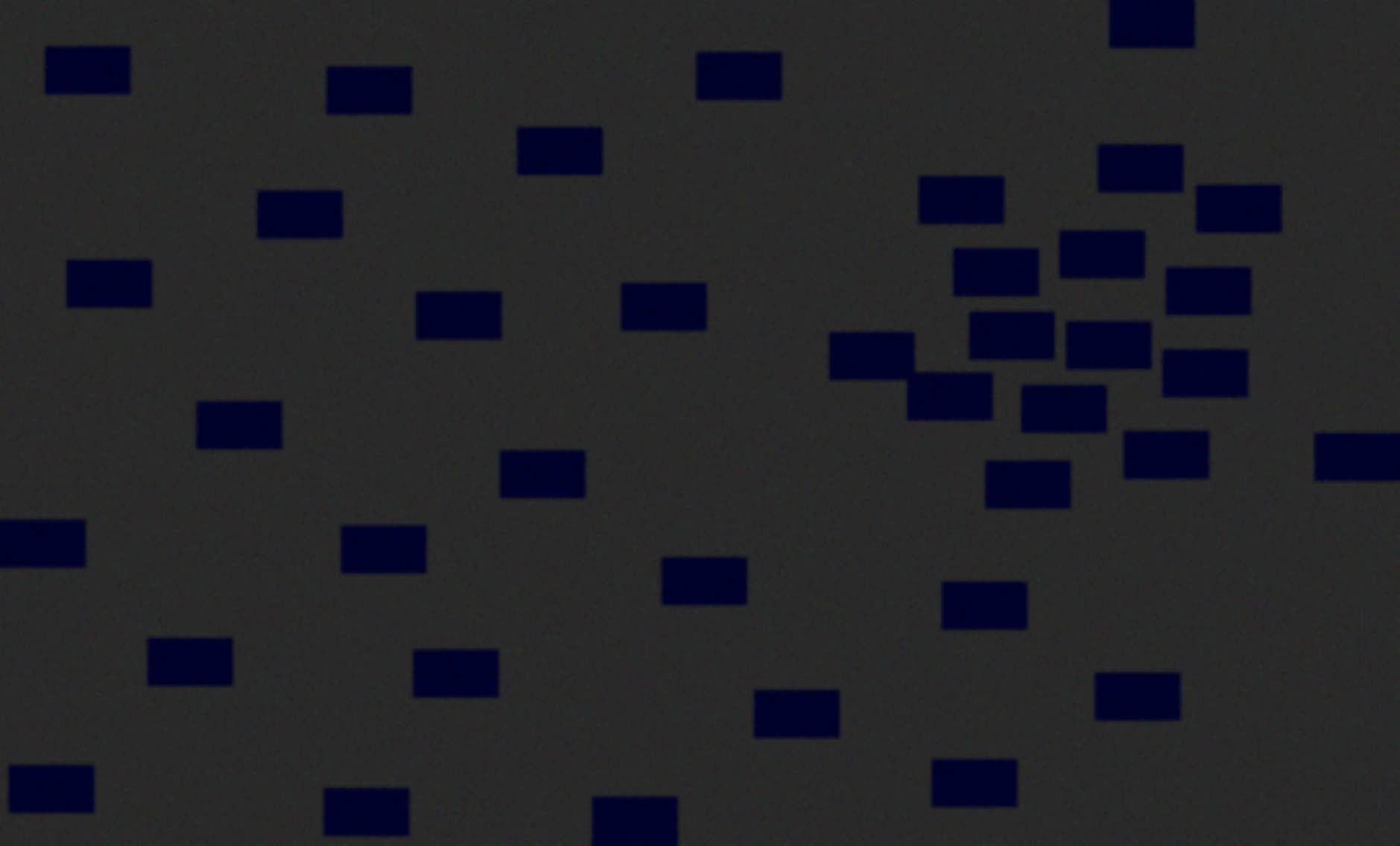}
\includegraphics[height=0.106\textheight]{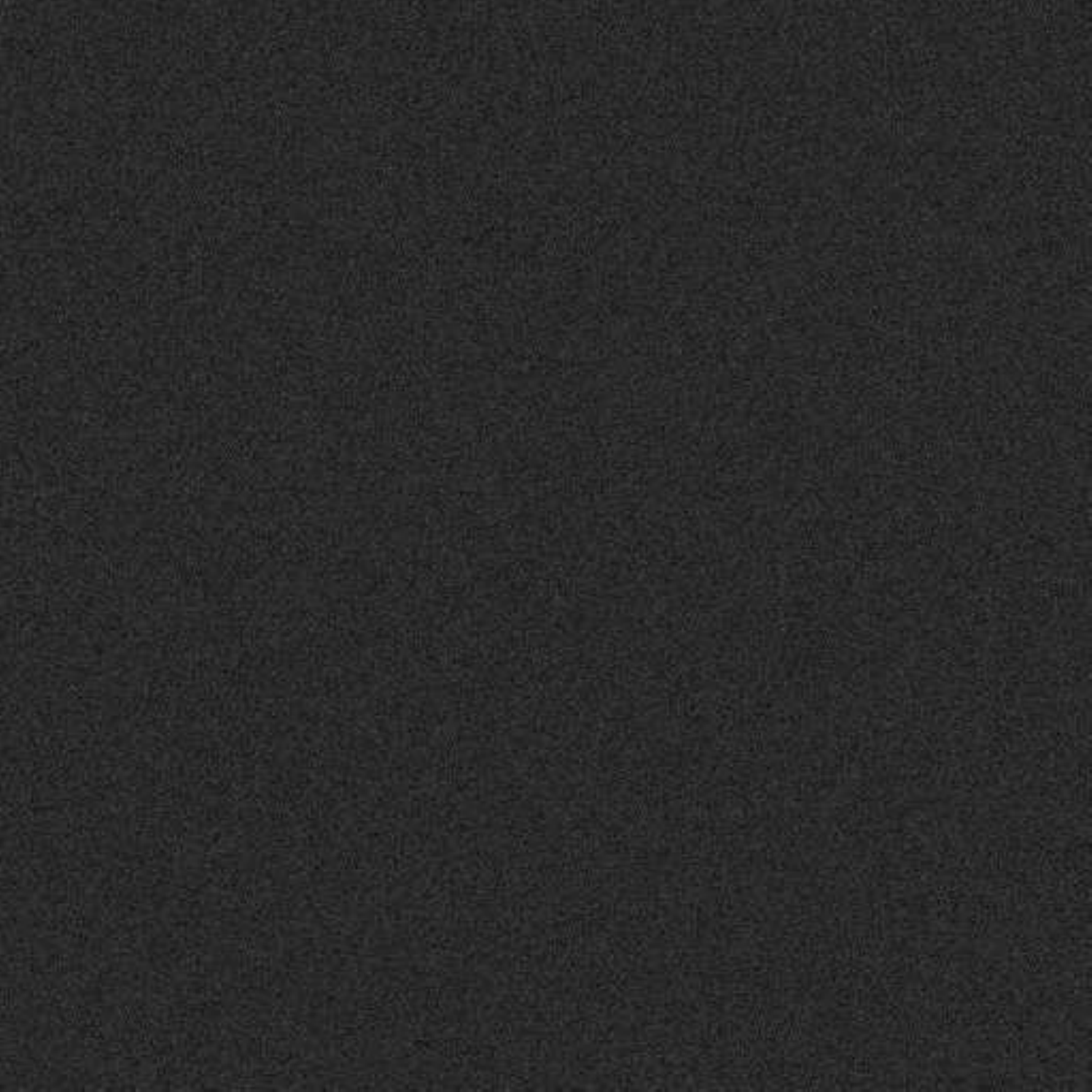}
\includegraphics[height=0.106\textheight]{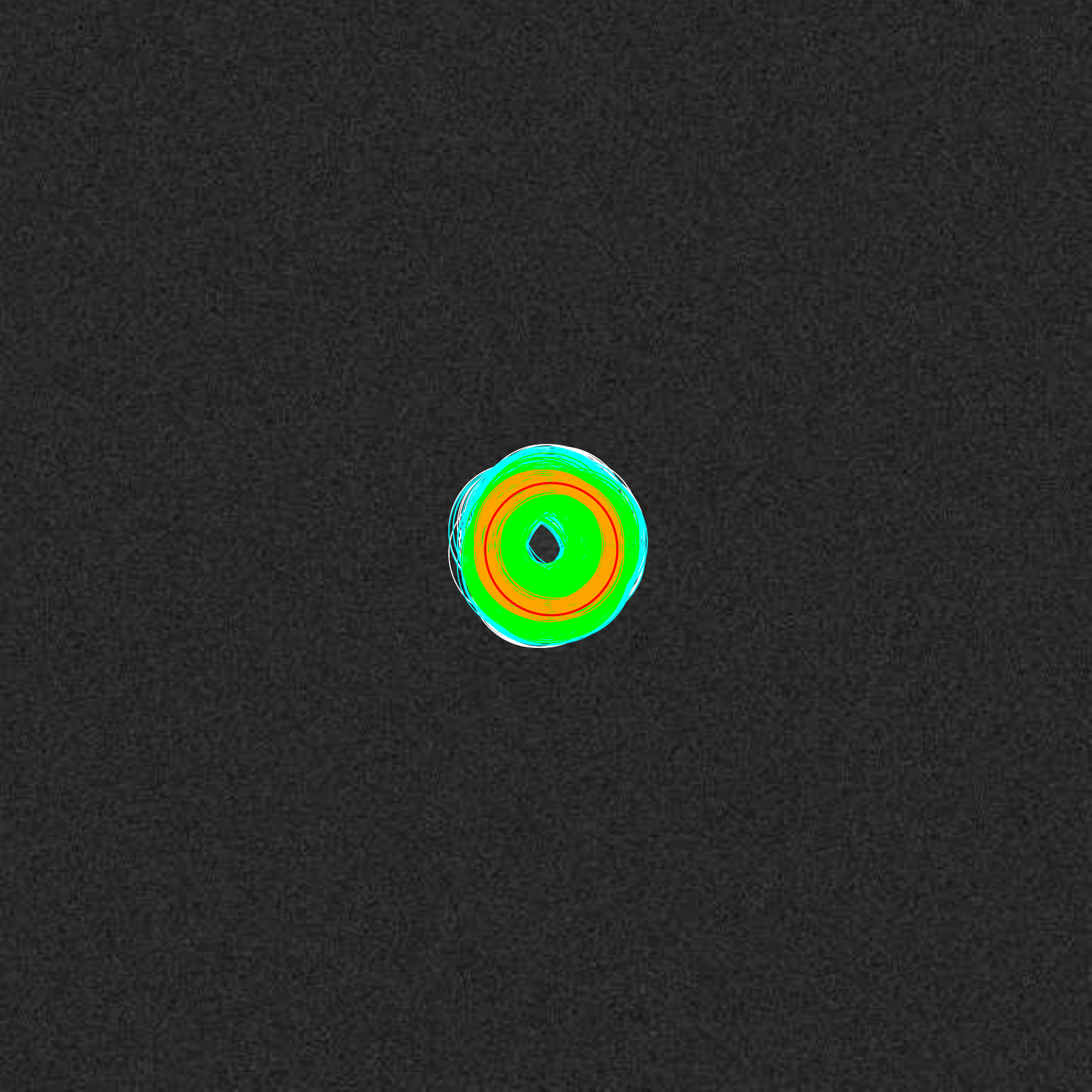}

\end{minipage}

\begin{minipage}[c]{\textwidth}
\centering
\rot{\hspace{1.0em}\vphantom{A}\vphantom{[}\texttt{Boracchi}~\cite{boracchi2014novelty}}
\includegraphics[height=0.106\textheight]{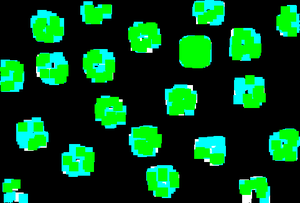}
\includegraphics[height=0.106\textheight]{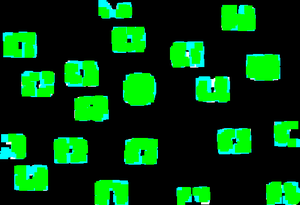}
\includegraphics[height=0.106\textheight]{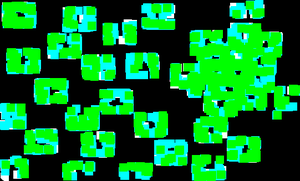}
\includegraphics[height=0.106\textheight]{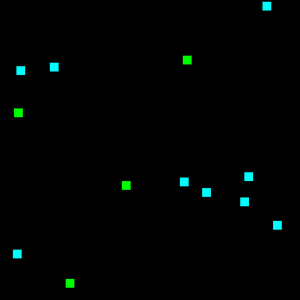}
\includegraphics[height=0.106\textheight]{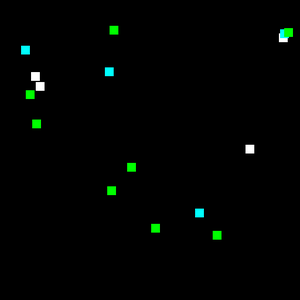}

\end{minipage}

\caption{From left to right: Image presenting an anomaly in colors, in shape and in density, image of pure noise,  and image of noise with an anomaly in the middle (from~\cite{grosjean2009contrario}). From top to bottom: The original image, the image residual of one of the scales computed in~\cite{davy2018reducing} (the scale shown is the one where the anomaly is the most salient, the contrast has been adjusted for visualization purpose), algorithm detections for:
\cite{davy2018reducing}, \cite{aiger2010phase}, \cite{zontak2010defect}, \cite{mishne2014multiscale}, \cite{grosjean2009contrario} and \cite{boracchi2014novelty}.  Detections are shown using the following color coding: white is a weak detection - threshold with NFA $\in [10^{-3}, 10^{-2}]$, cyan is a mild detection - threshold with NFA $\in [10^{-8}, 10^{-3}]$, green is a strong detection - threshold with NFA $\in [10^{-21}, 10^{-8}]$, and orange is very strong  - threshold with NFA $\le 10^{-21}$. When available, red is the detection with the threshold corresponding to the lowest NFA. For~\cite{mishne2014multiscale} we adopted a similar color coding: white between $0$ and $0.5$, cyan between $0.5$ and $0.7$, green between $0.7$ and $0.9$ and orange above $0.9$.}
\label{fig:results_1}
\end{figure*}

 The toy examples are probably the easiest to analyze. We show the results in Figure \ref{fig:results_1}. We generated images in  the classic form  used in anomaly detection benchmarks like in \cite{riche2013rare2012}, where the anomaly is the shape or the color that is unique in the figure. In the third toy example most rectangles are well spaced except in a small region. The anomaly therefore is a change in spatial density. Even though these examples are extremely simple to analyze, they appear to challenge several methods, as can be seen in Figure \ref{fig:results_1}. Only \citet{davy2018reducing} is able to detect accurately the anomaly in all three examples. This is  explained in the second row where the residual after background subtraction is shown. In the residual details of  the anomalies stand out on a noise-like background.  
 While \citet{aiger2010phase} works well with the color and the shape, it fails to detect the spatial density anomaly. \citet{zontak2010defect} detects well but also lots of false detection. The other methods \citet{zontak2010defect,mishne2014multiscale,grosjean2009contrario} and \citet{boracchi2014novelty} over-detect the contours of the non anomalous shapes, thus leading to many false positives. We also tried a sanity check with a pure white Gaussian noise image. This is done in the last two examples of Figure \ref{fig:results_1}. \citet{davy2018reducing},  and \citet{grosjean2009contrario} soundly detect no anomaly in white noise, as expected. However a few detections are made by \citet{boracchi2014novelty} and almost everything is detected by \citet{mishne2014multiscale}. It can be noted that the background model of the first three papers is directly respected in the case of white Gaussian noise, which explains the perfect result.  (In the case of the model of \citet{davy2018reducing}, it has to be noted that non-local means asymptotically transforms white Gaussian noise into white Gaussian noise \cite{buades2008nonlocal}). The  over-detection in \citet{mishne2014multiscale} can be explained by the lack of an automatic statistical threshold. The  few  spurious detections in \citet{boracchi2014novelty}  show that the feature used for the detection doesn't  follow a Gaussian distribution,  contrarily to  the method's testing assumption.  It is also clear that one cannot build a sound sparse dictionary for white noise.
 
 The same test was done after adding a small anomalous spot to the noise, and the conclusion is similar: \cite{davy2018reducing,grosjean2009contrario} perform well, \cite{boracchi2014novelty} has a couple of false detections and doesn't detect the anomaly. One method, \citet{zontak2010defect}, doesn't detect anything. Finally \citet{mishne2014multiscale} over-detects. Both noise images were taken from \citet{grosjean2009contrario}.

\begin{figure*}
\centering
\footnotesize
\begin{minipage}[c]{\textwidth}
\centering
\rot{\hspace{3em}\vphantom{A}\vphantom{[}input image}
\includegraphics[height=0.116\textheight]{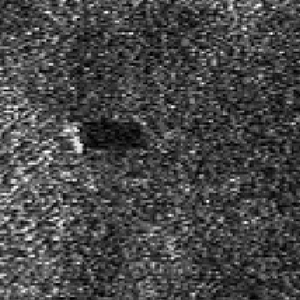}
\includegraphics[height=0.116\textheight]{Experiments/Papers/textile.png}
\includegraphics[height=0.116\textheight]{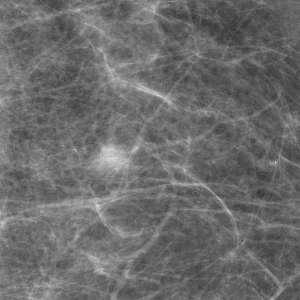}
\includegraphics[height=0.116\textheight]{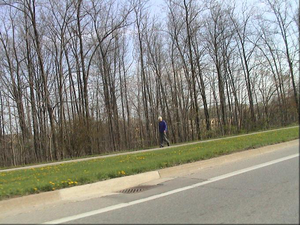}
\includegraphics[height=0.116\textheight]{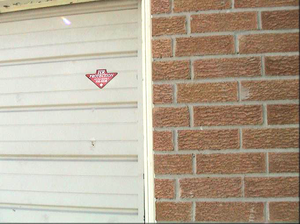}

\end{minipage}

\begin{minipage}[c]{\textwidth}
\centering
\rot{\hspace{1.5em}\vphantom{A}\vphantom{[}residual~\cite{davy2018reducing}}
\includegraphics[height=0.116\textheight]{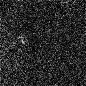}
\includegraphics[height=0.116\textheight]{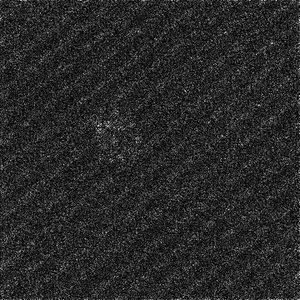}
\includegraphics[height=0.116\textheight]{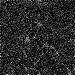}
\includegraphics[height=0.116\textheight]{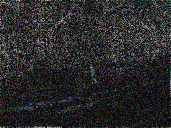}
\includegraphics[height=0.116\textheight]{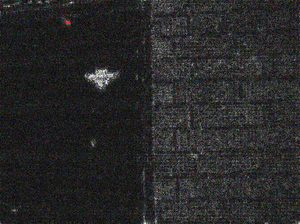}

\end{minipage}

\begin{minipage}[c]{\textwidth}
\centering
\rot{\hspace{1.5em}\vphantom{A}\vphantom{[}\texttt{Davy}~\cite{davy2018reducing}}
\includegraphics[height=0.116\textheight]{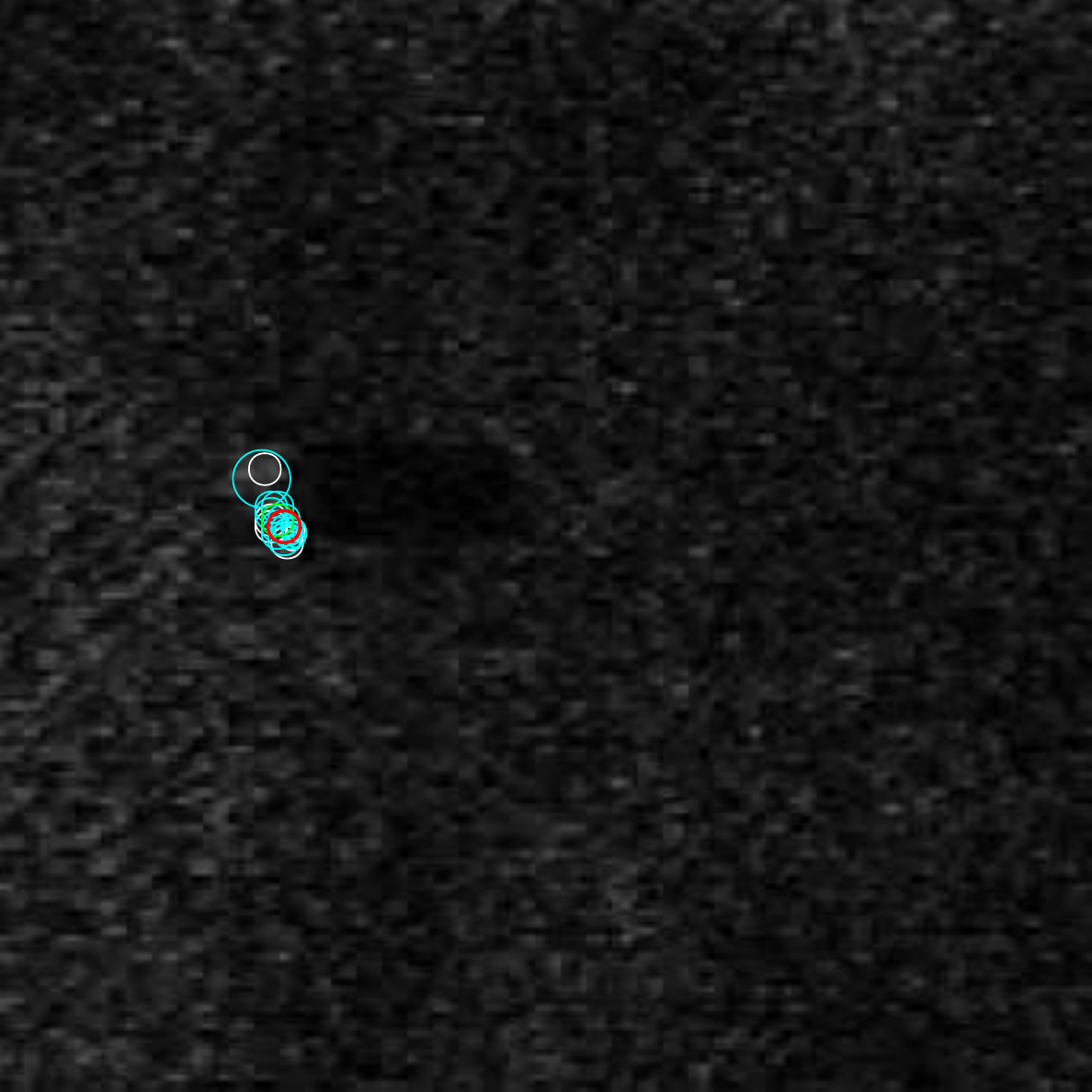}
\includegraphics[height=0.116\textheight]{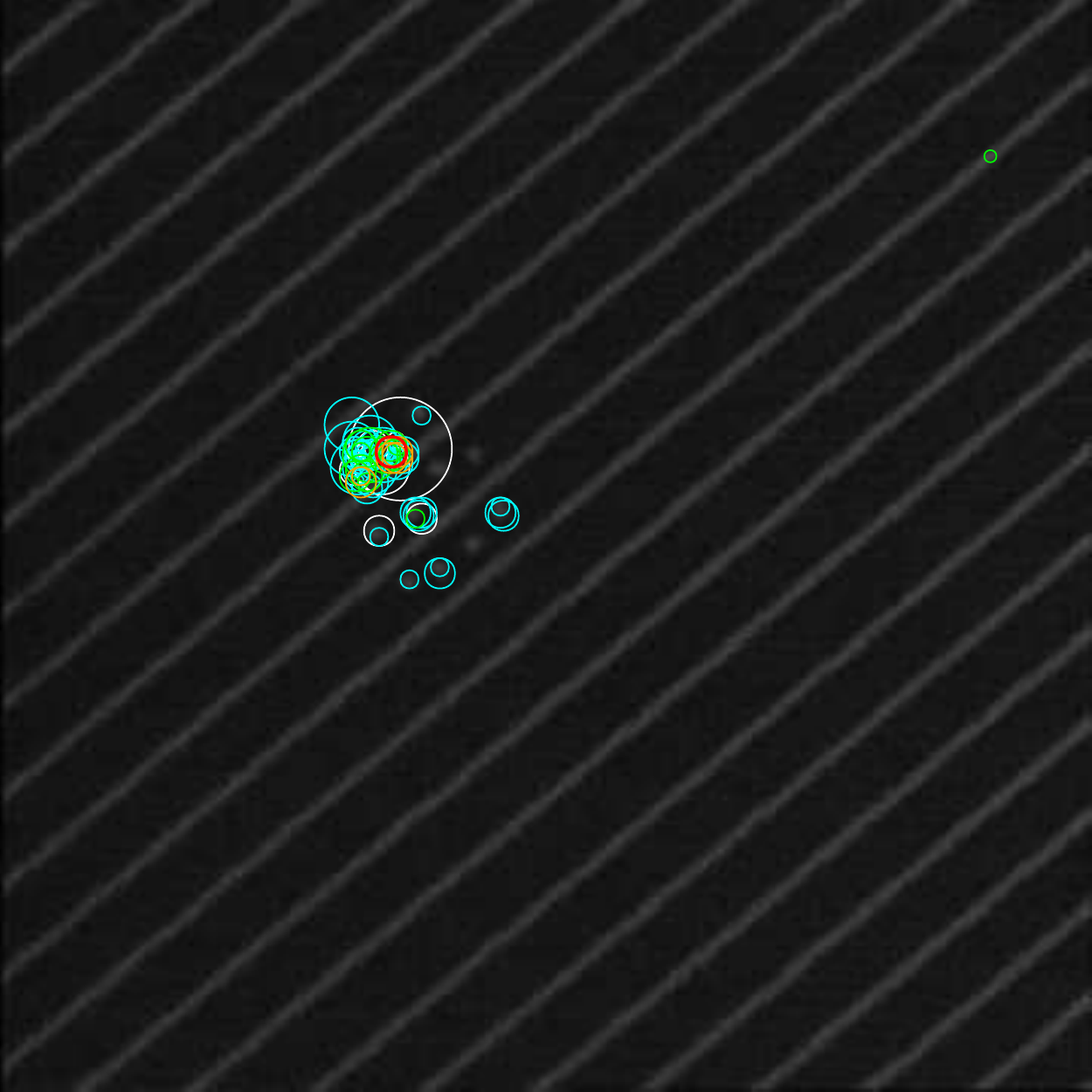}
\includegraphics[height=0.116\textheight]{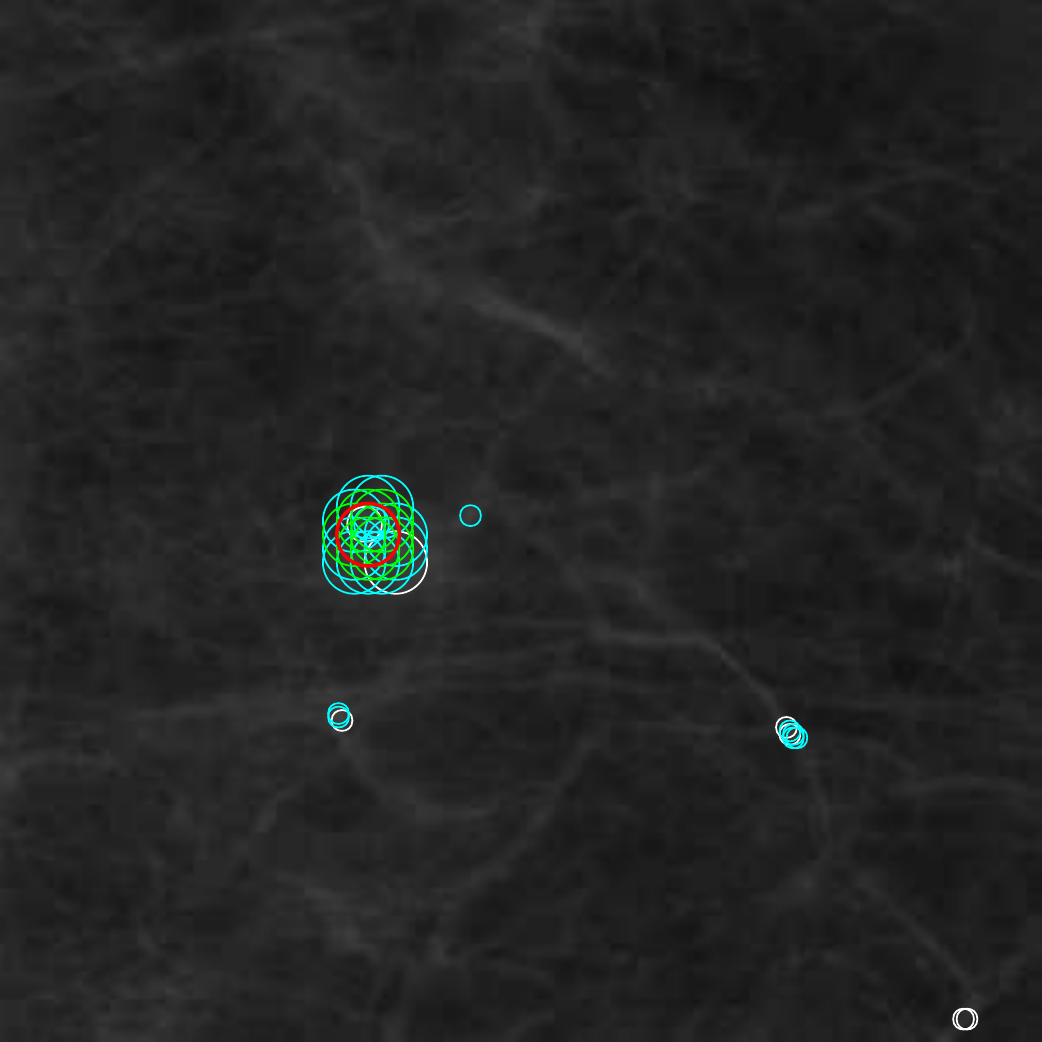}
\includegraphics[height=0.116\textheight]{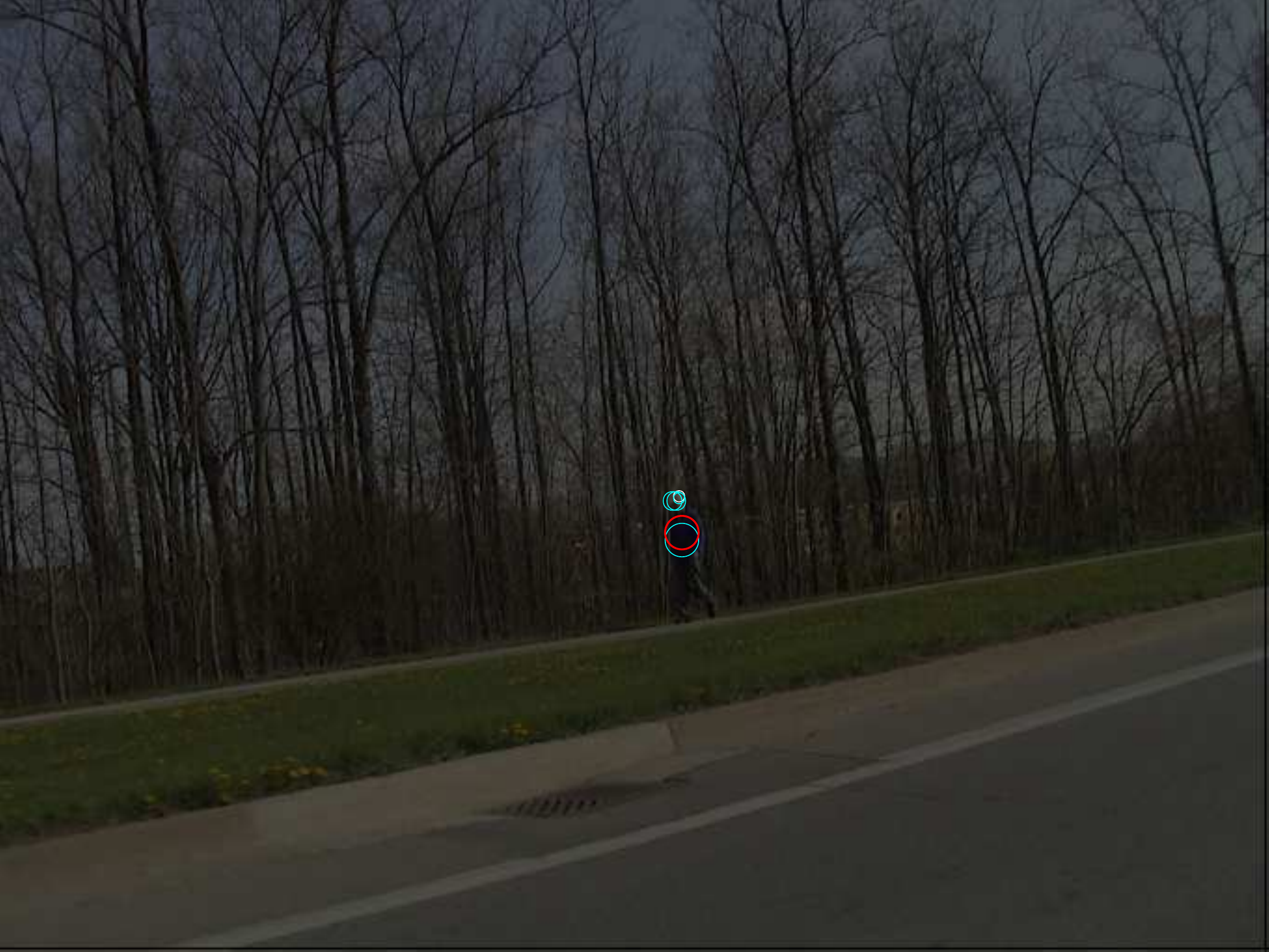}
\includegraphics[height=0.116\textheight]{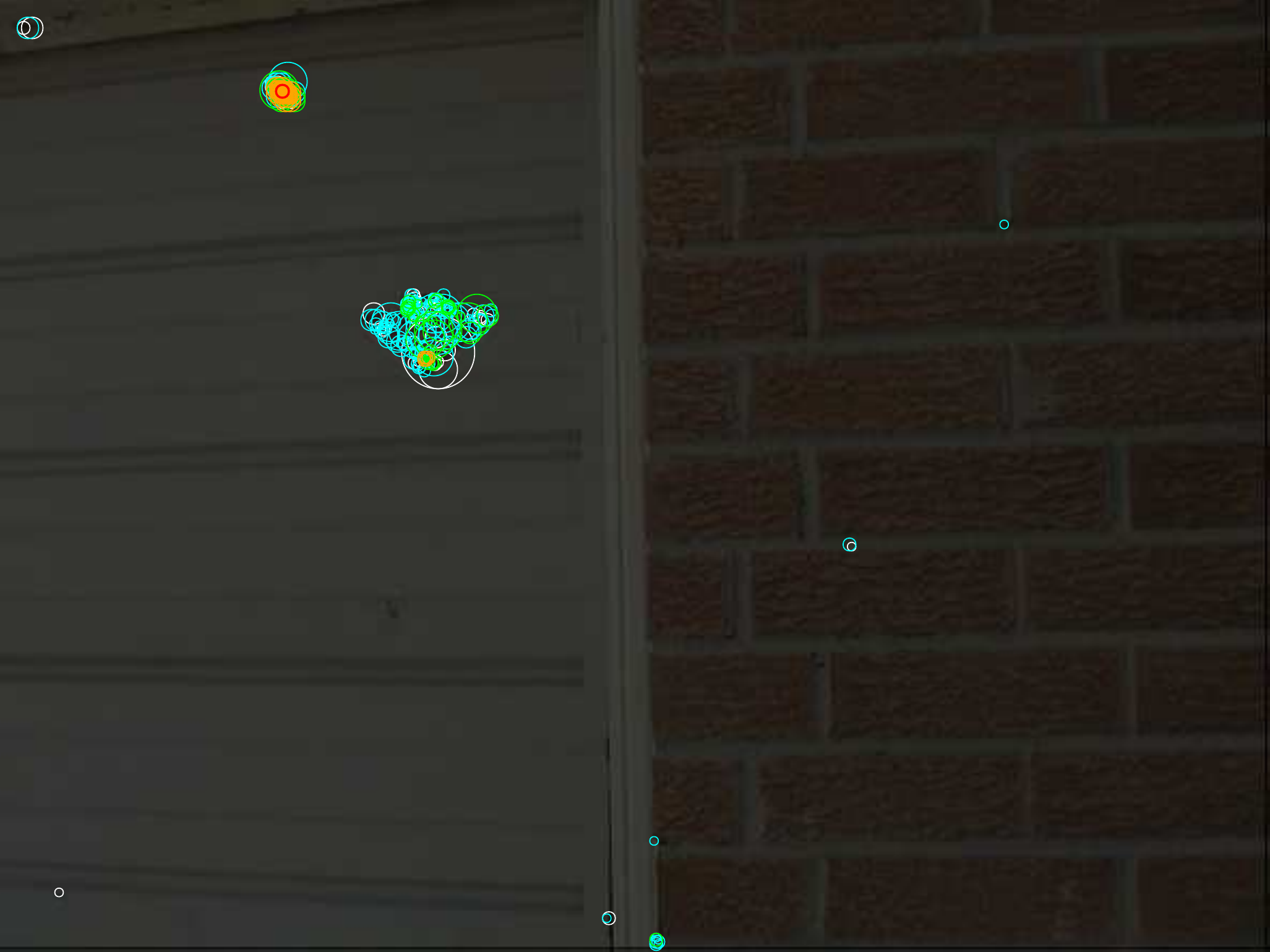}

\end{minipage}

\begin{minipage}[c]{\textwidth}
\centering
\rot{\hspace{2em}\vphantom{A}\vphantom{[}\texttt{Aiger}~\cite{aiger2010phase}}
\includegraphics[height=0.116\textheight]{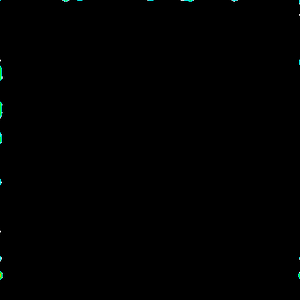}
\includegraphics[height=0.116\textheight]{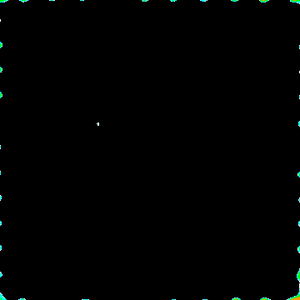}
\includegraphics[height=0.116\textheight]{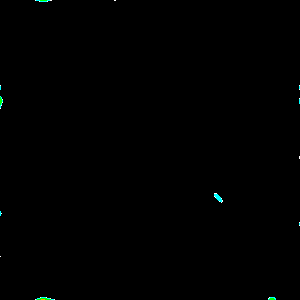}
\includegraphics[height=0.116\textheight]{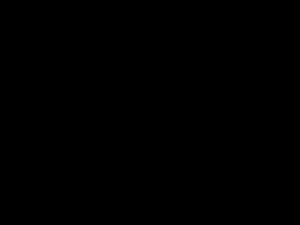}
\includegraphics[height=0.116\textheight]{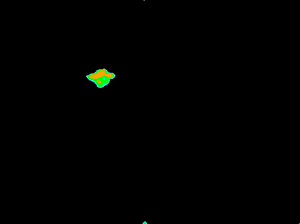}

\end{minipage}

\begin{minipage}[c]{\textwidth}
\centering
\rot{\hspace{2em}\vphantom{A}\vphantom{[}\texttt{Zontak}~\cite{zontak2010defect}}
\includegraphics[height=0.116\textheight]{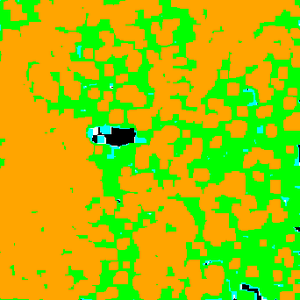}
\includegraphics[height=0.116\textheight]{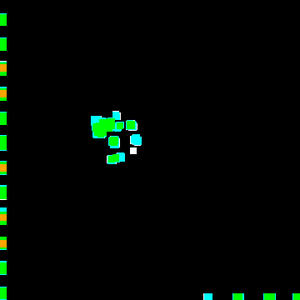}
\includegraphics[height=0.116\textheight]{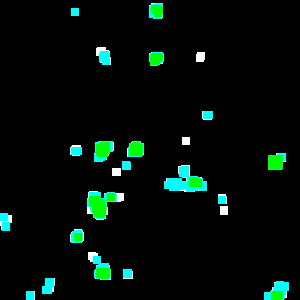}
\includegraphics[height=0.116\textheight]{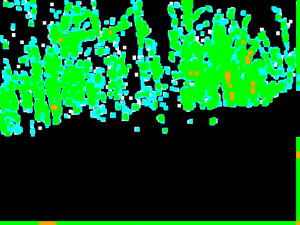}
\includegraphics[height=0.116\textheight]{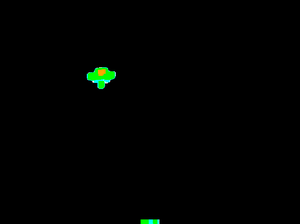}

\end{minipage}

\begin{minipage}[c]{\textwidth}
\centering
\rot{\hspace{2em}\vphantom{A}\vphantom{[}\texttt{Mishne}~\cite{mishne2014multiscale}}
\includegraphics[height=0.116\textheight]{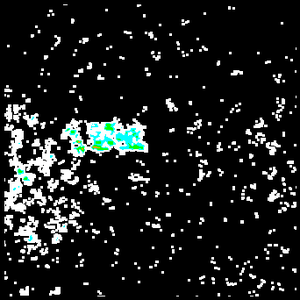}
\includegraphics[height=0.116\textheight]{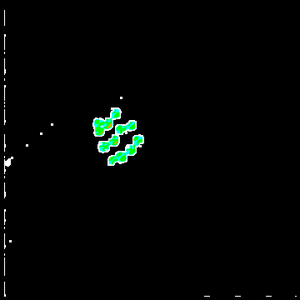}
\includegraphics[height=0.116\textheight]{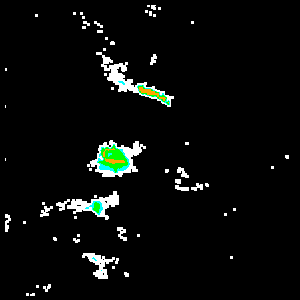}
\includegraphics[height=0.116\textheight]{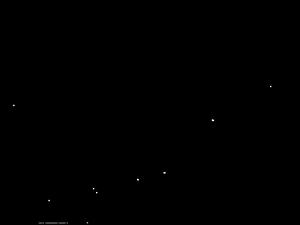}
\includegraphics[height=0.116\textheight]{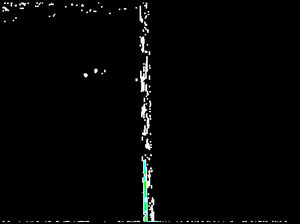}

\end{minipage}

\begin{minipage}[c]{\textwidth}
\centering
\rot{\hspace{2em}\vphantom{A}\vphantom{[}\texttt{Grosjean}~\cite{grosjean2009contrario}}
\includegraphics[height=0.116\textheight]{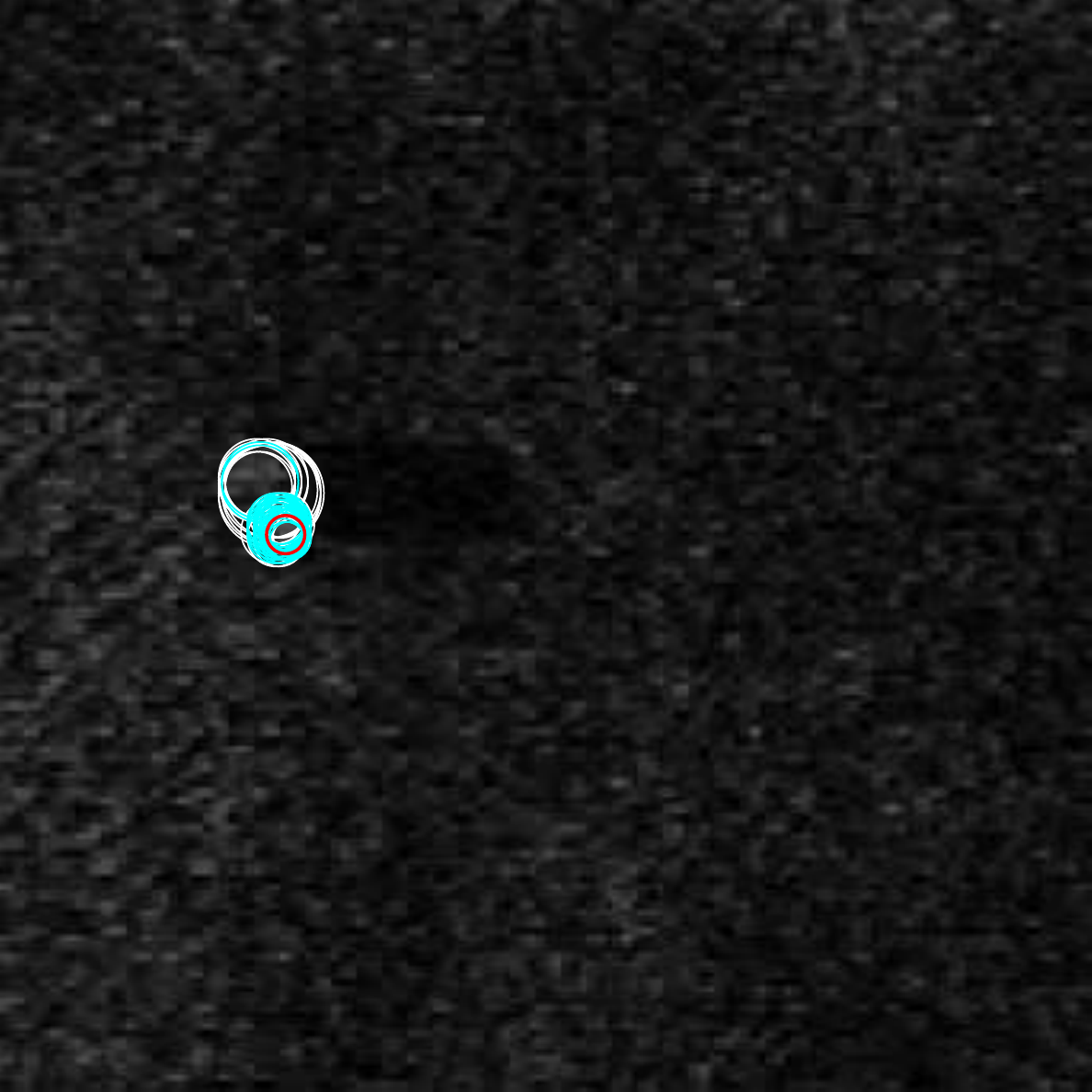}
\includegraphics[height=0.116\textheight]{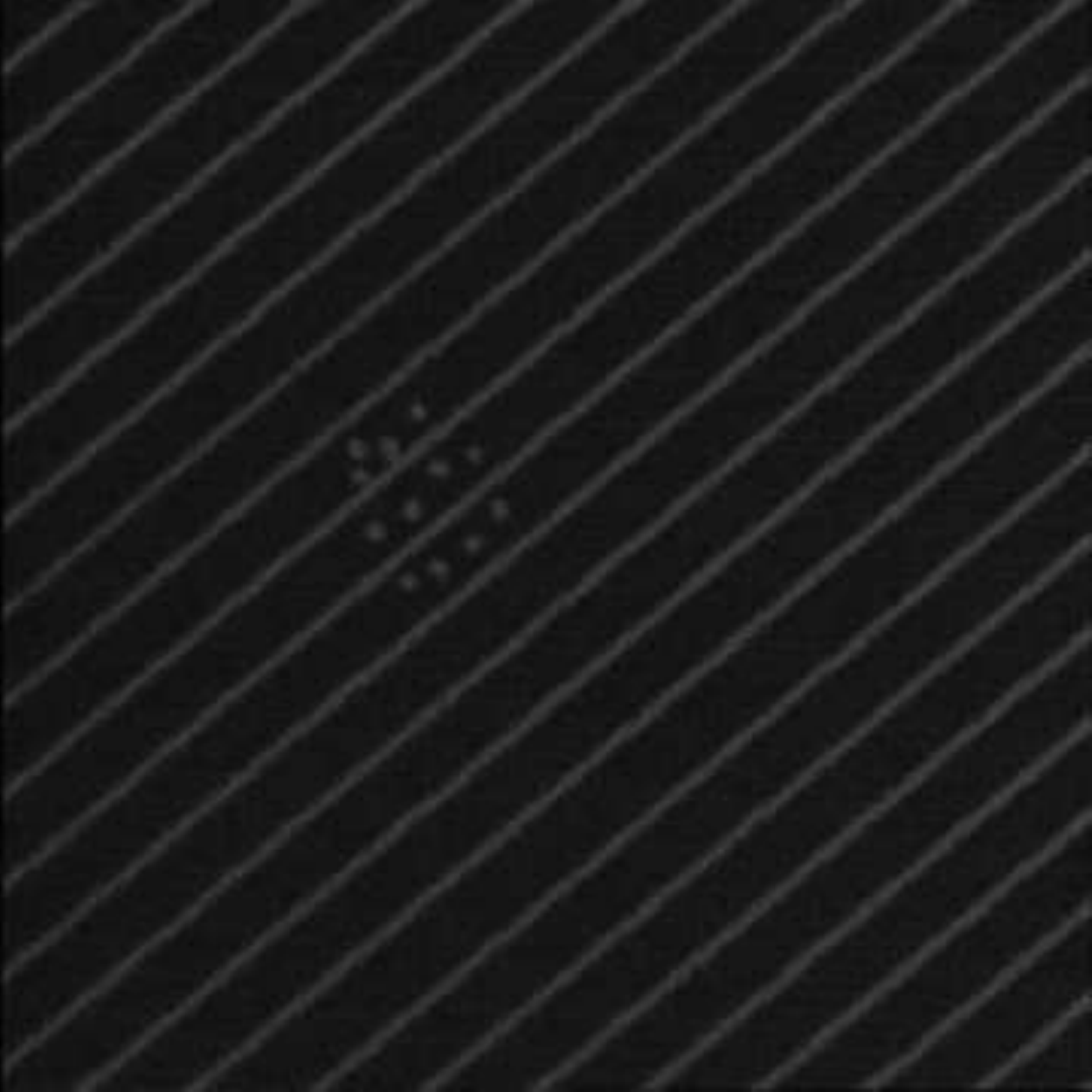}
\includegraphics[height=0.116\textheight]{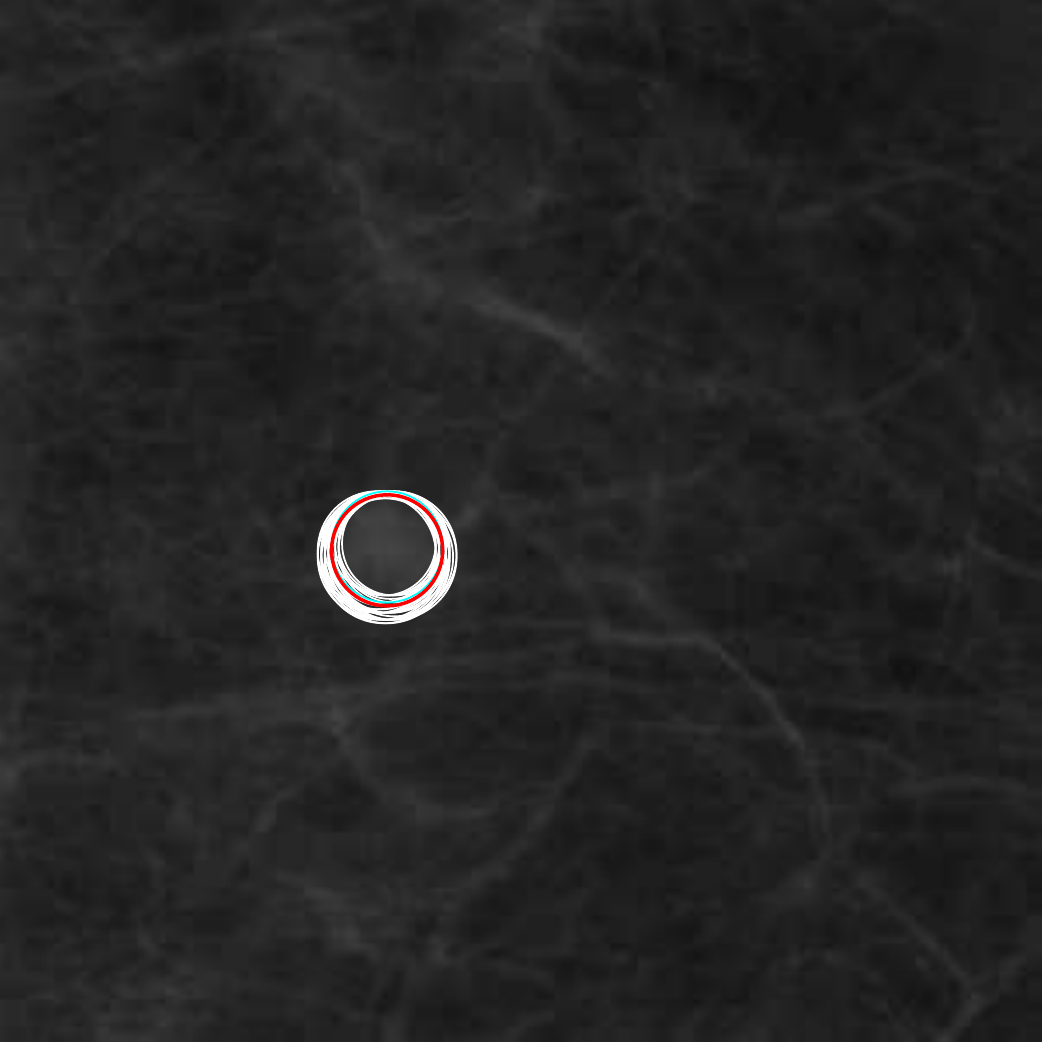}
\includegraphics[height=0.116\textheight]{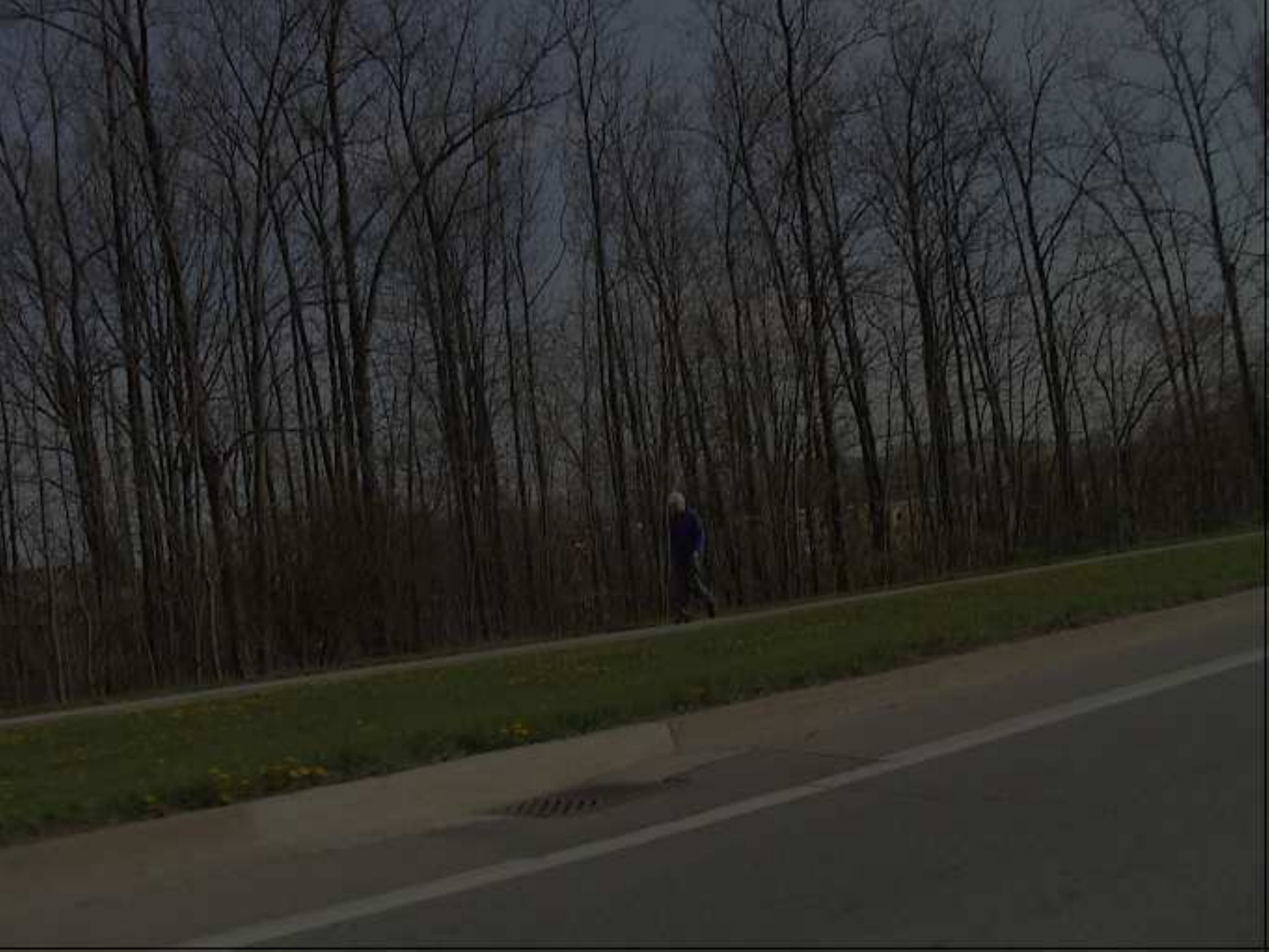}
\includegraphics[height=0.116\textheight]{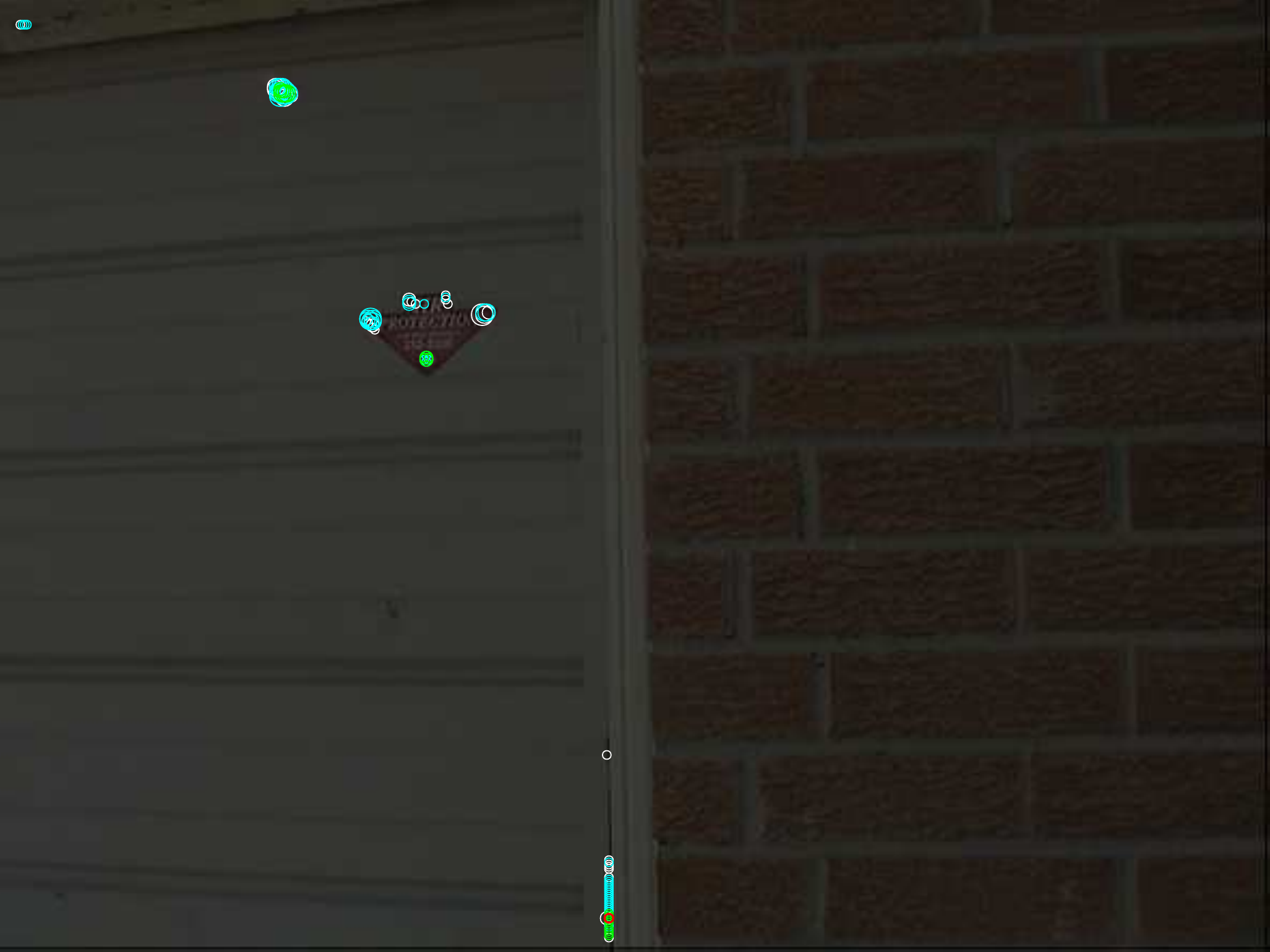}

\end{minipage}

\begin{minipage}[c]{\textwidth}
\centering
\rot{\hspace{1.0em}\vphantom{A}\vphantom{[}\texttt{Boracchi}~\cite{boracchi2014novelty}}
\includegraphics[height=0.116\textheight]{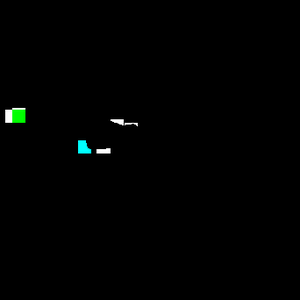}
\includegraphics[height=0.116\textheight]{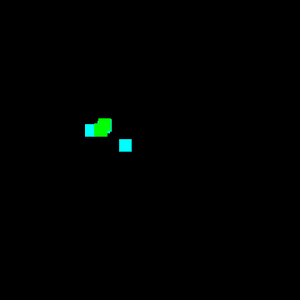}
\includegraphics[height=0.116\textheight]{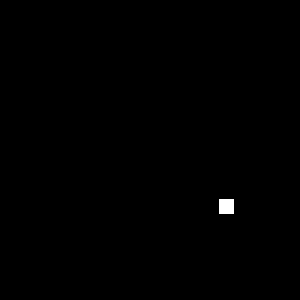}
\includegraphics[height=0.116\textheight]{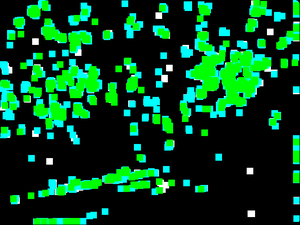}
\includegraphics[height=0.116\textheight]{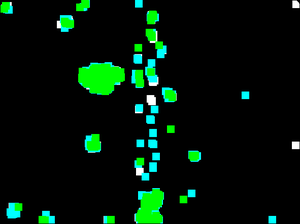}

\end{minipage}

\caption{From left to right: image of an undersea mine from~\cite{mishne2014multiscale}, image of a periodic textile from~\cite{tsai1999automated}, image of a tumor from~\cite{grosjean2009contrario}, image of a man from the Toronto dataset~\cite{bruce2006saliency}, image of a garage door from~\cite{bruce2006saliency}. From top to bottom: The original image, the image residual of one of the scales computed in~\cite{davy2018reducing} (the scale shown is the one where the anomaly is the most salient, the contrast has been adjusted for visualization purpose), algorithm detections for:
\cite{davy2018reducing}, \cite{aiger2010phase}, \cite{zontak2010defect}, \cite{mishne2014multiscale}, \cite{grosjean2009contrario} and \cite{boracchi2014novelty}.  Detections are shown using the following color coding: white is a weak detection - threshold with NFA $\in [10^{-3}, 10^{-2}]$, cyan is a mild detection - threshold with NFA $\in [10^{-8}, 10^{-3}]$, green is a strong detection - threshold with NFA $\in [10^{-21}, 10^{-8}]$, and orange is very strong  - threshold with NFA $\le 10^{-21}$. When available red is the detection with the threshold corresponding to the lowest NFA. For~\cite{mishne2014multiscale} we adopted a similar color coding: white between $0$ and $0.5$, cyan between $0.5$ and $0.7$, green between $0.7$ and $0.9$ and orange above $0.9$.}
\label{fig:results_2}
\end{figure*}

We then analyze three examples coming from previous papers. The first one (first column in Figure \ref{fig:results_2}) is a radar image of an undersea mine borrowed from \citet{mishne2014multiscale}. The mine is  detected by \citet{davy2018reducing,grosjean2009contrario} without any false detections. Both \citet{mishne2014multiscale,boracchi2014novelty} have false detections; \citet{zontak2010defect} over-detects and \citet{aiger2010phase} misses  the mine. The second example (second column in Figure \ref{fig:results_1}) shows an example of near-periodic texture. This is one of the examples where Fourier based methods are ideally well suited. It  was therefore important to check if more generic methods were still able to detect the anomaly. Two methods \citet{aiger2010phase} and \citet{grosjean2009contrario} fail to detect the anomaly, the other three methods performing really well. This makes the case for self-similarity and sparsity based methods,  that generalize nicely the background's periodicity assumption. The final example (third column from Figure \ref{fig:results_2}) is a real example of medical imaging borrowed from \citet{grosjean2009contrario} where the goal is to detect the tumor (the large white region). 
\citet{aiger2010phase,boracchi2014novelty} fail to detect the tumor. A strong detection is given by \citet{zontak2010defect, mishne2014multiscale} but the false alarms are also  strong and  numerous. Finally \citet{davy2018reducing} has stronger tumor detections than \citet{grosjean2009contrario}) (a NFA of $10^{-6.6}$ against $10^{-2.8}$) but it has several false alarms as well.

Finally we tested the  methods on real photographs taken from the Toronto dataset \cite{bruce2006saliency}. This  clearly takes several of the methods out of their specific context and type  of  images (tumors in  X-ray images, mine detection in  sonar scans, clot  detection in microfibers, wafer defects,...) On the other hand, the  principles of the  algorithms are general. So by testing on these examples, our goal is to explore the  limits of several detection  principles,  not to compare these specific algorithms.  
Clearly some of the methods are  more adapted for spatially homogeneous background than to an outdoor cluttered scene.

Another issue  when using real photographs is that anomalies detected by humans may be semantic. None of the   methods we consider was made to detect semantic anomalies,  that can only be learned on  human  annotated images. Nevertheless, the tests' results are  still enlightening. Detections are very different from one method to the other. The fourth example in Figure \ref{fig:results_2} shows a man walking in front of some trees. \citet{aiger2010phase,grosjean2009contrario} and \citet{mishne2014multiscale} don't detect anything. Both \citet{zontak2010defect,boracchi2014novelty} detect mostly the trees and the transition between the road and the sidewalk. Surprisingly \citet{davy2018reducing} only detects the man. Indeed in the noise like residual one can check  that  the man stands out.  The second example shows a garage door as well as a brick wall. This time the algorithms tend to agree more. The conspicuous sign on the door is well detected by all methods as well the lens flare. A gap at the bottom between the brick wall and the door is detected by \citet{davy2018reducing,mishne2014multiscale,grosjean2009contrario,boracchi2014novelty}. The methods \citet{mishne2014multiscale} and \citet{boracchi2014novelty} also detect the transition between the wall and and the brick wall. Finally some detections on the brick wall are made by \citet{davy2018reducing} and \citet{boracchi2014novelty}.  The  residuals of  \citet{davy2018reducing} on the  second row are  much closer to noise than the background,  which amply justify the  interest  of detecting  on  the residual  rather  than on the background. Nevertheless,  the  residual has  no reason to be uniform, as is apparent in the garage's residual.  Even if the detections look any way acceptable,  this non-uniformity of the residual noise suggests that center-surround detectors based on a local variance  (as done in \cite{grosjean2009contrario})  might eventually be preferable.

Fixing  a target number of $10^{-2}$ for the NFA means that  under the $(\mathcal{H}_0)$ model, only $10^{-2}$ false positives should occur per image.  Yet, many of them shown examples show several false positives. Given the mathematical justification of these thresholds, false positives come from discrepancies between the hypothetical $(\mathcal{H}_0)$ model and the image. In the case of \citet{zontak2010defect}, the over-detection in the trees of the picture with a man can be explained by the limited self-similarity of the trees: for this region, the nearest patches won't be close enough to the patch to reconstruct to fit the model, which requires at least one would-be-identical-except-for-the-noise patch in the neighborhood. The over-detection in the case of the undersea mine is likely a mismatch of the noise model with the picture noise. The many false alarms of this method for the other examples, makes us wonder if the model hypothesis is not too strong. The  \citet{boracchi2014novelty} method triggers many  false detections in almost all examples tested.  As we mentioned,  this  suggests that the Gaussian model for the detection  pairs is inaccurate. This is not necessarily a problem for specific fault detection applications where the false alarm curves can be learned. 

\subsection{Quantitative evaluation}
\label{sec:exp_quantitative}
\begin{figure}[h]
\centering
\includegraphics[width=0.325\columnwidth]{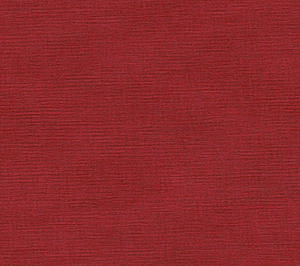}
\includegraphics[width=0.325\columnwidth]{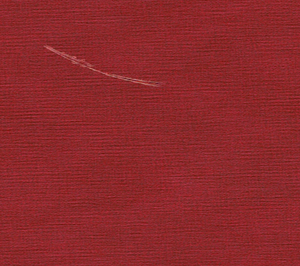}
\includegraphics[width=0.325\columnwidth]{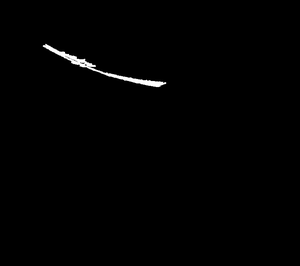}
\caption{A ground truth (on the right) for anomaly detection has been generated by introducing an anomaly in a RPN~\cite{galerne2011random} texture (on the left), which is anomaly free. The detection is then done on the result (in the middle).}
\label{fig:example_gt_anom}
\end{figure}

Estimating how well an anomaly detector  works ``in general'' is a  challenging  evaluation  task. Qualitative experiments such as the ones presented in  section \ref{qualitative evaluation} give no final decision. Our  goal now is to  address the performance  evaluation in terms of true positive rate  (TP) and false positive rate (FP). To that  aim, we  generated a set of ten RPN textures \cite{ipol.2011.ggm_rpn} which are deprived of any statistical anomalies. We then introduced one artificial anomaly per rpn by merging a small piece of another image inside each of them. This was made  by simple blending or by Poisson editing \cite{perez2003poisson} using the implementation of \cite{ipol.2016.163}. This method provides a set of images where a ground truth is known. Hence the detection quality measure can be clearly defined. Figure \ref{fig:example_gt_anom} shows one of the generated RPN images with an anomaly added and the anomaly's ground truth locus. Table \ref{tab:quantitative_results} shows the result for our six methods on this dataset.

\begin{table*}[h]
\setlength{\tabcolsep}{3pt}
\footnotesize
  \centering
  \begin{tabular}{ccccc}
    \toprule
    & TP pixels (in \%) & FP pixels (in \%) & TP anomalies (in \%) & FP anomalies (in \%)\\
    \midrule
    \citet{aiger2010phase} & $56.2$ & $7.60\times 10^{-4}$ & 90 & 40 \\
    \citet{zontak2010defect} & $0$ & $0$ & 0 & 0 \\
    \citet{mishne2014multiscale} & $23.4$ & $8.52$ & 90 & 90 \\
    \citet{boracchi2014novelty} & $78.2$ & $0.87$ & 100 & 100 \\
    \citet{grosjean2009contrario} & $11.6$ & $0.16$ & 30 & 20 \\
    \citet{davy2018reducing} & $33.1$ & $1.79\times 10^{-5}$  & 80 & 10 \\
    \bottomrule
  \end{tabular}
  \caption{Quantitative comparative results for anomaly detection. The number of true positive (TP) and false positive (FP) for different metrics is shown. TP pixels and FP pixels correspond to detections at a pixel level. A true positive is when an anomalous pixel is detected, and a false positive when a normal pixel is detected as anomalous. TP anomalies and FP anomalies evaluate if anomalies have been detected at all.  A true positive is counted when there is at least one detected pixel in an anomalous region, and a false positive when there is at least one detection completely outside an anomalous region (with a maximum of 1 FP per image). These results were computed on a dataset of random uniform textures with a single anomaly added to each image. The thresholds were set for a targer number of false alarms (NFA) of $10^{-2}$ per image (theoretical FP pixels of $4\times10^{-6}\%$).  An example of an image from the dataset is shown in Figure~\ref{fig:example_gt_anom}.  A method works correctly if it detects a high percentage of  anomalies (third column) while having a good pixel coverage (first  column), and a minimal false positive rate (second and fourth columns). Having a very low false positive rate is crucial for massive fault detection. In that sense, the  best methods are~\cite{aiger2010phase} and \cite{davy2018reducing}.} 
  \label{tab:quantitative_results}
\end{table*}
\begin{table*}[h]
\setlength{\tabcolsep}{3pt}
\footnotesize
  \centering
  \begin{tabular}{ccccc}
    \toprule
    & TP pixels (in \%) & FP pixels (in \%) & TP anomalies (in \%) & FP anomalies (in \%)\\
    \midrule
    \citet{aiger2010phase} & $79.1$ & $1.0$ & 100 & 100 \\
    \citet{zontak2010defect} & $27.2$ & $1.0$ & 60 & 60 \\
    \citet{mishne2014multiscale} & $12.5$ & $1.0$ & 50 & 90 \\
    \citet{boracchi2014novelty} & $80.1$ & $1.0$ & 100 & 100 \\
    \citet{grosjean2009contrario} & $24.2$ & $1.0$ & 70 & 100 \\
    \citet{davy2018reducing} & $65.0$ & $1.0$  & 100 & 100 \\
    \bottomrule
  \end{tabular}
  \caption{This table is similar to Table~\ref{tab:quantitative_results},  but in this case each method detection threshold is set so as there are $1\%$ false positives.  Hence,  the  criterion is  to detect as many anomalies as possible (third  column) while having a high true positive rate. The winners are clearly~\cite{boracchi2014novelty} and \cite{aiger2010phase}.  
  } 
  \label{tab:quantitative_results_2}
\end{table*}

Table \ref{tab:quantitative_results} demonstrates that for all methods, the predicted number of false positives (namely the theoretical  NFA) is not always achieved. Indeed, the threshold for Table \ref{tab:quantitative_results} was chosen so that the theoretical number of false detections per image should be $10^{-2}$. When taking into account the total number of pixels, this means that only around $4\times10^{-6}\%$ false detections should be made by any method in this table. Only two methods are close to this number: \cite{aiger2010phase} and \cite{davy2018reducing}, while the other compared methods make too many false detections. Such a false positive target might seem too strict. Yet, it is an important requirement of anomaly detectors in  fault  detection to minimize the  false alarm rate. Indeed excessive false alarms may put a production chain in jeopardy. Images are generally of the order of $10^7$ pixels. Therefore if one wants to limit the false detection rate in a series of tested images, the false positive rate needs to be really small. The methods compared - except \citet{mishne2014multiscale} - used the NFA framework as seen in Section \ref{sec:generalization}. Therefore the discrepancy between the theoretical target and the obtained number of false alarms is explained by an inadequate $(\mathcal{H}_0)$ for the images. In fact, only the background model of \citet{aiger2010phase} matches completely these really specific textures that are RPNs.

\begin{figure*}
\centering
\includegraphics[width=0.15\textwidth]{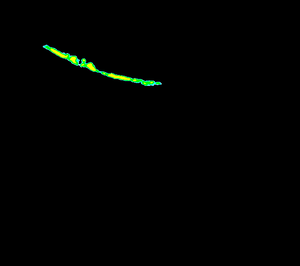}
\includegraphics[width=0.15\textwidth]{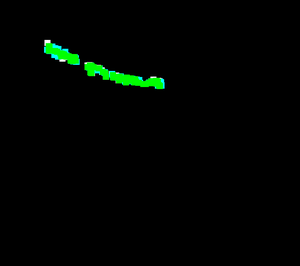}
\includegraphics[width=0.15\textwidth]{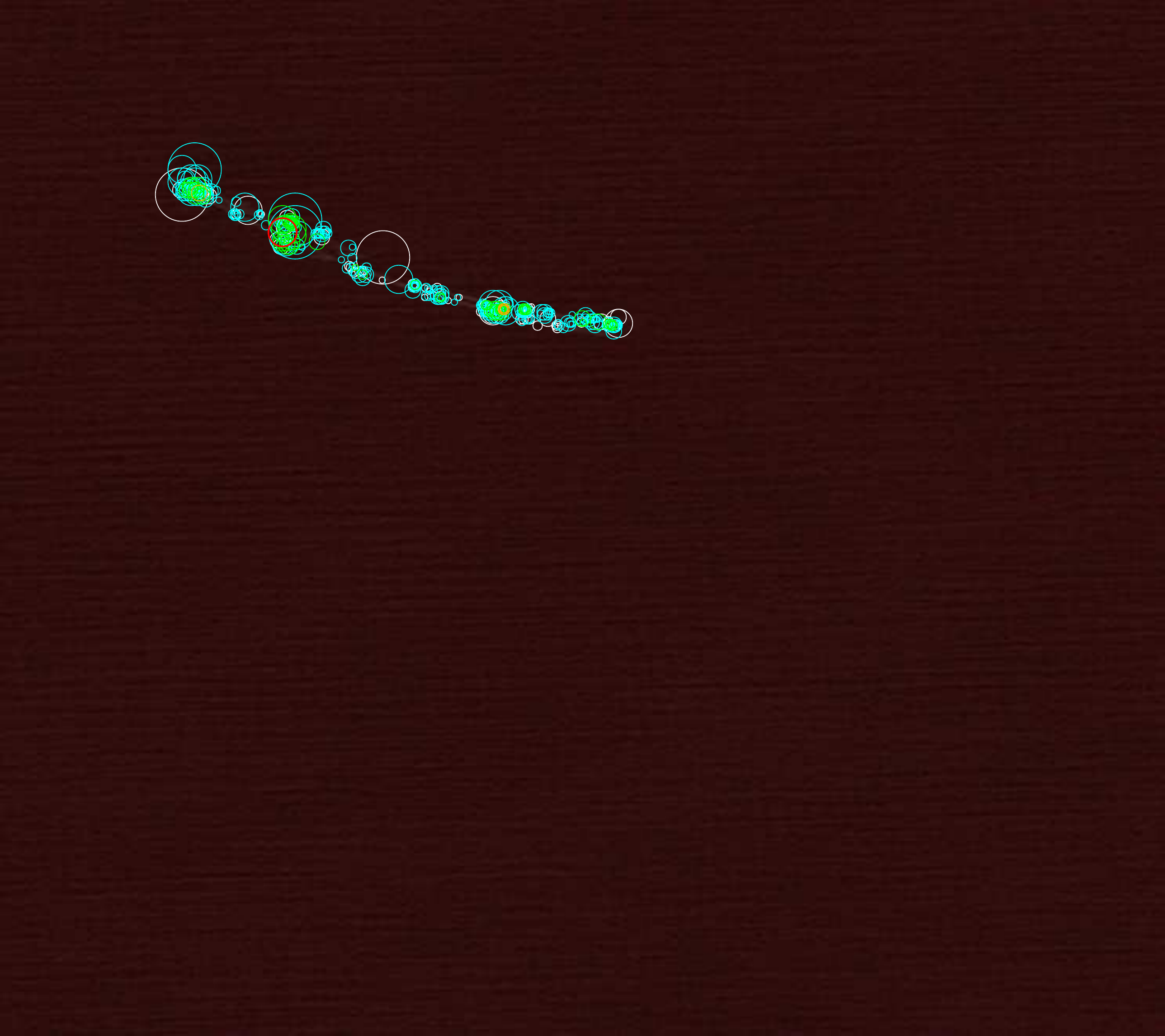}
\includegraphics[width=0.15\textwidth]{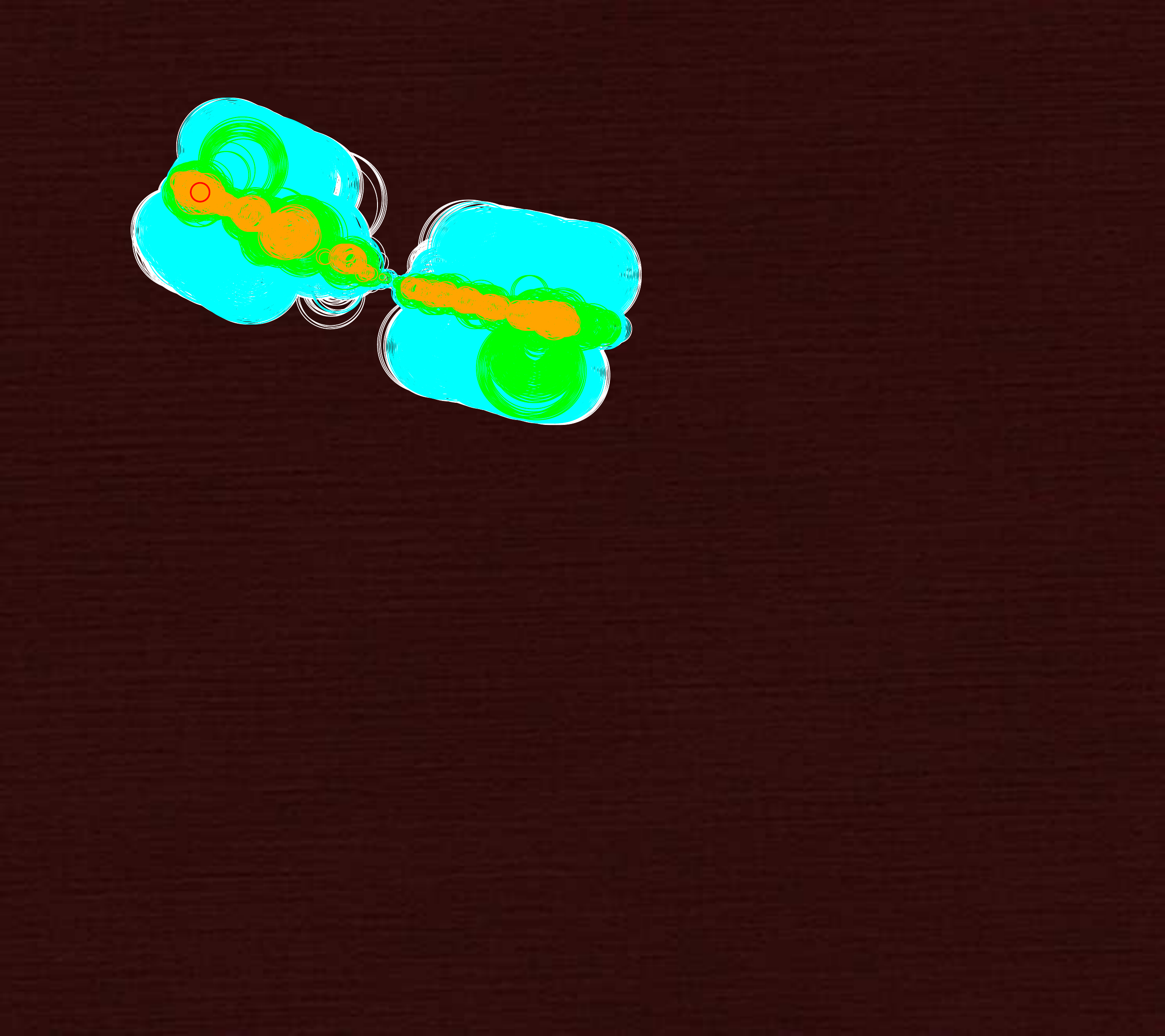}
\includegraphics[width=0.15\textwidth]{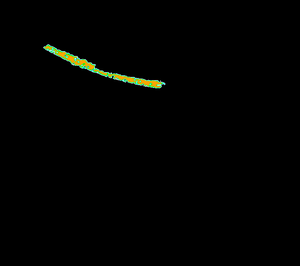}
\includegraphics[width=0.15\textwidth]{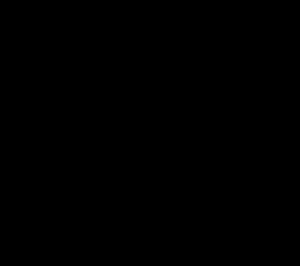}
\caption{Example of detections for all the different methods on \ref{fig:example_gt_anom}. It corresponds to the one showed in Table \ref{tab:quantitative_results}. From left to right: \citet{aiger2010phase}, \citet{boracchi2014exploiting}, \citet{davy2018reducing}, \citet{grosjean2009contrario}, \citet{mishne2017diffusion} and \citet{zontak2010defect}.}
\label{fig:quant_examples}
\end{figure*}

\begin{figure}
\includegraphics[width=\columnwidth]{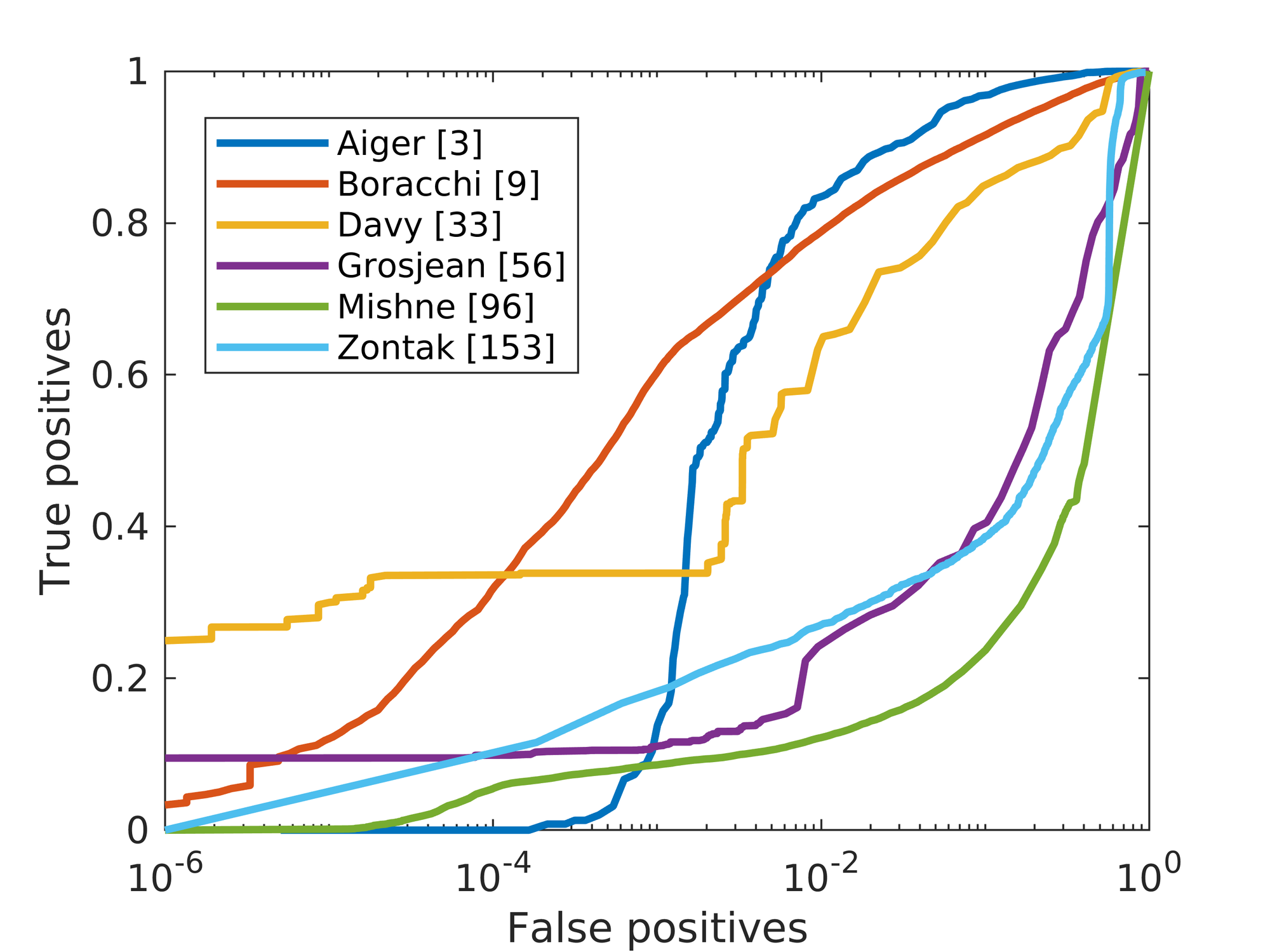}
\caption{ROC curve computed on the dataset of synthetic images. A true positive corresponds to an anomalous pixel detected. A false positive corresponds to a normal pixel that has been detected as anomalous.
In deep blue \citet{aiger2010phase} (Area Under the Curve (AUC) $7.52$), in red \citet{boracchi2014novelty} (AUC $5.79$), in yellow \citet{davy2018reducing} (AUC $7.03$), in purple \citet{grosjean2009contrario} (AUC $3.30$), in green \citet{mishne2014multiscale} (AUC $1.98$) and in light blue \citet{zontak2010defect} (AUC $2.92$).}
\label{fig:roc_curve}
\end{figure}

\begin{figure}
\includegraphics[width=\columnwidth]{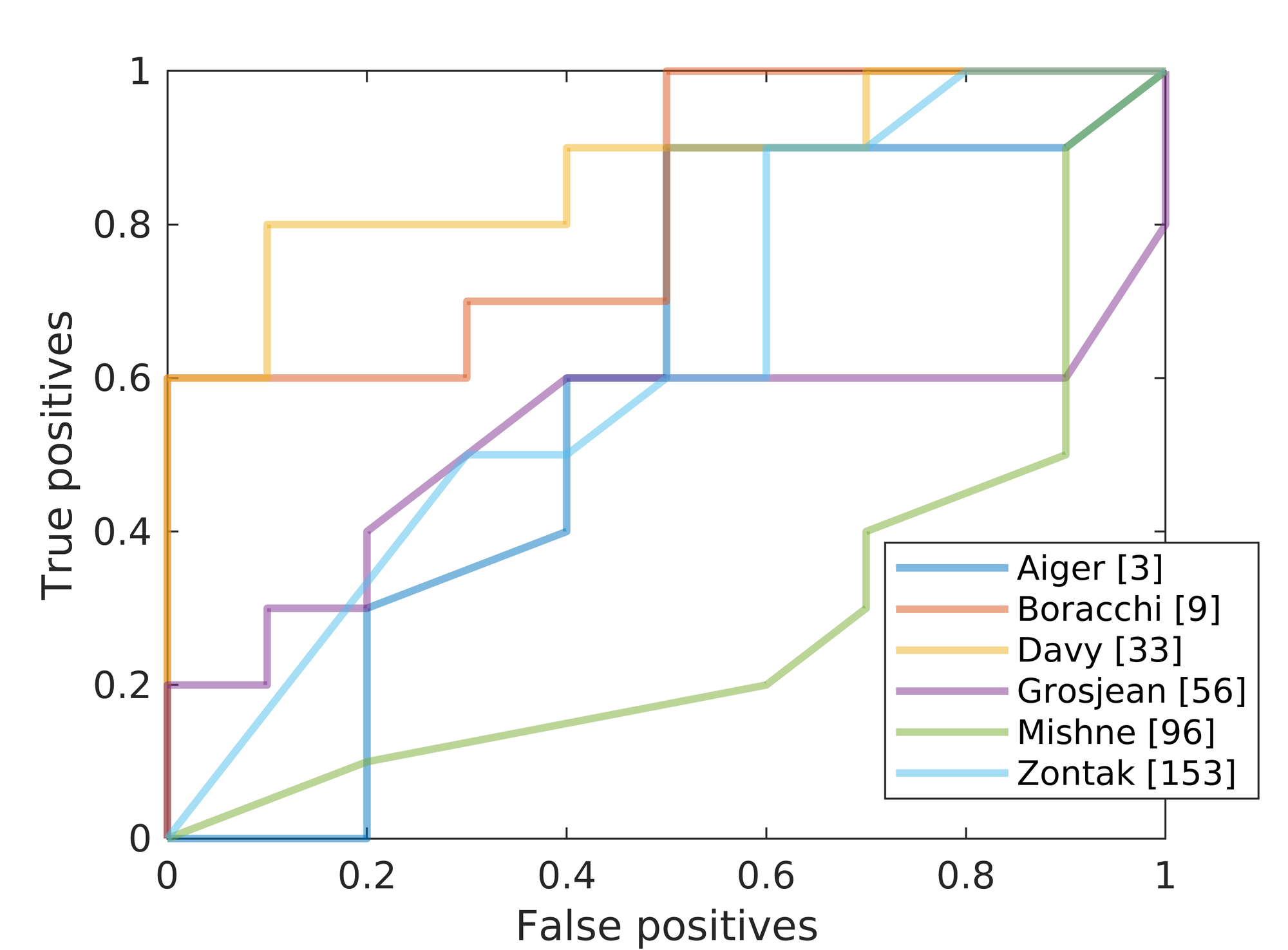}
\caption{ROC curve computed on the dataset of synthetic images. A true positive is when an anomaly is detected (in the sense that at least one detection has been made inside the anomalous region). A false positive is when there is a detection outside the anomalous region. In deep blue \citet{aiger2010phase} (Area Under the Curve (AUC) $0.82$), in red \citet{boracchi2014novelty} (AUC $0.585$), in yellow \citet{davy2018reducing} (AUC $0.87$), in purple \citet{grosjean2009contrario} (AUC $0.52$), in green \citet{mishne2014multiscale} (AUC $0.28$) and in light blue \citet{zontak2010defect} (AUC $0.625$).}
\label{fig:roc_curve_detections}
\end{figure}

To better compare the methods, we also computed ROC curves for all methods, Figures \ref{fig:roc_curve} and \ref{fig:roc_curve_detections}, as well as the table of true positive areas and false positive areas for a fixed positive rate of $1\%$ (Table \ref{tab:quantitative_results_2}). The ROC curve aren't impacted by the choice of thresholds. Figure \ref{fig:roc_curve} is shown with a log scale for the number of false positives because its low or very low false positive section is much more relevant for anomaly detection than the rest. From these ROC curves and tables we can conclude, for this specific example, that \cite{aiger2010phase} (Area under the curve (AUC) 7.52) (which theoretically should be optimal for this problem) performs the best followed closely by \cite{davy2018reducing} (AUC 7.03). It's worth noting that \cite{davy2018reducing} is performing better than \cite{aiger2010phase} for very low false positive region. We then have \cite{boracchi2014novelty} (AUC 5.79). The trailing methods are \cite{grosjean2009contrario} (AUC 3.30), \cite{zontak2010defect} (AUC 2.92) and finally \cite{mishne2014multiscale} (AUC 1.98) . Nevertheless, if a moderate  number of false positives can be tolerated, then \cite{boracchi2014novelty} becomes really attractive because of its high detection precision.  Figure \ref{fig:roc_curve_detections} illustrates the problem of false detections. Most methods requires many false detections to achieve a reasonable detection rate. Only  \citet{aiger2010phase} (AUC 0.82) and \citet{davy2018reducing} (AUC 0.87) detect well while still keeping a zero false detection rate. This confirms the results from Table \ref{tab:quantitative_results}. Table \ref{tab:quantitative_results_2} also shows that having a $1\%$ detection is useful to obtain a good precision but leads to almost all images getting false positives. In practice $1\%$ is too large a tolerance for images. In Figure \ref{fig:quant_examples} we show the result of the detections on \ref{fig:example_gt_anom} corresponding to Table \ref{tab:quantitative_results} for the different methods.


\subsection{Impact of the parameters}

\begin{figure*}
\centering
\footnotesize
\hspace{1em}\vphantom{A}\vphantom{[} $4\times 4$ patches\hspace{6em}$8\times 8$ patches\hspace{5em}$16\times 16$ patches\hspace{5em}$32\times 32$ patches
\begin{minipage}[c]{\textwidth}
\centering
\rot{\hspace{1.5em}\vphantom{A}\vphantom{[} $4$ neighbors}
\includegraphics[height=0.116\textheight]{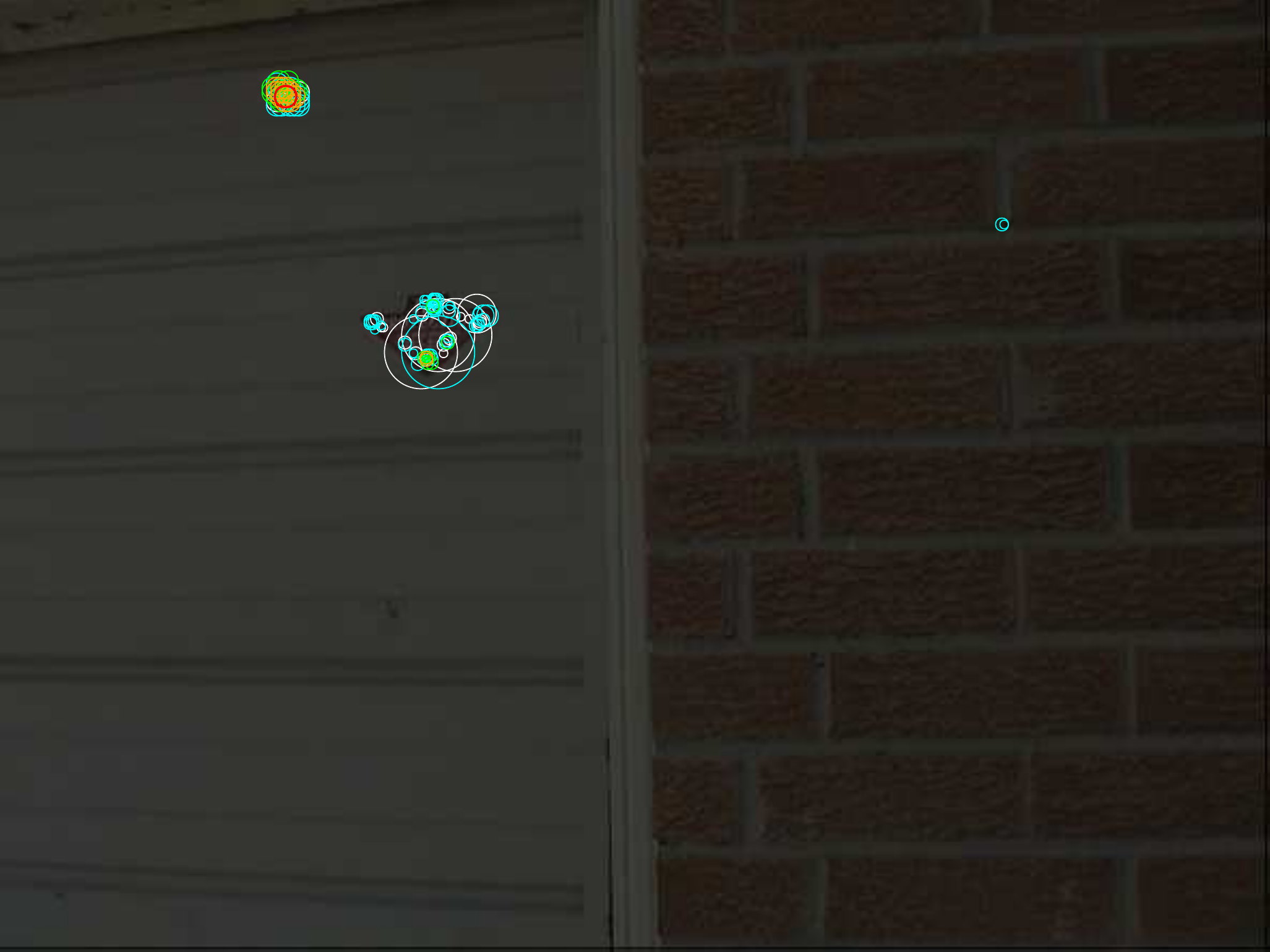}
\includegraphics[height=0.116\textheight]{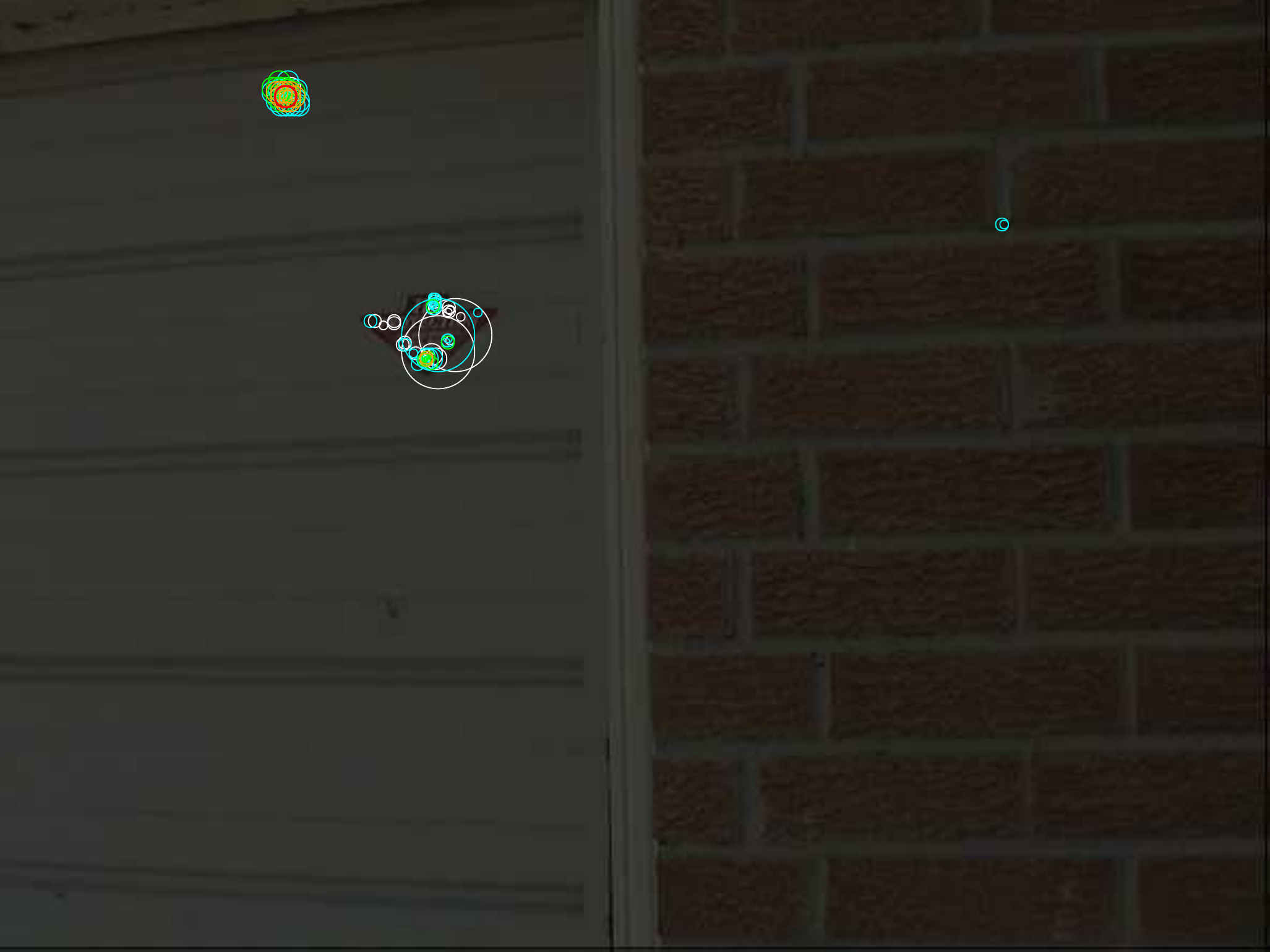}
\includegraphics[height=0.116\textheight]{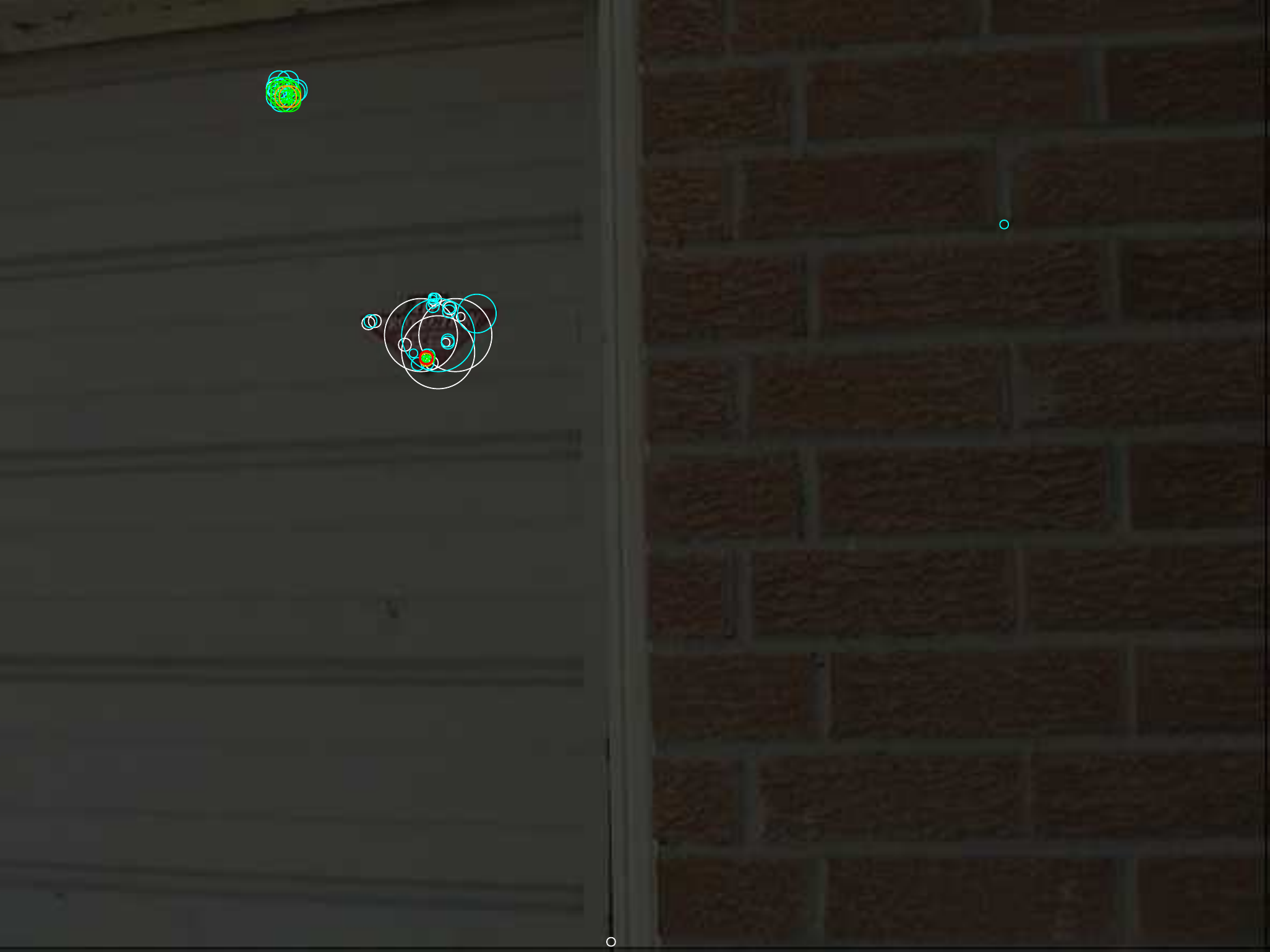}
\includegraphics[height=0.116\textheight]{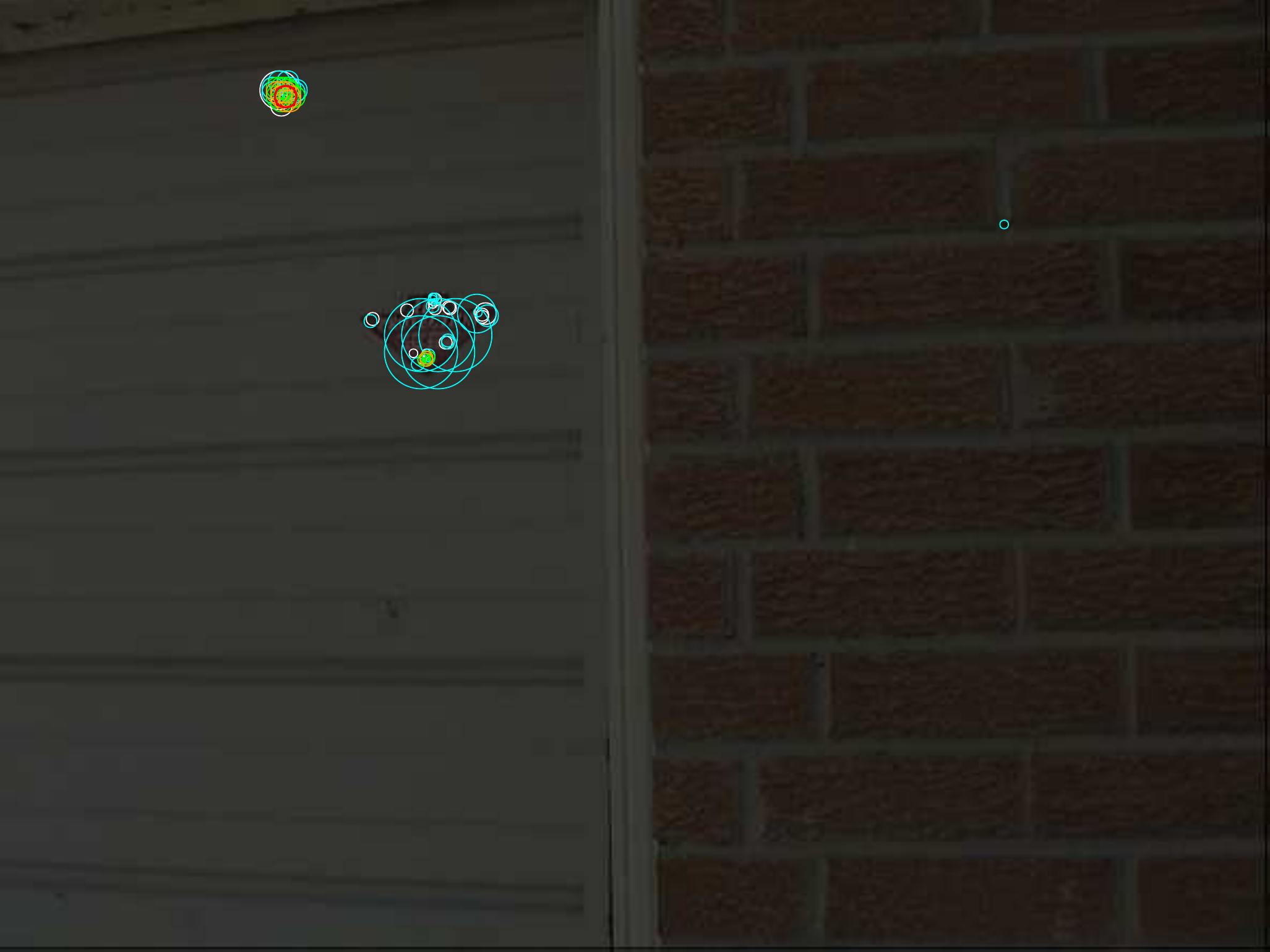}
\end{minipage}

\begin{minipage}[c]{\textwidth}
\centering
\rot{\hspace{1.5em}\vphantom{A}\vphantom{[} $8$ neighbors}
\includegraphics[height=0.116\textheight]{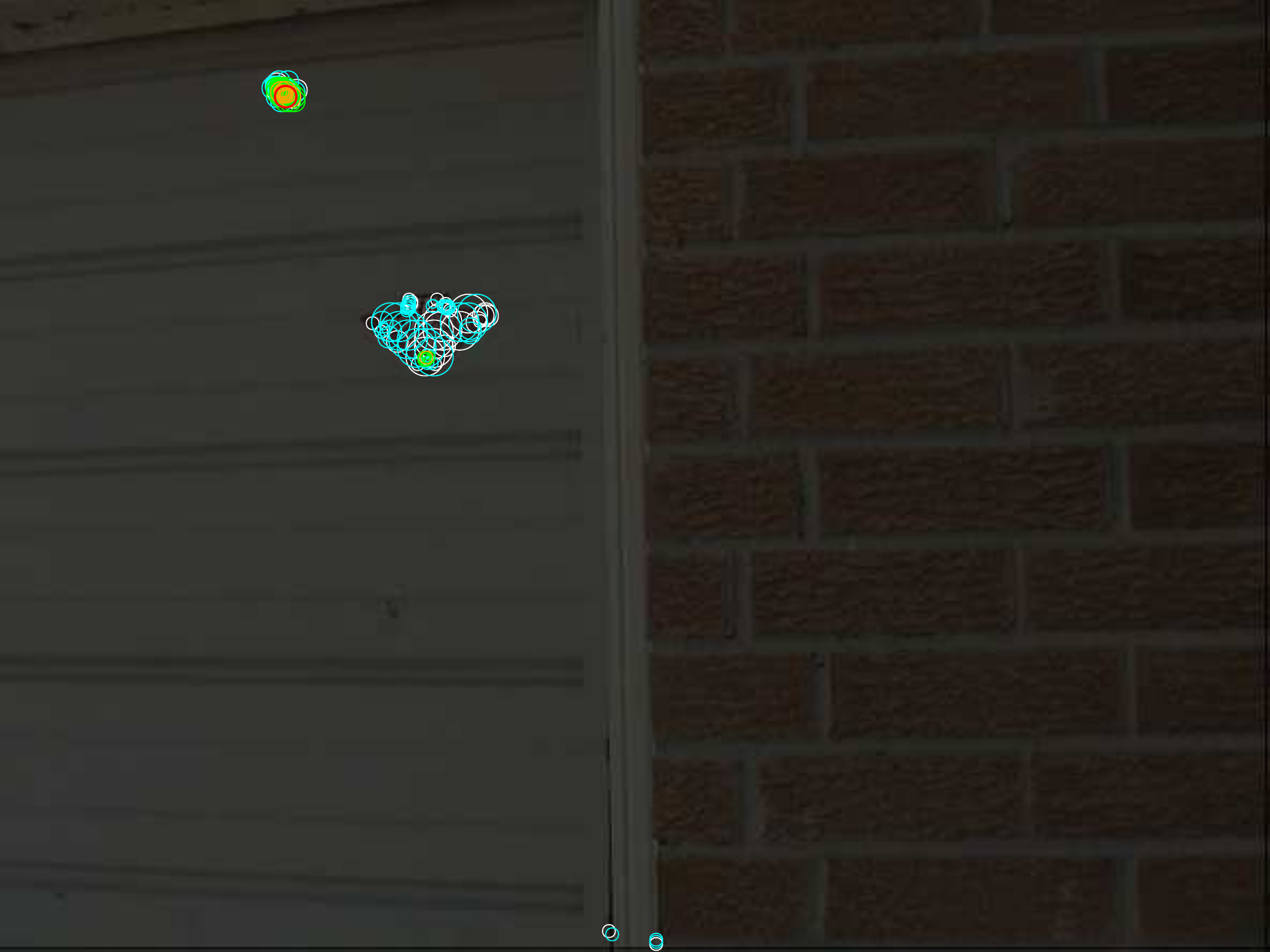}
\includegraphics[height=0.116\textheight]{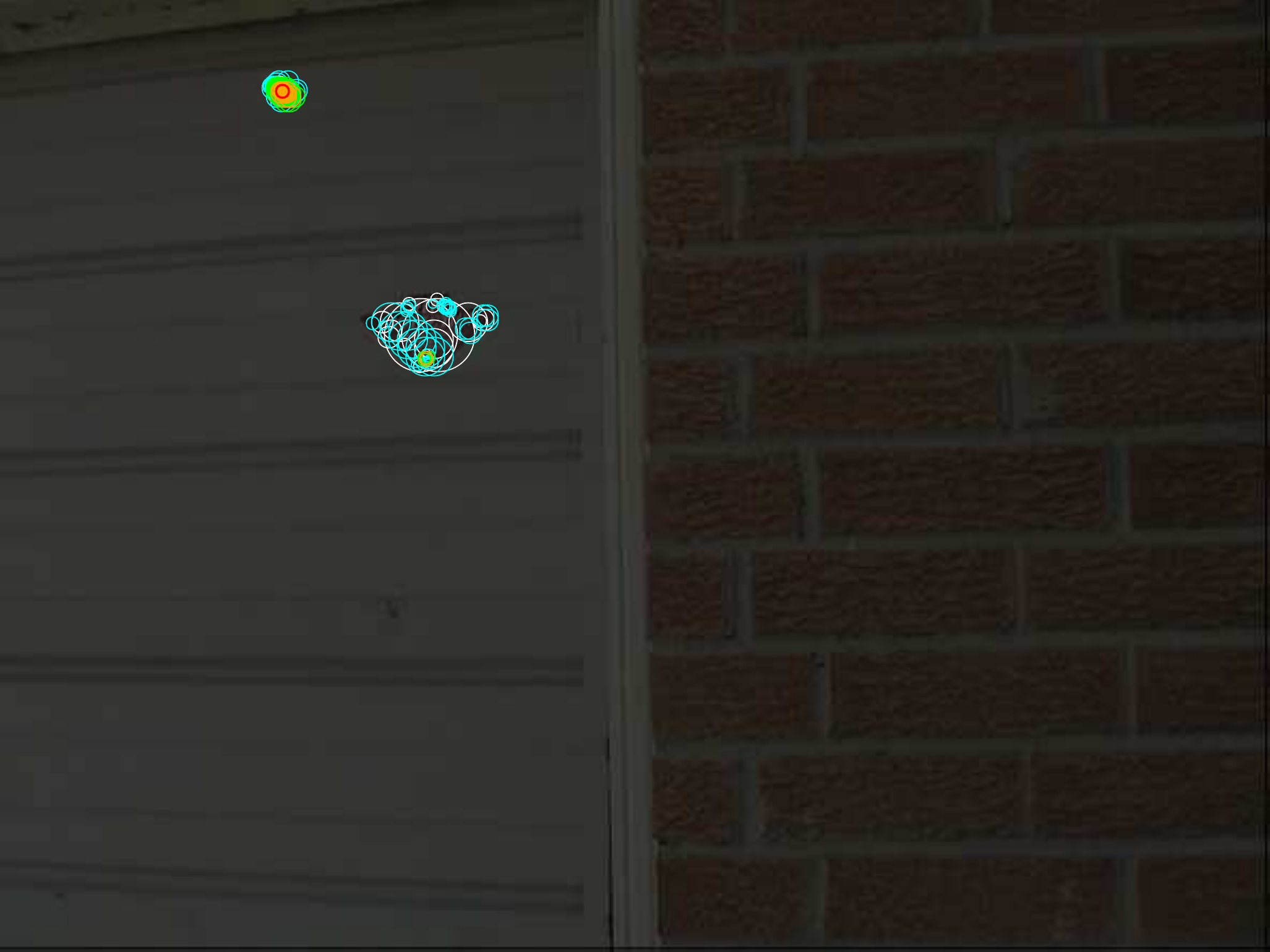}
\includegraphics[height=0.116\textheight]{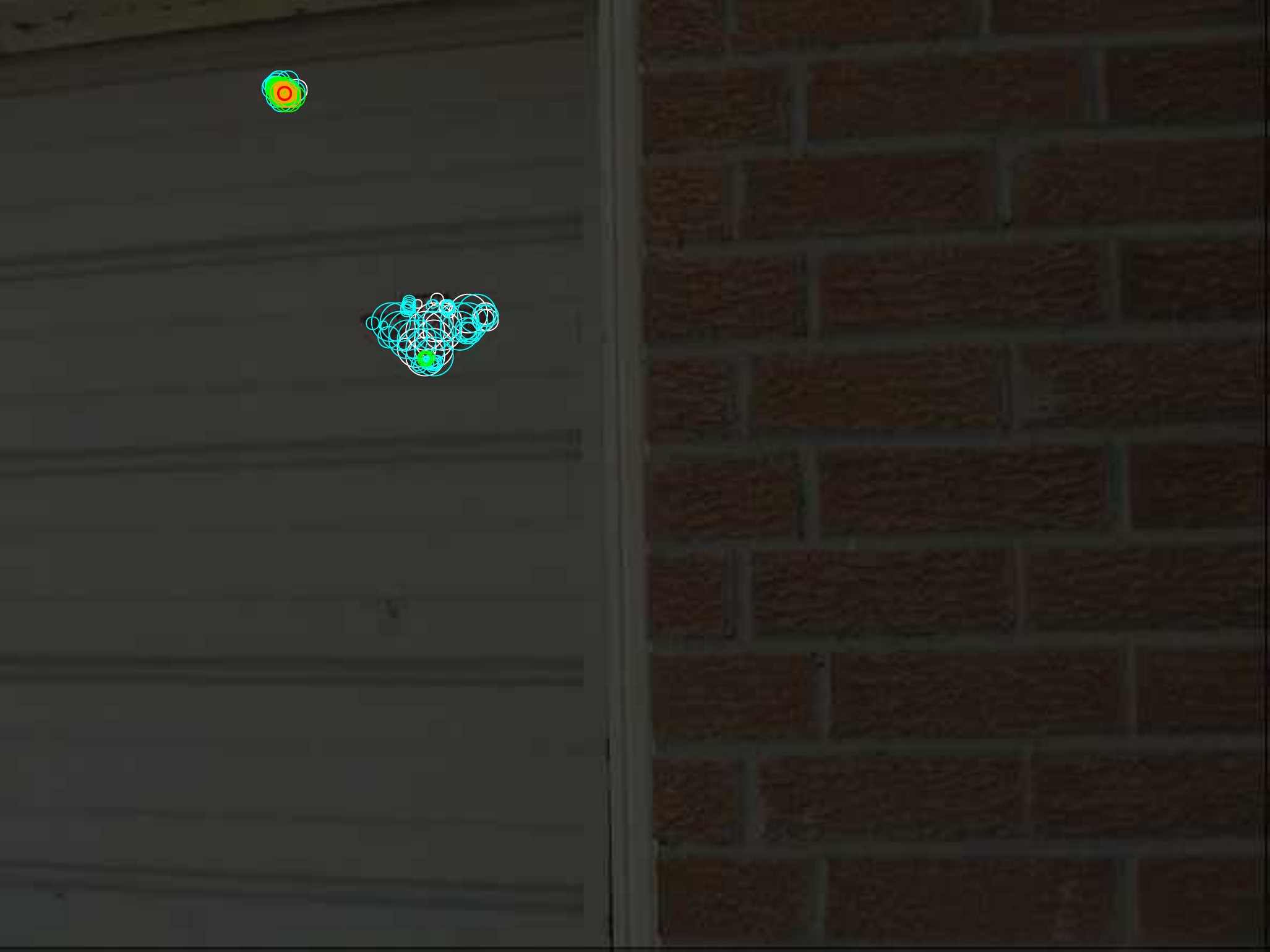}
\includegraphics[height=0.116\textheight]{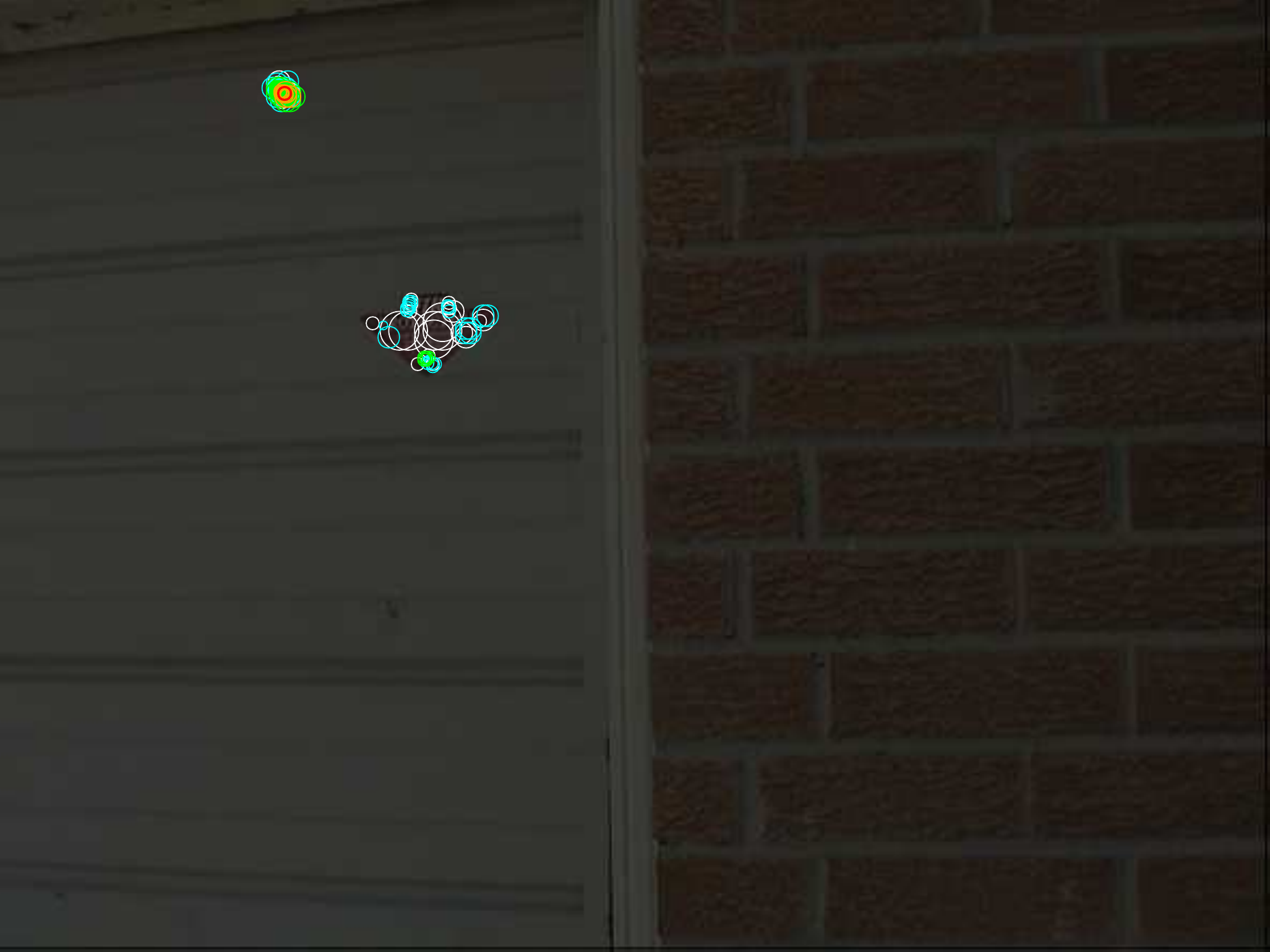}
\end{minipage}

\begin{minipage}[c]{\textwidth}
\centering
\rot{\hspace{1.2em}\vphantom{A}\vphantom{[} $16$ neighbors}
\includegraphics[height=0.116\textheight]{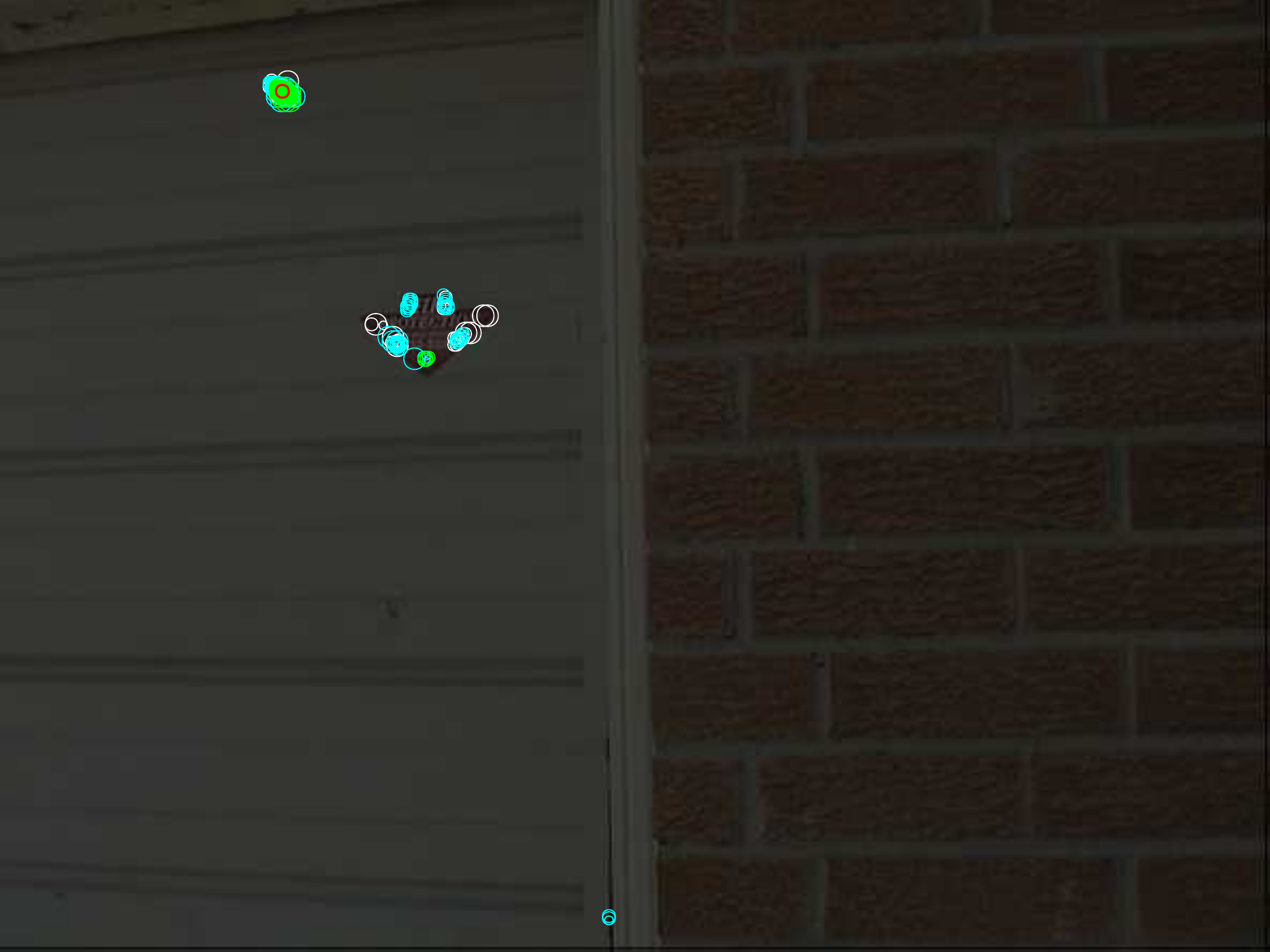}
\includegraphics[height=0.116\textheight]{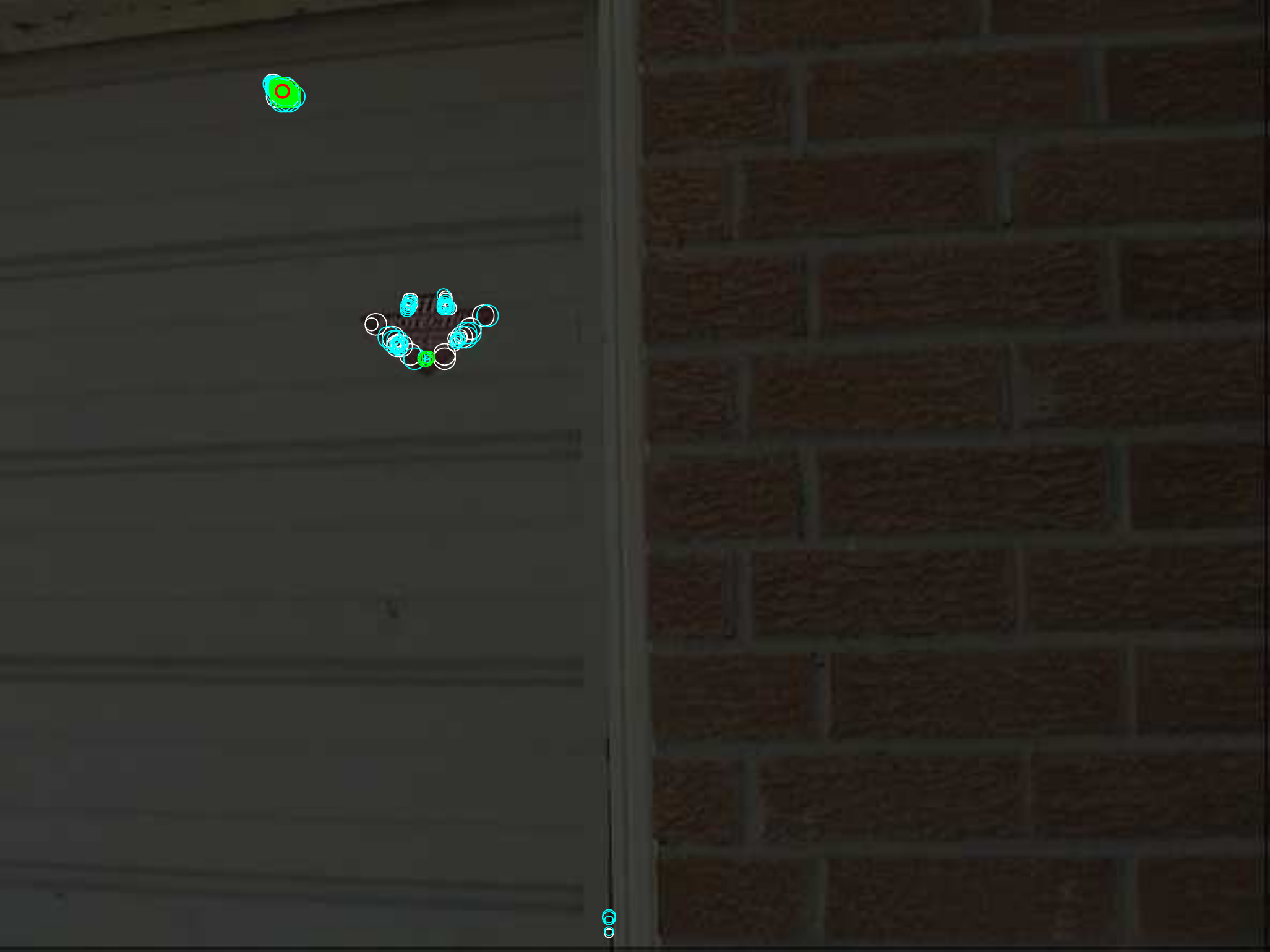}
\includegraphics[height=0.116\textheight]{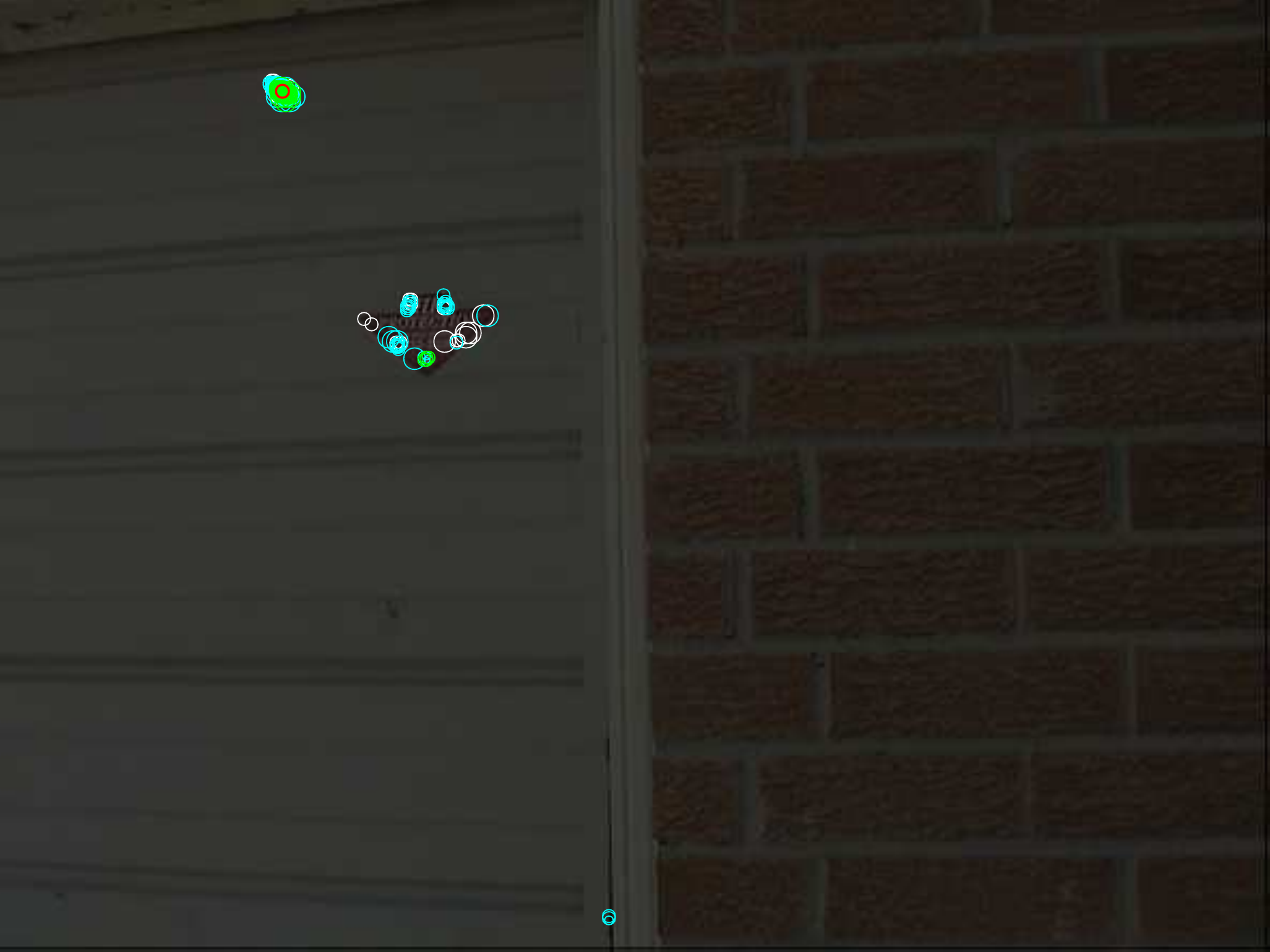}
\includegraphics[height=0.116\textheight]{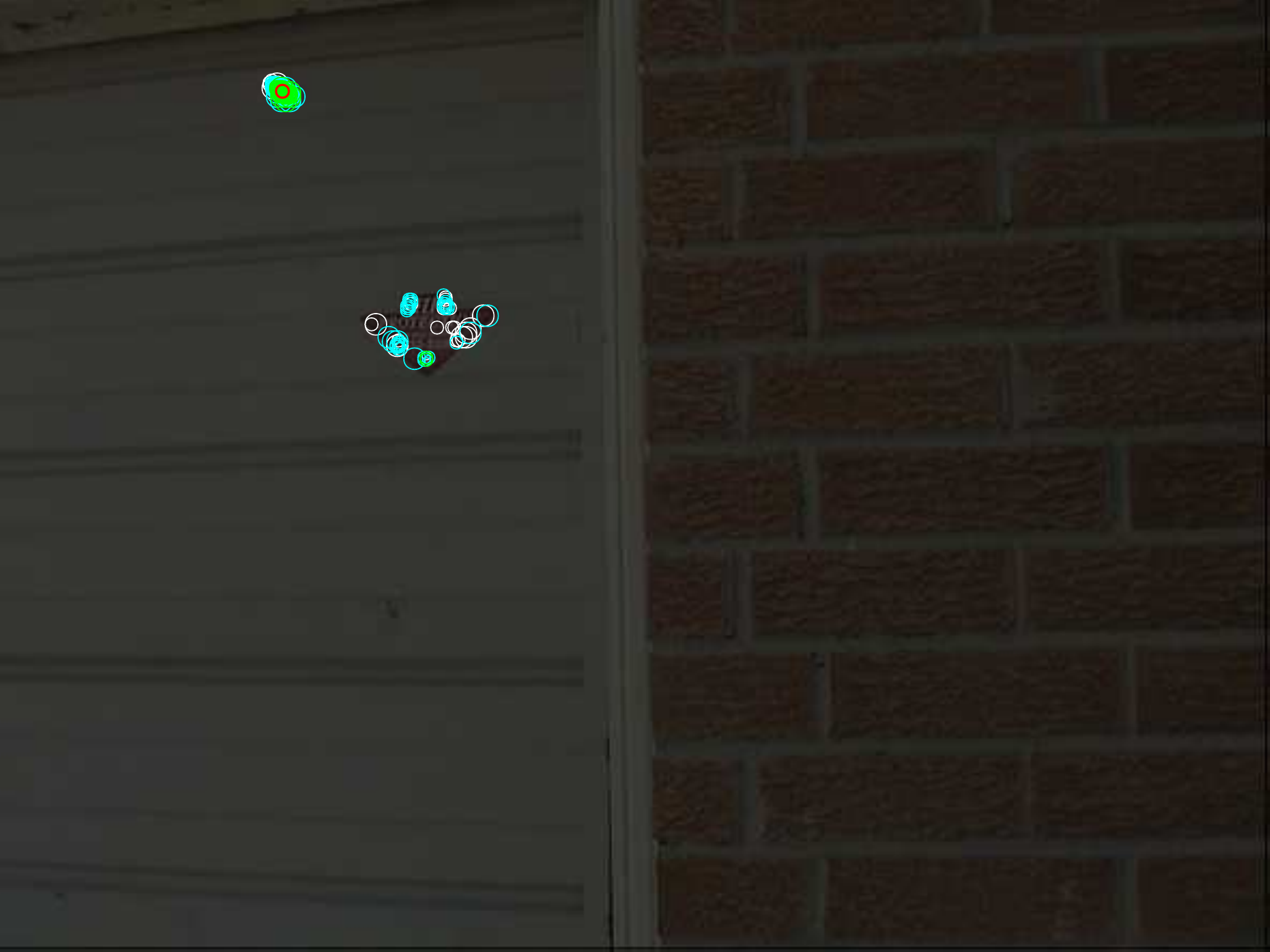}
\end{minipage}

\begin{minipage}[c]{\textwidth}
\centering
\rot{\hspace{1.2em}\vphantom{A}\vphantom{[} $32$ neighbors}
\includegraphics[height=0.116\textheight]{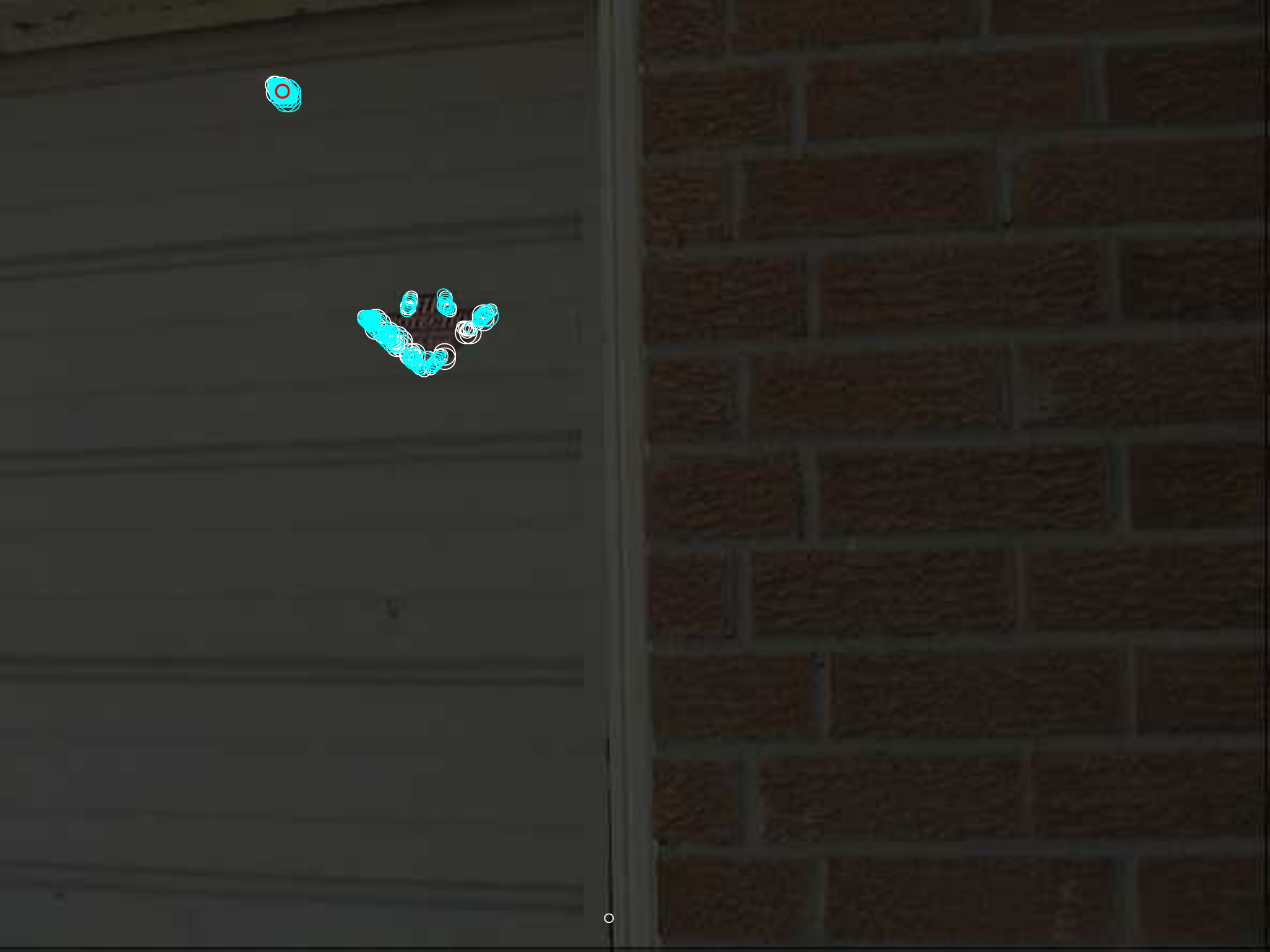}
\includegraphics[height=0.116\textheight]{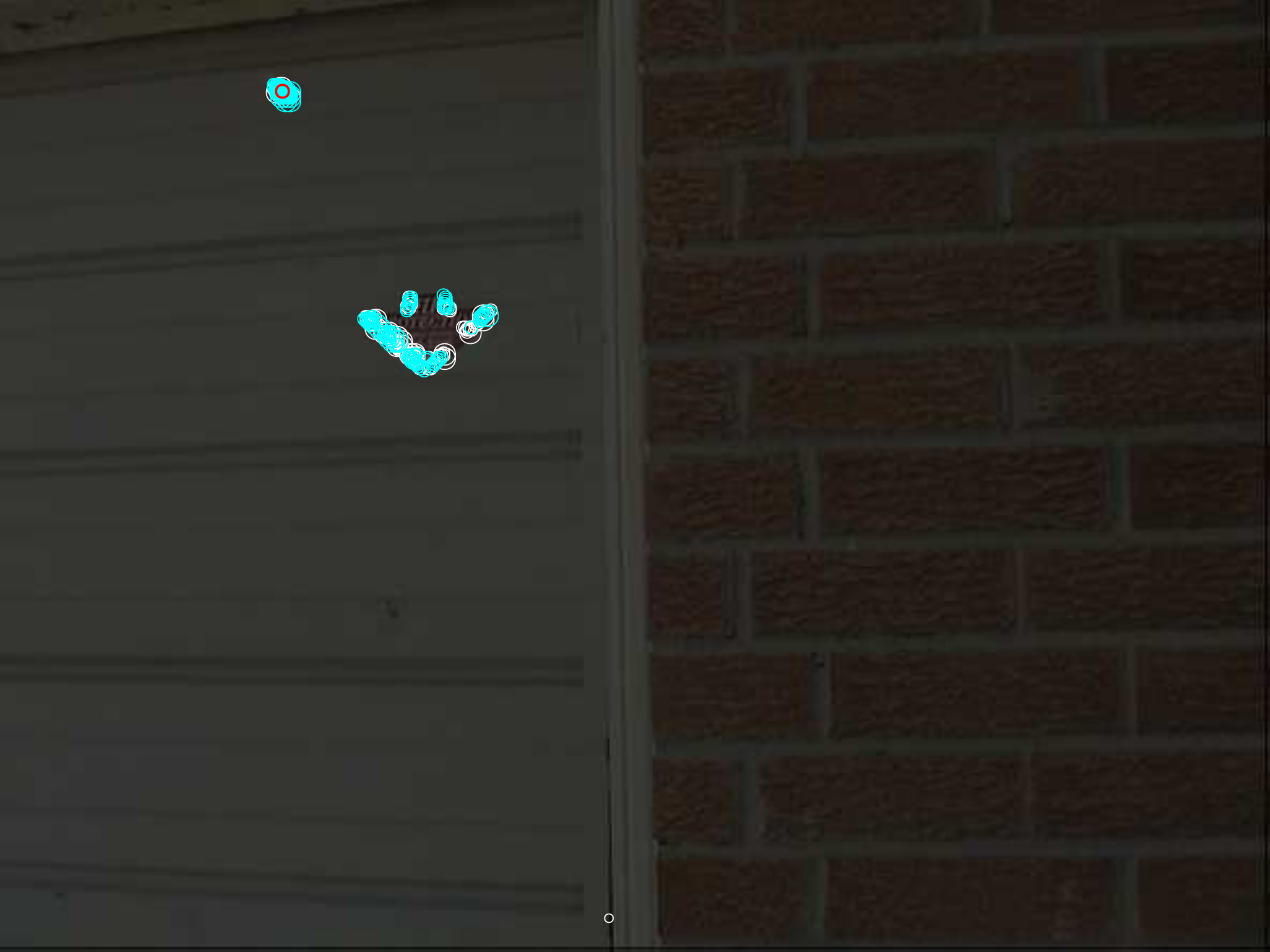}
\includegraphics[height=0.116\textheight]{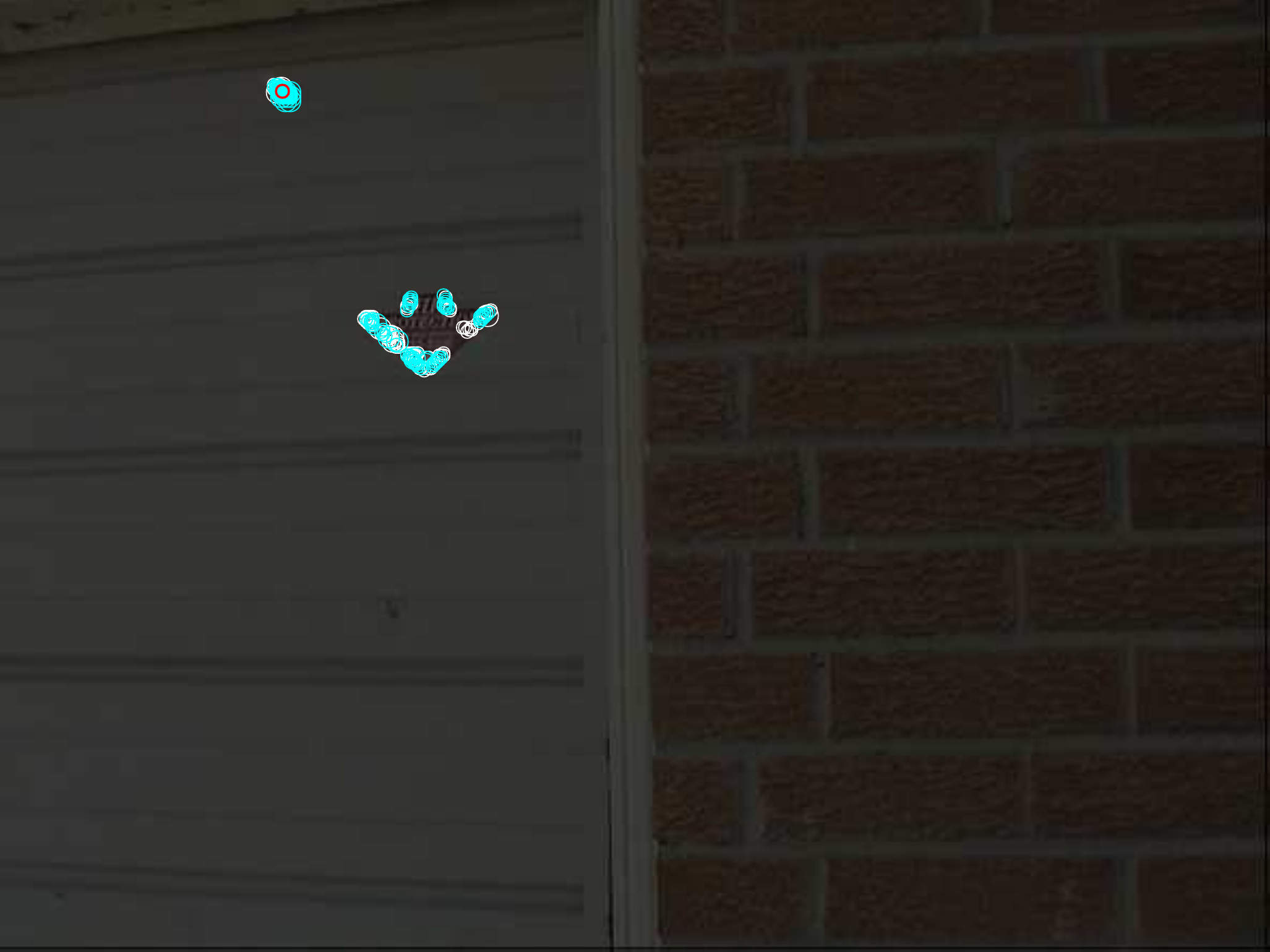}
\includegraphics[height=0.116\textheight]{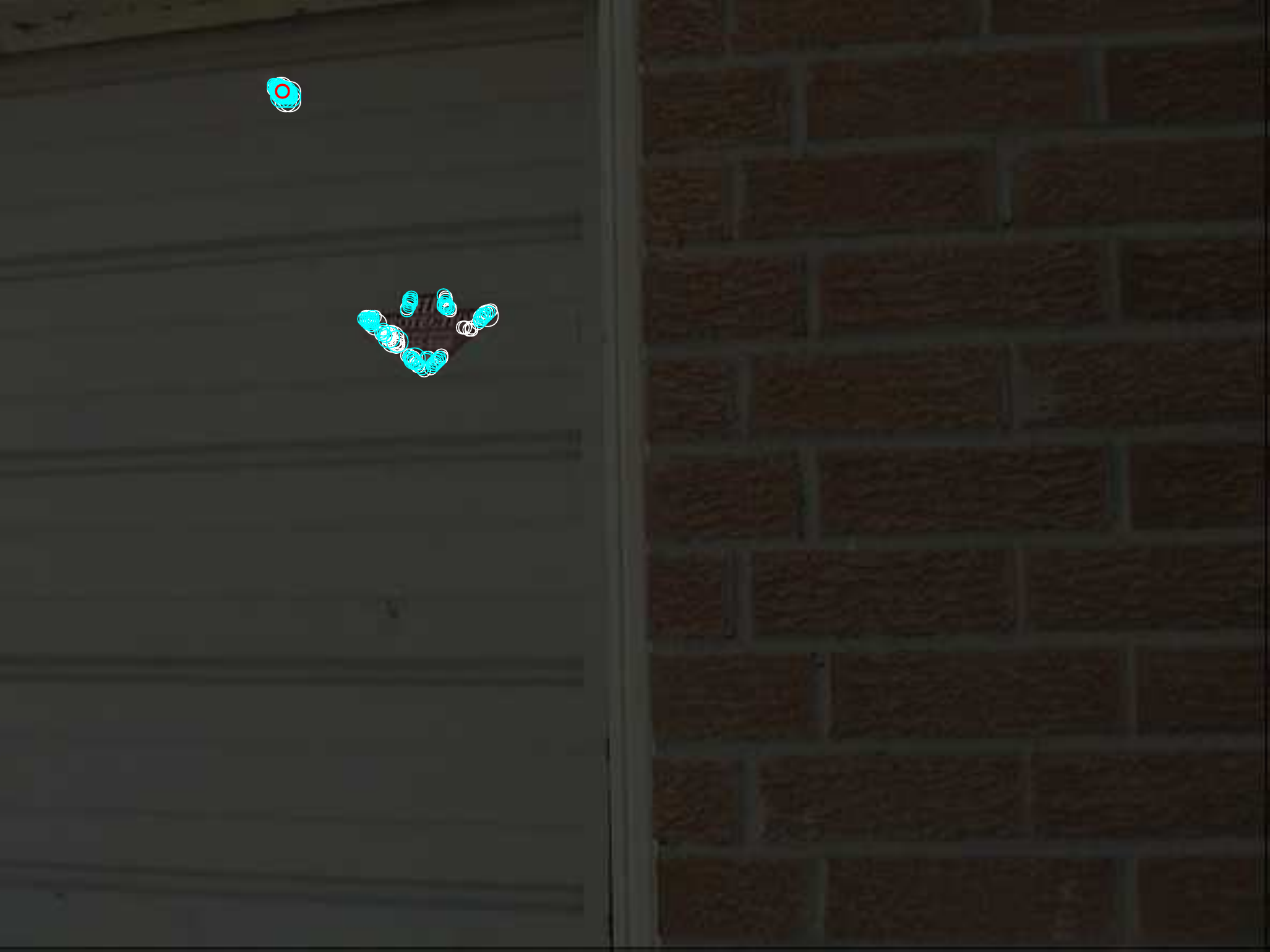}
\end{minipage}
\caption{Impact of parameters for the detection using \citet{davy2018reducing}. The two parameters studied are the size of the patch and the number of patches.}
\label{fig:davy_impact_parameters}
\end{figure*}

\begin{figure*}
\centering
\footnotesize
\hspace{1em}\vphantom{A}\vphantom{[} $4\times 4$ patches\hspace{6em}$8\times 8$ patches\hspace{5em}$16\times 16$ patches\hspace{5em}$32\times 32$ patches
\begin{minipage}[c]{\textwidth}
\centering
\rot{\hspace{1em}\vphantom{A}\vphantom{[} $20\times 20$ region}
\includegraphics[height=0.116\textheight]{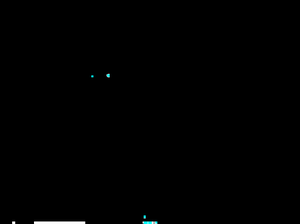}
\includegraphics[height=0.116\textheight]{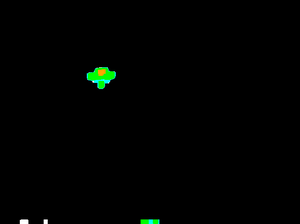}
\includegraphics[height=0.116\textheight]{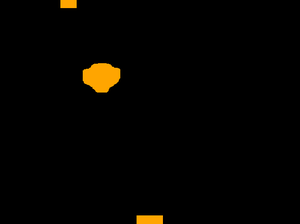}
\includegraphics[height=0.116\textheight]{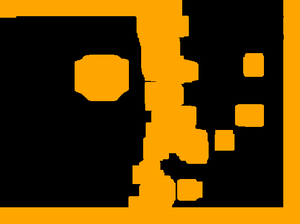}
\end{minipage}

\begin{minipage}[c]{\textwidth}
\centering
\rot{\hspace{1em}\vphantom{A}\vphantom{[} $40\times 40$ region}
\includegraphics[height=0.116\textheight]{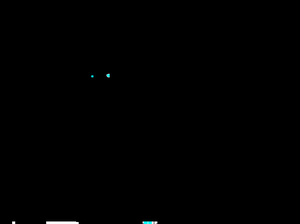}
\includegraphics[height=0.116\textheight]{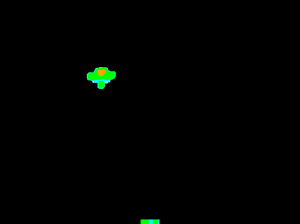}
\includegraphics[height=0.116\textheight]{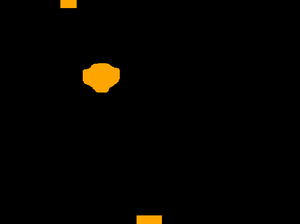}
\includegraphics[height=0.116\textheight]{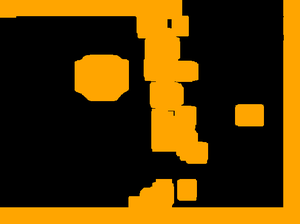}
\end{minipage}

\begin{minipage}[c]{\textwidth}
\centering
\rot{\hspace{1em}\vphantom{A}\vphantom{[} $80\times 80$ region}
\includegraphics[height=0.116\textheight]{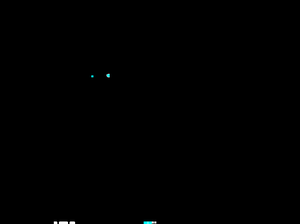}
\includegraphics[height=0.116\textheight]{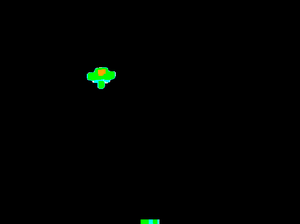}
\includegraphics[height=0.116\textheight]{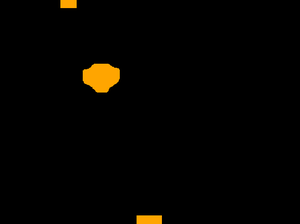}
\includegraphics[height=0.116\textheight]{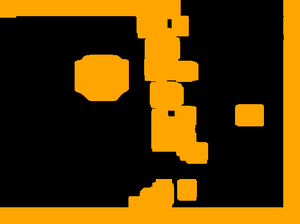}
\end{minipage}

\begin{minipage}[c]{\textwidth}
\centering
\rot{\hspace{0.6em}\vphantom{A}\vphantom{[} $160\times 160$ region}
\includegraphics[height=0.116\textheight]{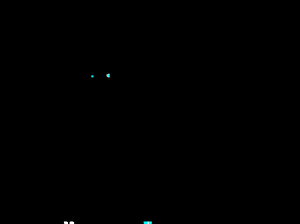}
\includegraphics[height=0.116\textheight]{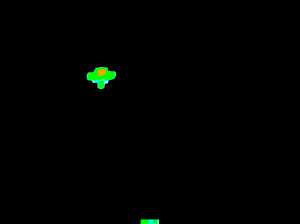}
\includegraphics[height=0.116\textheight]{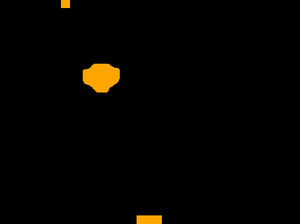}
\includegraphics[height=0.116\textheight]{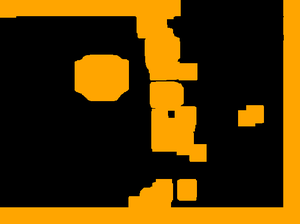}
\end{minipage}
\caption{Impact of parameters for the detection using \citet{zontak2010defect}. The two parameters studied are the size of the patch and the size of the region used for the computation.}
\label{fig:zontak_impact_parameters}
\end{figure*}

\begin{figure*}
\centering
\footnotesize
\hspace{1em}\vphantom{A}\vphantom{[} $8\times 8$ patches\hspace{5em}$16\times 16$ patches\hspace{5em}$32\times 32$ patches
\begin{minipage}[c]{\textwidth}
\centering
\rot{\hspace{1.5em}\vphantom{A}\vphantom{[} $8$ neighbors}
\includegraphics[height=0.116\textheight]{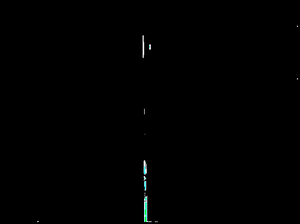}
\includegraphics[height=0.116\textheight]{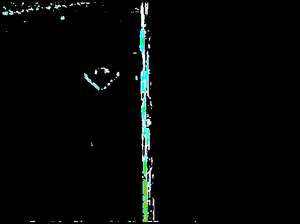}
\includegraphics[height=0.116\textheight]{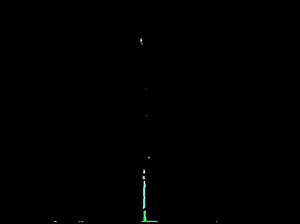}
\end{minipage}

\begin{minipage}[c]{\textwidth}
\centering
\rot{\hspace{1.2em}\vphantom{A}\vphantom{[} $16$ neighbors}
\includegraphics[height=0.116\textheight]{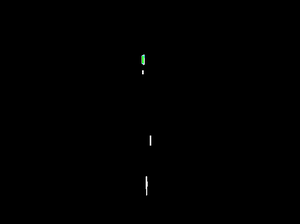}
\includegraphics[height=0.116\textheight]{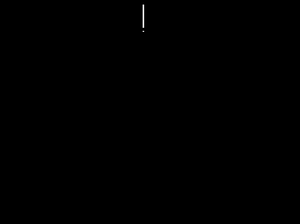}
\includegraphics[height=0.116\textheight]{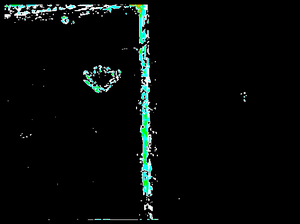}
\end{minipage}

\begin{minipage}[c]{\textwidth}
\centering
\rot{\hspace{1.2em}\vphantom{A}\vphantom{[} $32$ neighbors}
\includegraphics[height=0.116\textheight]{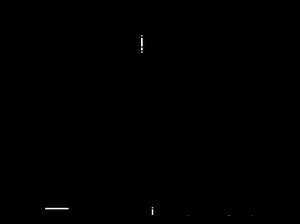}
\includegraphics[height=0.116\textheight]{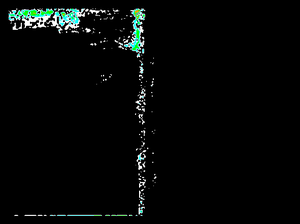}
\includegraphics[height=0.116\textheight]{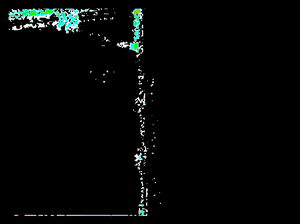}
\end{minipage}
\caption{Impact of parameters for the detection using \citet{mishne2014multiscale}. The two parameters studied are the size of the patch and the number of patches.}
\label{fig:mishne_impact_parameters}
\end{figure*}

\begin{figure*}
\centering
\footnotesize
\hspace{1em}\vphantom{A}\vphantom{[} $5\times 5$ patches\hspace{6em}$10\times 10$ patches\hspace{5em}$15\times 15$ patches\hspace{5em}$20\times 20$ patches
\begin{minipage}[c]{\textwidth}
\centering
\rot{\vphantom{A}\vphantom{[} $1.2\times$ redundancy}
\includegraphics[height=0.116\textheight]{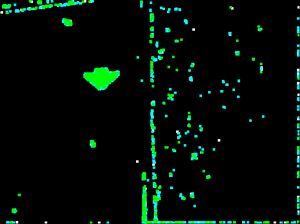}
\includegraphics[height=0.116\textheight]{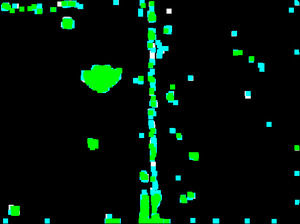}
\includegraphics[height=0.116\textheight]{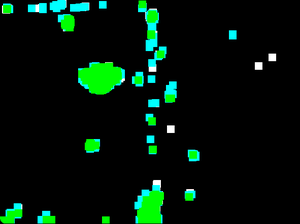}
\includegraphics[height=0.116\textheight]{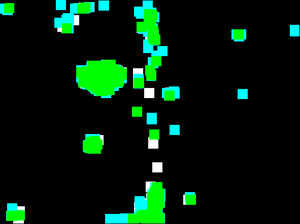}
\end{minipage}

\begin{minipage}[c]{\textwidth}
\centering
\rot{\vphantom{A}\vphantom{[} $1.5\times$ redundancy}
\includegraphics[height=0.116\textheight]{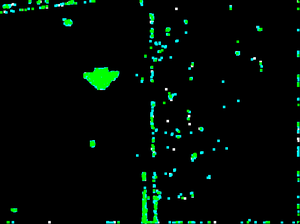}
\includegraphics[height=0.116\textheight]{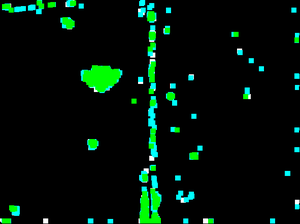}
\includegraphics[height=0.116\textheight]{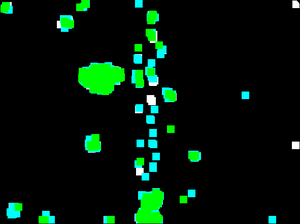}
\includegraphics[height=0.116\textheight]{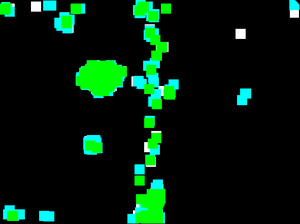}
\end{minipage}

\begin{minipage}[c]{\textwidth}
\centering
\rot{\hspace{0.5em}\vphantom{A}\vphantom{[} $2\times$ redundancy}
\includegraphics[height=0.116\textheight]{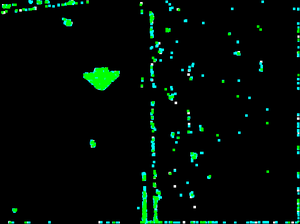}
\includegraphics[height=0.116\textheight]{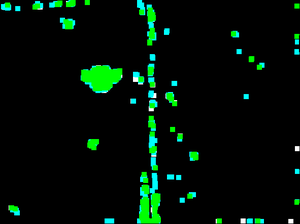}
\includegraphics[height=0.116\textheight]{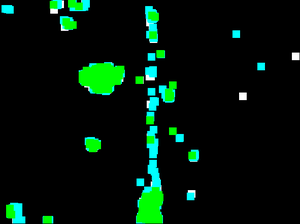}
\includegraphics[height=0.116\textheight]{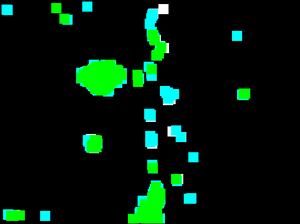}
\end{minipage}

\begin{minipage}[c]{\textwidth}
\centering
\rot{\hspace{0.5em}\vphantom{A}\vphantom{[} $4\times$ redundancy}
\includegraphics[height=0.116\textheight]{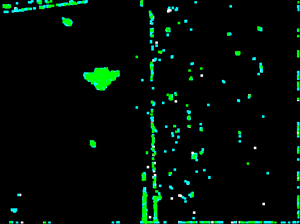}
\includegraphics[height=0.116\textheight]{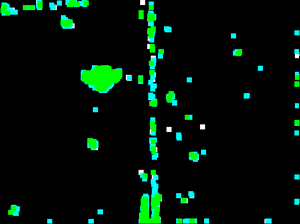}
\includegraphics[height=0.116\textheight]{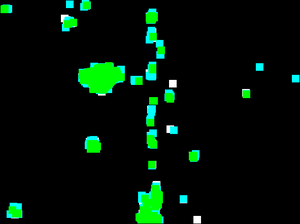}
\includegraphics[height=0.116\textheight]{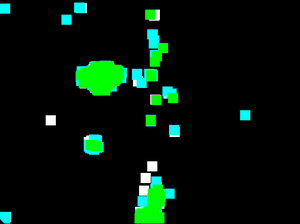}
\end{minipage}
\caption{Impact of parameters for the detection using \citet{boracchi2014novelty}. The two parameters studied are the size of the patch and the redundancy of the dictionary.}
\label{fig:boracchi_impact_parameters}
\end{figure*}

Until now we considered the parameters suggested in the corresponding papers. While it can be interesting to fine tune parameters depending on the application we wanted to stay as generic as possible which led us to fix the same parameters for all the experiments, whatever the type of images, for a given method. In this section we show qualitatively that the parameters impact little on the detection results: playing with the parameters neither adds new interesting detections, nor reduces the quantity of false detections. To evaluate that we selected a few images from our qualitative testing set and computed the results with different sets of parameters. The different experiments are presented in Figures \ref{fig:davy_impact_parameters}, \ref{fig:zontak_impact_parameters}, \ref{fig:mishne_impact_parameters} and \ref{fig:boracchi_impact_parameters}. There is actually a non negligible difference for \citet{zontak2010defect}, the reason is probably that the model assumed during the derivation of the NFA is not completely valid. It is also interesting to see that using not too big patches allows to keep a good precision of the detected region. Nevertheless this experiment validates the choice of parameters for the different models, as the detections are not too drastically different for most methods. We specify here the different parameters used for the different methods:
\begin{enumerate}
    \item \citet{boracchi2014novelty}: $15\times 15$ patches with a redundancy of $1.5$;
    \item \citet{davy2018reducing}: $8\times 8$ patches with $16$ nearest neighbors;
    \item \citet{mishne2017diffusion}: $8\times 8$ patches with $16$ nearest neighbors;
    \item \citet{zontak2010defect}: $8\times 8$ patches with a region of size $160\times 160$, we also set $h$ the similarity parameter to the known noise level $\sigma$ as it seems to work best in practice.
\end{enumerate}

\subsection{Computation time analysis}
In this section we do a brief computation time analysis. All algorithms have wildly different computation times. For example  \citet{aiger2010phase} method is really fast as no really complex computations are needed. On the contrary the  \citet{mishne2014multiscale} method is really slow. Table \ref{tab:computation_time} summarizes the computation time for the different algorithms for the parameter used for the experiment. It's worth noting that for the larger parameters the  \citet{mishne2014multiscale} method requires many hours to compute a single result. It's also worth noting that even though the  \citet{boracchi2014novelty} and \citet{davy2018reducing} algorithms are not the fastest ones, the dictionaries of patches and indexes for the searches can be precomputed and therefore accelerated for fast industrial applications. For example the processing of \citet{boracchi2014novelty} only takes 12s when the dictionary is prelearned. The computation time estimation was done on a core i7-7820HQ 2.90GHz using authors' code whenever it was available (\cite{boracchi2014novelty}, \cite{davy2018reducing} and \cite{mishne2014multiscale} are multithreaded so actual computation times are reported. We report $1/8$ of the actual computation time for \cite{aiger2010phase}, \cite{grosjean2009contrario} and \cite{zontak2010defect} for a fair comparison). 

\begin{table*}[]
    \centering
    \resizebox{\textwidth}{!}{
    \begin{tabular}{ @{}c  c  c  c  c  c @{}}
            \toprule
            \citet{aiger2010phase} & \citet{boracchi2014novelty} & \citet{davy2018reducing} & \citet{grosjean2009contrario} & \citet{mishne2014multiscale} & \citet{zontak2010defect} \\
            \midrule
            0.09 & 1375 & 57 & 1.4 & 749 & 394 \\
            \bottomrule
    \end{tabular}
    }
    \caption{Computation time (in seconds) for the different methods reviewed in details with the parameter chosen for the experiments for the door image (size: $600\times450$.}
    \label{tab:computation_time}
\end{table*}

\section{Discussion and conclusions}
\label{sec:discussion}
Our  analysis and experiments  seem to confirm the view that generic anomaly detection methods  can be built on purely  qualitative assumptions. Such methods do not  require a learning database for the  background or the anomalies,  but  can  learn directly normality from a  single image in which  anomalies may  be present. Why not using more images?  Certainly disposing of a ``golden reference'' or even of a database of ``golden references'' may seem to be ideal situation. But the majority and the best methods succeed to work with a single image. For some methods though, or applications, disposing of a database can help enhance the results and the computation time (by precomputing a dictionary for example).
This success of detecting on a single image is of course possible  only  under  the  assumption that anomalies are a minor part  of  the  image. Some of the  most performing methods use anyway only a small part of the image samples,  processing locally in the image domain on in the  sample domain. Using the present image also has the  advantage of providing an updated background. 

Since anomalies cannot be  modeled,  the focus of attention of all methods  is the  background model. Methods giving a stochastic model to the  background, parametric or not,    could only be applied to restricted classes of  background.  For this reason,  our attention  has been drawn to the thriving \textit{qualitative} background  models. 
Any  assumption  about  a kind of global or  local background homogeneity is \textit{a priori} acceptable.  The  most  restrictive models assume that  the background is periodic,  or  smooth or  even low  dimensional.  This kind of strong  regularity assumption is not  extensible to any  image.  

Another common sense principle is put forward by local contrast center-surround detectors,  which anomalies generate local anomalous contrast.  Yet center-surround methods suffer from the  difficulty of defining a universal detection  rule. 

A more clever idea  has emerged with the  \citet{aiger2010phase} method, which is  to transform the  background into a homogeneous texture while the  anomalies would still stand out.  

Meanwhile the old idea  of  performing a background subtraction  remains  quite valid.  Indeed, as pointed out still  very recently in \cite{tout2016fully}, background subtraction  may be used to return to an elementary  background model for the \textit{residual} that might contain only noise.

The  most  general background models are  merely qualitative.  We singled out two  of them as the most  recent and  powerful ones: the  \textit{sparsity} assumption and the \textit{self-similarity} assumption. 
We found that  two recent exponents use these assumptions to perform a sort of background subtraction:  \citet{carrera2017defect} for sparsity and \citet{davy2018reducing} for self-similarity.

We compared methods on various examples in Section \ref{sec:experiments} and found some methods tend to work better on these various inputs than others, but no method stands out as the best on all images. For applications of anomaly detection, we advise using methods which background model describes the best the expected anomaly-free background, as it will generally lead to the best performance. In our quantitative experiments, Section \ref{sec:exp_quantitative}, \citet{aiger2010phase}'s background model was closest to the background of our synthetic examples, and got the best AUC.

Furthermore, we found that  all methods required a strict  control of the  number of  false alarms to become  universal. Indeed most methods were originally presented with at best an empirical threshold and at worst a comment saying that the threshold depends on the application. The  first  method  proposing this  is the one by~\citet{grosjean2009contrario}, and it was recently extended in \citet{davy2018reducing}.  
Since~\cite{grosjean2009contrario} requires a background stochastic model,  we concluded that a  good  universal model  should:  
\begin{itemize}
\item subtract a background model that  is  merely  qualitative (self-similar,  sparse);
\item handle the residual as a stochastic process to detect anomalies as anomalies in a colored noise;
\item possibly also whiten the residual before detecting the  anomaly.
\end{itemize}

This way, most methods are generalized in  a common framework.
We tested three such syncretic methods and compared them favorably with the three other most relevant methods taken from the  main classes of  background models.  Our  comparative  tests were made on  very diverse  images.  Our quantitative comparison tests were made on simulated  ground truths with stochastic background.  

Both tests seem to validate  the  possibility of  detecting anomalies with very few false alarms using a merely qualitative background model.  This fact is  both surprising and exciting.  It  confirms that  there has been significant progress in the  past  decade.
We  hope that  this study, at the  very least, provides users with useful generic tools that  can be combined for  any  detection  task.

\appendix 

\section{Appendix: Dual formulation of sparsity models\label{ap:dual_sparsity}}

Sparsity based variational methods lack the  direct interpretation  enjoyed by other methods  as  to the proper definition of an  anomaly.  By reviewing the first simplest method of this  kind proposed   in \cite{boracchi2014novelty}, we shall see that its dual interpretation  points to the  detection  of the worst anomaly.
Let $D$ a dictionary representing ``normal'' image patches. For a given patch $p$ the normal patch corresponding to $p$ is $\hat{p}=D\hat{x}$ where
$$
\hat{x} = \argmin_x \left\{ \frac{1}{2}\|p-Dx\|_2^2 + \lambda \|x\|_1 \right\}.
$$
One can derive the following dual optimization problem: Let $z = p-Dx$,
$$
\min_x \left\{ \frac{1}{2}\|z\|_2^2 + \lambda \|x\|_1 \right\}
\text{ s.t } z=p-Dx .
$$
The Lagrangian is in this case
\begin{align*}
\mathcal{L}(x,z,\eta) &= \frac{1}{2}\|z\|_2^2 + \lambda \|x\|_1 + \eta^T(p - Dx - z)\\
&= \eta^Tp + \left(\frac{1}{2}\|z\|_2^2 - \eta^Tz\right) + (\lambda \|x\|_1 - \eta^TDx).
\end{align*}
The dual problem is then 
\begin{align*}
\mathcal{G}(\eta) &= \inf_{x,z} \mathcal{L}(x,z,\eta)\\
&= \eta^T 
p
+ \inf_{z}\left(\frac{1}{2}\|z\|_2^2 - \eta^Tz\right) + \inf_{x}(\lambda \|x\|_1 - \eta^TDx).
\end{align*}

Consider  first $\inf_{z}\left(\frac{1}{2}\|z\|_2^2 - \eta^Tz\right)$: This part is differentiable in $z$ so that
$$
\partial_z \left(\frac{1}{2}\|z\|_2^2 - \eta^Tz\right) = z - \eta
$$
therefore the inf is achieved for $z=\eta$.
The inf is in this case
$$
\inf_{z}\left(\frac{1}{2}\|z\|_2^2 - \eta^Tz\right) = -\frac{1}{2}\|\eta\|_2^2
$$

As for $\inf_{x}(\lambda \|x\|_1 - \eta^TDx)$:
This part is not differentiable (because not smooth) nevertheless the subgradient exists. Let $v$ such that $\|x\|_1 = v^Tx$ (for all i $v_i \in {-1, 1}$). The subgradient of $\|.\|_1$ gives $v$.
\begin{align*}
\partial_x \left(\lambda \|x\|_1 - \eta^TDx\right) &= \partial_x \left(\lambda v^Tx - \eta^TDx\right)\\
&= \lambda v - D^T\eta
\end{align*}
A necessary condition to attain the infimum is then $0 \in \{\lambda v - D^T\eta\}$.
This leads to $v = \frac{D^T\eta}{\lambda}$ with the condition that $\|D^T\eta\|_\infty \leq \lambda$ (because $\|v\|_\infty \leq 1$) which can be injected into the previous equation which gives 
\begin{align*}
\inf_{x}(\lambda \|x\|_1 - \eta^TDx) &= \inf_{x}(\lambda v^Tx - \eta^TDx)\\
&= \lambda (\frac{D^T\eta}{\lambda})^T x - \eta^T Dx\\
&= \eta^T Dx - \eta^T Dx\\
&= 0
\end{align*}
Finally,
\begin{equation*}
\mathcal{G}(\eta) = \eta^Tp - \frac{1}{2}\|\eta\|_2^2
\end{equation*}
Therefore the dual problem is 
\begin{equation*}
\sup_\eta \left\{\eta^Tp - \frac{1}{2}\|\eta\|_2^2\right\} \text{ s.t. } \|D^t\eta\|_\infty \leq \lambda
\end{equation*}
which is equivalent to 
\begin{equation*}
\sup_\eta \left\{-\frac{1}{2}\|p - \eta\|_2^2\right\} \text{ s.t. } \|D^t\eta\|_\infty \leq \lambda.
\end{equation*}
It can be reformulated in a penalized version as 
\begin{equation}\label{dualsparsity_appendice}
\hat{\eta} = \argmin_\eta \left\{\frac{1}{2}\|p - \eta\|_2^2 + \lambda' \|D^T\eta\|_\infty\right\}.
\end{equation}

While $D\hat{x}$ represents the ``normal'' part of the patch $p$, $\hat{\eta}$ represents the anomaly. Indeed, the condition $\|D^T\eta\|_\infty \leq \lambda$ imposes to $\eta$ to be far from the patches represented by $D$. 
Moreover, for a solution $\eta^*$ of the dual to exist (and so that the duality gap doesn't exist) it requires that $\eta^* = p - Dx^*$ \textit{i.e.} $p = Dx^* + \eta^*$ which confirms the previous observation. Notice that  the  solution of \eqref{dualsparsity_appendice} exists by an obvious compactness argument and is unique by the strict convexity of  the dual functional.

\bibliographystyle{spmpscinat}

\bibliography{main}

\setcounter{tocdepth}{3}
\end{document}